\documentclass[11pt]{article}
\usepackage{shortcuts_OPT}
\RequirePackage[OT1]{fontenc}
\RequirePackage{amsthm,amsmath}
\usepackage{graphicx}
\usepackage[top=1in, left=1in, right=1in, bottom=1in]{geometry}
\setcitestyle{open={(},close={)}}

\begin{document}

\title{A Class of Two-Timescale Stochastic EM Algorithms for Nonconvex Latent Variable Models}

\author{\textbf{Belhal Karimi} \ and \ \textbf{Ping Li}\\\\
Cognitive Computing Lab\\
Baidu Research\\
10900 NE 8th St. Bellevue, WA 98004, USA\\
\texttt{\{belhalkarimi, liping11\}@baidu.com}
}

\date{\vspace{0.5in}}

\maketitle

\begin{abstract}
\noindent The\footnote{Preliminary results appeared in Proceedings of the {IEEE} International Symposium on Information Theory (ISIT), 2021.} Expectation-Maximization (EM) algorithm is a popular choice for learning latent variable models.
Variants of the EM have been initially introduced by~\citet{neal1998view}, using incremental updates to scale to large datasets, and by~\citet{wei1990monte, delyon1999}, using Monte Carlo (MC) approximations to bypass the intractable conditional expectation of the latent data for most nonconvex models.
In this paper, we propose a general class of methods called Two-Timescale EM Methods based on a two-stage approach of stochastic updates to tackle an essential nonconvex optimization task for latent variable models.
We motivate the choice of a double dynamic by invoking the variance reduction virtue of each stage of the method on both sources of noise: the index sampling for the incremental update and the MC approximation.
We establish finite-time and global convergence bounds for nonconvex objective functions.
Numerical applications on various models such as deformable template for image analysis or nonlinear models for pharmacokinetics are also presented to illustrate~our~findings.
\end{abstract}

\newpage

\section{Introduction}

Learning latent variable models is critical for many important modern machine learning problems, see for instance~\citet{mclachlan2007algorithm} for references.
We formulate the training of this type of model as the following {empirical risk minimization} problem:
\begin{align} \label{eq:em_motivate}
\begin{split}
 \min_{ \param \in \Param }~ \overline{\calL} ( \param ) \eqdef  \calL ( \param ) + \Pen (\param) \quad \text{with}~~\calL ( \param ) = \frac{1}{n} \sum_{i=1}^n \calL_i( \param) \eqdef  \frac{1}{n} \sum_{i=1}^n \big\{ - \log g( y_i ; \param ) \big\}\eqs,
\end{split}
\end{align}
where $\{y_i\}_{i=1}^n$ are observations, $\Param \subset \rset^d$ is the parameters set and $\Pen : \Param \rightarrow \rset$ is a smooth regularizer.
The objective $ \overline{\calL} ( \param )$ is possibly {nonconvex} and is assumed to be lower bounded.
In the latent data model, the likelihood $g(y_i ; \param)$, is the marginal distribution of the complete~data likelihood, noted $f(z_i,y_i; \param)$, such that for a compact set $\Zset \subset \rset^p$
\begin{align}
g(y_i; \param) = \int_{\Zset} f (z_i,y_i;\param) \mu(\rmd z_i)\, ,
\end{align}
where $\{ z_i \}_{i=1}^n$ are the vectors of latent variables associated to the observations $\{y_i\}_{i=1}^n$.
In this paper, we assume that the complete data likelihood belongs to the curved exponential family~\citep{efron1975defining}, i.e.,
\beq \label{eq:exp}
f(z_i,y_i; \param) = h  (z_i,y_i) \exp ( \pscal{S(z_i,y_i)}{\phi(\param)} - \psi(\param) )\eqs,
\eeq
where $\psi(\param)$, $h(z_i,y_i)$ are scalar functions, $\phi(\param) \in \rset^k$ is a vector function, and $\{S(z_i,y_i) \in \rset^k\}_{i=1}^n$ is the vector of sufficient statistics.
Batch EM~\citep{dempster1977Maximum, wu1983convergence}, the method of reference for~\eqref{eq:em_motivate}, is comprised of two steps.
At iteration $k$, the {E-step} computes the conditional expectation of the sufficient statistics of~\eqref{eq:exp}, noted $\overline{\bss}(\param)= \frac{1}{n} \sum_{i=1}^n \overline{\bss}_i(\param)$, where for all $\param \in \Param$ and $i \in \inter$, where $\inter \eqdef \{1, \cdots, n\}$:
\begin{align}\label{eq:definition-overline-bss}
 \overline{\bss}_i(\hat{\param}^{(k)}) \eqdef \int_{\Zset} S(z_i,y_i) p(z_i|y_i;\hat{\param}^{(k)}) \mu(\rmd z_i) \eqsp,
\end{align}
and the {M-step} is given by
\begin{align}\label{eq:mstep}
\hat{\param}^{(k+1)}  = \overline{\param}( \overline{\bss}(\hat{\param}^{(k)}) ) \eqdef \argmin_{ \vartheta \in \Param } ~\big\{ \Pen( \vartheta ) + \psi( \vartheta) - \pscal{ \overline{\bss}(\hat{\param}^{(k)})}{ \phi ( \vartheta) } \big\} \eqsp.
\end{align}

There are two main caveats of such a method: {(a)} with the explosion of data, the first step of the EM is computationally inefficient as it requires, at each iteration, a full pass over the dataset; and {(b)} the complexity of modern  models makes the expectation in~\eqref{eq:definition-overline-bss} intractable.
Both of these constraints occur in the E-step of the EM algorithm, see the integral and finite sum structure of~\eqref{eq:definition-overline-bss} and we tackle them jointly in this contribution.

\subsection{Prior Work}
Inspired by stochastic optimization procedures,~\citet{neal1998view,cappe2009line} developed respectively an incremental and an online variant of the {E-step} in models where the expectation is computable, and were then extensively used and studied in~\citet{nguyen2020mini, liang2009online,cappe2011online}.
Some improvements of those methods have been provided and analyzed, globally and in finite-time, in~\citet{karimi2019global} where variance reduction techniques taken from the optimization literature have been efficiently applied to scale the EM algorithm to large datasets. Follow-up studies on variance reduced stochastic EM include~\citet{fortem2020,fort2021geom}.
Regarding the computation of the expectation under the posterior distribution, the Monte Carlo EM (MCEM) has been introduced in~\citet{wei1990monte} where a Monte Carlo (MC) approximation for this expectation is computed. Its convergence is established in \citet{fort2003convergence}. A variant of that algorithm is the Stochastic Approximation of the EM (SAEM) in~\citet{delyon1999} leveraging the power of Robbins-Monro update~\citep{robbins1951stochastic} to ensure pointwise convergence of the vector of estimated parameters using a decreasing stepsize rather than increasing the number of MC samples.
The MCEM and the SAEM have been successfully applied in mixed effects models~\citep{mcculloch1997maximum,hughes1999mixed,baey2016nonlinear} or to do inference for joint modeling of time-to-event data coming from clinical trials in~\citet{das2010Inferences}, unsupervised clustering in~\citet{ngChoice2003}, variational inference of graphical models in~\citet{BleiVariational2017} among other applications.
An incremental variant of the SAEM was proposed in~\citet{kuhn2019properties} but its analysis is limited to asymptotic consideration.
Gradient-based methods have been developed and analyzed in~\citet{zhu2017high} but remain out of the scope of this paper as they tackle the high-dimensionality issue.

\subsection{Contributions}
This paper introduces and analyzes a new class of methods which purpose is to update two proxies for the target expected quantities in a two-timescale manner.
Those approximated quantities are then used to optimize the objective function~\eqref{eq:em_motivate} for challenging examples (nonlinear) and settings (large-scale) using the {M-step} of the EM algorithm.
Our main contributions can be summarized as follows:
\begin{itemize}
\item We propose a two-timescale method based on (i) stochastic approximation (SA), to alleviate the burden of computing MC approximations, and on (ii) incremental updates, scaling to large datasets. We describe the edges of each level of our method based on variance reduction arguments. Such class of algorithms has two advantages. First, it naturally leverages variance reduction and Robbins-Monro type of updates to tackle large-scale and highly nonconvex learning tasks. Then, it gives a simple formulation as a {scaled-gradient method} which makes the analysis and implementation accessible.
\item We also establish global (independent of the initialization) and finite-time (true at each iteration) upper bounds on a classical sub-optimality condition~\citep{jain2017non, ghadimi2013stochastic}, \ie the second order moment of the gradient of the objective function.
We discuss the double dynamic of those bounds due to the two-timescale property of our algorithm update and we theoretically show the advantages of introducing variance reduction in a {stochastic approximation}~\citep{robbins1951stochastic} scheme.
\item Our theoretical findings include MC sampling noise contrary to existing studies related to the EM where the expectations are computed exactly.
Adding a layer of MC approximation and the SA step to reduce its variance introduce new challenges that need careful considerations and account for the originality of our research paper, both on the algorithmic and theoretical plans.
\item Numerical experiments are presented in this contribution on a variety of models and datasets.
In particular, we provide empirical insights on the edges of our method for learning latent variable models in image analysis and pharmacokinetics.
\end{itemize}


In Section~\ref{sec:tts} we formalize both incremental and Monte Carlo variants of the EM.
We introduce our two-timescale class of EM (TTSEM) algorithms for which we derive several statistical guarantees in Section~\ref{sec:mainanalysis} for possibly {nonconvex} functions.
Section~\ref{sec:proofs} corresponds to the sketches of the proofs for our main results.
Section~\ref{sec:numerical} is devoted to the numerical experiments showing the benefits of our methods on several tasks and datasets.
Proofs and additional experimental details are deferred to the Appendix.

\section{Two-Timescale Stochastic EM Algorithms}\label{sec:tts}

We recall and formalize in this section the different methods found in the literature that aim at solving the intractable expectation problem and the large-scale problem.
We then introduce our class of stochastic methods that efficiently tackles the optimization problem in~\eqref{eq:em_motivate}.

\subsection{Monte Carlo Integration and Stochastic Approximation}

As mentioned in the introduction, for complex and possibly nonconvex models, the expectation under the posterior distribution defined in~\eqref{eq:definition-overline-bss} is not tractable. In that case, the first solution involves computing a Monte Carlo integration of that expectation.
For all $ i \in \inter$, draw $M$ samples, noted $\{z_{i,m} \sim p(z_i|y_i;\param)\}_{m=1}^{M}$, and compute the MC integration, noted $\tilde{S}$, of $\overline{\bss}(\param) \eqdef \frac{1}{n} \sum_{i=1}^n \overline{\bss}_i(\param)$ where each of its component is defined by~\eqref{eq:definition-overline-bss}:
\beq\label{eq:mcstep}
\textrm{MC-step}:~ \tilde{S} \eqdef \frac{1}{n} \sum_{i=1}^n\frac{1}{M} \sum_{m=1}^M S(z_{i,m}, y_i)\eqs.
\eeq
Then, update the parameter via the maximization function $\overline{\param}(\tilde{S})$.
This algorithm, called the MCEM~\citep{wei1990monte}, bypasses the intractable expectation issue but is rather computationally expensive.
Indeed, in order to reach pointwise convergence, the number of samples $M$ needs to be increasingly large.
An alternative to the MCEM is to use a Robbins-Monro (RM) type of update, see~\citet{robbins1951stochastic}.
We denote, for $k >0 $, the number of samples $M_k$ and the approximation by $\tilde{S}^{(k+1)}$:
\beq\label{eq:stats}
\begin{split}
 \tilde{S}^{(k+1)} \eqdef \frac{1}{n} \sum_{i=1}^n \tilde{S}^{(k+1)}_i = \frac{1}{n} \sum_{i=1}^n\frac{1}{M_k} \sum_{m=1}^{M_k} S(z_{i,m}^{(k)}, y_i) \eqs,
\end{split}
\eeq
where for $m \in [M_k]$, $z_{i,m}^{(k)} \sim p(z_i|y_i;\param^{(k)})$.
Then, the RM update of the statistics $\hat{\bss}^{(k+1)}$ reads:
\beq\label{eq:rmstep}
\textrm{SA-step}:~ \hat{\bss}^{(k+1)} =  \hat{\bss}^{(k)}  + \gamma_{k+1}(\tilde{S}^{(k+1)} - \hat{\bss}^{(k)} )\eqs,
\eeq
where $\{ \gamma_{k} \}_{k>1} \in (0,1)$ is a sequence of decreasing stepsizes to ensure asymptotic convergence.
The combination of~\eqref{eq:stats} and~\eqref{eq:rmstep} is called the Stochastic Approximation of the EM (SAEM) and has been shown to converge to a maximum likelihood of the observations under very general conditions, see~\citet{delyon1999} for a proof of convergence.
In simple scenarios, the samples $\{z_{i,m}\}_{m=1}^{M}$ are conditionally independent and identically distributed with distribution $p(z_i,\param)$.
Nevertheless, in most cases, since the loss function between the observed data $y_i$ and the latent variable $z_i$ can be nonconvex, sampling exactly from this distribution is not an option and the MC batch is sampled by Markov Chain Monte Carlo (MCMC) algorithm~\citep{brooks2011handbook,meyn2012markov}.
It is proved in~\citet{kuhn2004coupling} that~\eqref{eq:rmstep} converges almost surely when coupled with an MCMC procedure.

\vspace{0.1in}
\noindent \textbf{Role of the stepsize $\gamma_k$:}  The sequence of decreasing positive integers $\{ \gamma_{k} \}_{k>1}$ controls the convergence of the algorithm.
It is inefficient to start with small values for the stepsize $\gamma_k$ and large values for the number of simulations $M_k$.
Rather, it is recommended that one decreases $\gamma_k$, as in $\gamma_k = 1/k^\alpha$, with $\alpha \in (0,1)$, and keeps a constant and small number of samples $M_k$, hence bypassing the computationally involved sampling step in~\eqref{eq:mcstep}.
 In practice, $\gamma_k$ is set equal to $1$ during the first few iterations to let the iterates explore the parameter space without memory and converge quickly to a neighborhood of the target estimate.
 The Stochastic Approximation is performed during the remaining iterations ensuring the almost sure convergence of the vector of estimates.
This Robbins-Monro type of update constitutes the {first level} of our algorithm, needed to temper the variance and noise introduced by the Monte Carlo integration.
In the next section, we derive variants of this algorithm to adapt to the sheer size of data of modern applications and formalize the {second level} of our class of two-timescale EM methods.

\subsection{Incremental and Two-Timescale Stochastic EM Methods} \label{sec:sEM}

Efficient strategies to scale to large datasets include incremental~\citep{neal1998view} and variance reduced~\citep{johnson:zhang:2013,chen2018stochastic} methods.
We explicit a general update that covers those latter variants and that represents the {second level} of our algorithm, \ie the incremental update of the noisy statistics $\tilde{S}^{(k+1)}$ in~\eqref{eq:stats}.
Instead of computing its full batch $\tilde{S}^{(k+1)}$ as in~\eqref{eq:stats}, the MC approximation is incrementally evaluated through $\stt^{(k+1)}$ as:
\beq \label{eq:sestep}
\textrm{Inc-step}:~\stt^{(k+1)} = \stt^{(k)} + \rho_{k+1} ( \StocEstep^{(k+1)}- \stt^{(k)}  )\eqs.
\eeq
Note that $\{ \rho_{k} \}_{k>1} \in (0,1)$ is a sequence of stepsizes, $\StocEstep^{(k)}$ is a proxy for $\tilde{S}^{(k)}$ defined in~\eqref{eq:stats}.
When $\rho_{k} = 1$ and $\StocEstep^{(k)} = \tilde{S}^{(k)}$, i.e., computed in a full batch manner as in~\eqref{eq:stats}, then we recover the SAEM algorithm, if $\rho_{k}=1$, $\gamma_{k}=1$ and $\StocEstep^{(k)} = \tilde{S}^{(k)}$, then we recover the MCEM algorithm.

\vspace{0.2in}
\noindent \textbf{Two-Timescale Stochastic EM methods:}
We introduce the general method derived using the two variance reduction techniques described above.
Beforehand, we list in Table~\ref{alg:prox}, variants of the {Inc-step}, stated in~\eqref{eq:sestep}, of Algorithm~\ref{alg:ttsem} for the quantity $\StocEstep^{(k+1)}$, at iteration $k >0$.
Our paper introduces two new variants of the \ISAEM\ (incremental SAEM), introduced by \citet{kuhn2019properties}, namely \SAEMVR\ (variance reduced SAEM) and \FISAEM\ (fast incremental SAEM) in order to accelerate the convergence of the parameters.
For each method, we define a random index noted $i_k \in \inter$ and drawn at iteration $k$, and $\tau_i^k = \max \{ k' : i_{k'} = i,~k' < k \}$ as the iteration index where $i \in \inter$ is last drawn prior to iteration $k$.

\vspace{0.1in}

 \begin{protocol}[h]
  \floatname{algorithm}{Table}
\caption{Proxies for the Incremental-step~\eqref{eq:sestep}}\label{alg:prox}
  \begin{algorithmic}[1]
  \STATE {\ISAEM\ }$\hspace{0.7cm} \StocEstep^{(k+1)} = \StocEstep^{(k)} + n^{-1} ( \tilde{S}_{i_k}^{(k)}  - \tilde{S}_{i_k}^{(\tau_{i_k}^k)} ) \label{eq:isaem}$
    \STATE {\SAEMVR\ }$\hspace{0.5cm} \StocEstep^{(k+1)}  = \stt^{(\ell(k))} +  ( \tilde{S}_{i_k}^{(k)}  -\tilde{S}_{i_k}^{(\ell(k))} ) \label{eq:vrsaem}$
      \STATE {\FISAEM\ }$\hspace{0.6cm} \StocEstep^{(k+1)} = \overline{\StocEstep}^{(k)} + ( \tilde{S}_{i_k}^{(k)}  - \tilde{S}_{i_k}^{(t_{i_k}^k)} )\quad \label{eq:fisaem}$ and
             $ \quad \overline{\StocEstep}^{(k+1)} = \overline{\StocEstep}^{(k)} + n^{-1} ( \tilde{S}_{j_k}^{(k)}  - \tilde{S}_{j_k}^{(t_{j_k}^k)} )$ \label{line:fisaem}
  \end{algorithmic}
\end{protocol}

\vspace{0.1in}

Note that the proposed \FISAEM\ update, Line~3 in Table~\ref{alg:prox}, draws {two} {independent} and {uniform} indices $(i_k, j_k) \in \inter$.
Thus, we define $t_j^k = \{ k' : j_{k'} = j , k' < k \}$ to be the iteration index where the sample $j \in \inter$ is last drawn as $j_k$ prior to iteration $k$ in addition to $\tau_i^k$ which was defined w.r.t. $i_k$.
We recall that $\tilde{S}_{i_k}^{(k)} \eqdef  \frac{1}{M_k} \sum_{m=1}^{M_k} S(z_{i_k,m}^{(k)}, y_{i_k})$ where $z_{i_k,m}^{(k)}$ are samples drawn from $ p(z_{i_k}|y_{i_k};\param^{(k)})$.
The stepsize in~\eqref{eq:sestep} is set to $\rho_{k+1} = 1$ for the \ISAEM\ method initializing with $\StocEstep^{(0)} = \tilde{S}^{(0)}$; $\rho_{k+1} = \rho$ is  constant for the \SAEMVR\ and \FISAEM.
Note that we initialize with $\overline{\StocEstep}^{(0)} = \tilde{S}^{(0)}$ for the \FISAEM\ which can be seen as a slightly modified version of SAGA inspired by~\citet{reddi2016fast}.
For \SAEMVR\, we set an epoch size of $m$ and we define $\ell(k) \eqdef m \lfloor k/m \rfloor$ as the first iteration number in the epoch that iteration $k$ is in.
Then, our general class of methods, see Algorithm~\ref{alg:ttsem}, leverages both levels~\eqref{eq:rmstep} and~\eqref{eq:sestep} in order to output a vector of fitted parameters $\hat{\param}^{({ K}_{ f })}$ where ${ K}_{ f }$ is the total number of iterations.
\begin{algorithm}[t]
\caption{Two-Timescale Stochastic EM methods.}\label{alg:ttsem}
  \begin{algorithmic}[1]
  \STATE \textbf{Input:} $\hat{\param}^{(0)} \leftarrow \param_0$, $\hat{\bss}^{(0)} \leftarrow \tilde{S}^{(0)}$, $\{\gamma_k\}_{k>0}$, $\{\rho_k\}_{k>0}$ and $ { K}_{ f }\in \mathbb{N}^*$.
  \STATE Set the terminating iteration number, $K \in \{0,\dots,{ K}_{ f }-1\}$, as a discrete r.v.~with:
  \begin{equation} \label{eq:random}
   P( K = k ) = \frac{ \gamma_{k} }{\sum_{\ell=0}^{{ K}_{ f }-1} \gamma_\ell} = \frac{\gamma_k}{{ P}_{ m}}\eqs.
  \end{equation}
  \FOR {$k=0,1,2,\dots, { K}_{ f } - 1$}
  \STATE Draw index $i_k \in \inter$ uniformly (and $j_k \in \inter$ for \FISAEM).
     \STATE Compute $\tilde{S}_{i_k}^{(k)}$ using the { MC-step}~\eqref{eq:mcstep},  for the drawn indices.
   \STATE Compute the surrogate sufficient statistics $\StocEstep^{(k+1)}$ using Lines~1, 2, and 3 in Table~\ref{alg:prox}.
   \STATE Compute $\hat{\bss}^{(k+1)}$ and $\stt^{(k+1)}$ using resp.~\eqref{eq:rmstep} and~\eqref{eq:sestep}:
\beq \label{eq:twolevels}
\begin{split}
& \stt^{(k+1)} = \stt^{(k)} + \rho_{k+1} ( \StocEstep^{(k+1)}- \stt^{(k)}  ) \eqs,\\
&  \hat{\bss}^{(k+1)} =  \hat{\bss}^{(k)}  + \gamma_{k+1}(\stt^{(k+1)} - \hat{\bss}^{(k)} ) \eqs.
\end{split}
\eeq
 \STATE Update $\hat{\param}^{(k+1)} = \overline{\param}(  \hat{\bss}^{(k+1)}) $ via the M-step~\eqref{eq:mstep}.
\ENDFOR
  \end{algorithmic}
\end{algorithm}
The update in~\eqref{eq:twolevels} is said to have a {two-timescale} property as the stepsizes satisfy $\lim_{k \to \infty} \gamma_k/\rho_k < 1$ such that $ \tilde{S}^{(k+1)} $  is updated at a faster time-scale, determined by $\rho_{k+1}$, than $\hat{\bss}^{(k+1)}$, determined by $\gamma_{k+1}$.
The next section introduces the main results of this paper and establishes global and finite-time bounds for the different updates of our scheme.

\newpage

\section{Finite Time Analysis of Two-Timescale EMs} \label{sec:mainanalysis}
\textbf{Notations reminder:}
$\tilde{S}$ represents the MC approximation of its expected counterpart $\overline{\bss}$ at index $i \in \inter$.
$\stt$ denotes the variance-reduced quantity in~\eqref{eq:sestep}, related to the stepsize $\rho$ (assumed constant here), and leveraging the incrementally updated quantity $\StocEstep$ via Table~\ref{alg:prox}.
The quantity noted $\hat{\bss}$ stands for the sufficient statistics resulting from the RM procedure in~\eqref{eq:rmstep} and is updated using the SA stepsize $\gamma$.

Following~\citet{cappe2009line}, it can be shown that stationary points of the objective function~\eqref{eq:em_motivate} corresponds to the stationary points of the following {nonconvex} Lyapunov function:
\beq\label{eq:em_sspace}
\min_{ {\bss} \in \Sset }~  V ( {\bss} ) \eqdef \overline\calL( \op(\bss) ) =  \frac{1}{n} \sum_{i=1}^n  \calL _i (  \op(\bss) )+ \Pen (  \op(\bss) ) \eqs,
\eeq
that we propose to study in this paper and where $\calL _i$ is defined in~\eqref{eq:em_motivate}.

\subsection{Assumptions and Intermediate Lemmas}
In order to derive the desired convergence guarantees, several important assumptions are given below:
\begin{assumption}\label{ass:compact}
The sets $\Zset, \Sset$ are compact set of $\rset^p$. Besides, there exist constants $C_{\Sset}, C_{\Zset}$ such that
\beq \textstyle \notag
\begin{split}
C_{\Sset} \eqdef \max_{ \bss, \bss' \in \Sset } \| \bss - \bss' \| < \infty \quad \textrm{and} \quad  C_{\Zset} \eqdef \max_{i \in \inter} \int_{\Zset} | S(z,y_i) | \mu( \rmd z ) < \infty.
\end{split}
\eeq
\end{assumption}
\begin{assumption}\label{ass:expected}
For any $i \in \inter$, $z \in \Zset$, $\param, \param' \in {\rm int} (\Param)^2$, where ${\rm int} (\Param)$ denotes the interior of $\Param$, we have $\big| p( z | y_i; \param ) - p( z | y_i; \param' ) \big| \leq  \Lip{p} \| \param - \param' \|$.
\end{assumption}
We recall that we consider the curved exponential family such that the objective function satisfies:
\begin{assumption} \label{ass:reg}
For any $\bm{s} \in \Sset$, the function $\param \mapsto L(s,\param) \eqdef \Pen( \param ) + \psi( \param) - \pscal{ \bss}{ \phi ( \param) }$ admits a unique global minimum $\mstep{\bss} \in {\rm int}(\Param)$.
In addition, $\jacob{\phi}{\param}{\overline{\param}(\bss )}$, the Jacobian of the function $\phi$ at $\param$, is full rank, $\Lip{p}$-Lipschitz and $\overline{\param}( \bss )$ is $\Lip{t}$-Lipschitz.
\end{assumption}
We denote by $\hess{L}{\param}(\bss,\param)$ the Hessian (w.r.t to $\param$ for a given value of $\bss \in \Sset$) of the function $\param \mapsto L(\bss,\param)= \Pen(\param) + \psi(\param) -\pscal{\bss}{\phi(\param)}$, and define
$$\operatorname{B}( \bss ) \eqdef\jacob{ \phi }{ \param }{ \mstep{\bss} } ( \hess{L}{\param}( {\bss},  \mstep{\bss} )  )^{-1} \jacob{ \phi }{ \param }{ \mstep{\bss} }^\top \, .$$
\begin{assumption}\label{ass:eigen}
It holds that $ \upsilon_{\max} \eqdef \sup_{\bss \in \Sset} \| \operatorname{B}( \bss ) \| < \infty$ and $0 < \upsilon_{\min}  \eqdef \inf_{\bss \in \Sset} \lambda_{\rm min} ( \operatorname{B}( \bss ) )$.
There exists a constant $\Lip{b}$ such that for all $\bss, \bss' \in \Sset^2$, we have $ \| \operatorname{B}( \bss ) - \operatorname{B}( \bss' )  \| \leq \Lip{b} \| {\bss} - {\bss}' \|$.
\end{assumption}

\vspace{0.1in}

The class of TTSEM methods, summarized in Algorithm~\ref{alg:ttsem}, is composed of two levels where the second stage corresponds to the variance reduction trick used in~\citet{karimi2019global} in order to accelerate incremental methods and reduce the variance introduced by the index sampling step.
The first stage is the Robbins-Monro update that aims at reducing the Monte Carlo noise of $\tilde{S}^{(k+1)}$ at iteration $k$, defined as follows:
\beq\label{eq:mcerror}
\eta_{i}^{(k)} \eqdef \tilde{S}_{i}^{(k)} -  \overline{\bss}_i(\vartheta^{(k)})\quad  \textrm{for all} \quad  i \in \inter \quad \textrm{and} \quad  k > 0 \eqs.
\eeq
We consider that the MC approximation is unbiased if for all $ i \in \inter$ and $m \in [M]$, the samples $z_{i,m} \sim p(z_i|y_i;\param)$ are i.i.d. under the posterior distribution, \ie $\EE[\eta_{i}^{(k)}|{\cal F}_k] = 0$ where  ${\cal F}_k$ is the filtration up to iteration $k$.
The following results are derived under the assumption that the fluctuations implied by the approximation are bounded:
\begin{assumption}\label{ass:mcerror}
For all $k >0$, $i \in \inter$, it holds that
$\EE [\| \eta_{i}^{(k)}\|^2 ] < \infty$ and $\EE[\| \EE[\eta_{i}^{(k)}|{\cal F}_k]\|^2] < \infty \eqs.$
\end{assumption}

Note that typically, the controls exhibited above are vanishing when the number of MC samples $M_k$ increases~with~$k$.

\vspace{0.1in}

We now state two important results on the Lyapunov function; its smoothness:
\begin{lemmacoloured} \label{lem:smooth}
\citep{karimi2019global} Assume A\ref{ass:compact}-A\ref{ass:eigen}.  For all $\bss,\bss' \in \Sset$ and $i \in \inter$, we have
\beq \label{eq:smooth}
\| \overline{\bss}_i ( \overline{\param} ({\bss})) - \overline{\bss}_i ( \overline{\param} ({\bss}' )) \| \leq \Lip{{\bss}} \| {\bss} - {\bss}' \|,~~\| \grd  V ( {\bss} ) - \grd  V ( {\bss}' ) \| \leq \Lip{V} \| {\bss} - {\bss}' \|\eqs,
\eeq
where $\Lip{\bss} \eqdef C_{\Zset} \Lip{p} \Lip{t}$ and $\Lip{V}  \eqdef \upsilon_{\max} ( 1 + \Lip{{\bss}} ) + \Lip{b} C_{\Sset}$.
\end{lemmacoloured}
We also establish a growth condition on the gradient of $V$ related to the mean field of the algorithm:
\begin{lemmacoloured}\label{lem:growth}
Assume A\ref{ass:reg} and A\ref{ass:eigen}. For all $\bss \in \Sset$,
\beq \label{eq:semigrad}
\begin{split}
\upsilon_{\min}^{-1} \pscal{\grd V ( {\bss} ) }{ {\bss} - \os( \op ({\bss})) } \geq \| {\bss} - \os( \op ({\bss})) \|^2 \geq \upsilon_{\max}^{-2} \| \grd V ( {\bss} ) \|^2\eqs.
\end{split}
\eeq
\end{lemmacoloured}
We present in the following a finite-time analysis of our general method described in Algorithm~\ref{alg:ttsem}.
\subsection{Global Convergence of Incremental and Two-Timescale Stochastic EM}
Then, the following non-asymptotic convergence rate can be derived for the \ISAEM\ algorithm:
\begin{theoremcoloured}\label{thm:isaem}
Assume A\ref{ass:compact}-A\ref{ass:mcerror}.
Consider the \ISAEM\ sequence $\{\hat{\bss}^{(k)}\}_{k>0} \in \mathcal{S}$ obtained with $\rho_{k+1}=1$ for any $k \leq { K}_{ f }$ where ${ K}_{ f } > 0$.
Let $\{\gamma_{k} = 1/(k^a \alpha c_1 \overline{L})\}_{k>0}$, where $a \in (0,1)$, be a sequence of stepsizes, $c_1 = \upsilon_{\min}^{-1}$, $\alpha = \max\{8, 1+6\upsilon_{\min}\}$, $\overline{L} = \max\{ \Lip{\bss} , \Lip{V} \}$, $\beta = c_1 \overline{L}/n$, then:
\beq\notag
\begin{split}
 \upsilon_{\max}^{-2}\sum_{k=0}^{{ K}_{ f }} \tilde{\alpha}_k \EE [\|\grd V( \hs{k} )\|^2] \leq   \EE  [V( \hs{0} ) - V( \hs{{ K}_{ f }} ) ] + \sum_{k=0}^{{ K}_{ f }-1} \tilde{\Gamma}_k         \EE [\| \eta_{i_k}^{(k)}\|^2] \eqs.
\end{split}
\eeq
\end{theoremcoloured}
Observe that, in Theorem~\ref{thm:isaem}, the convergence bound is composed of an initialization term $V( \hs{0} ) - V( \hs{{ K}_{ f }} )$ and suffers from the Monte Carlo noise introduced by the posterior sampling step, see the second term on the RHS of the inequality.
We observe, in the next section, that when variance reduction is applied ($\rho_k < 1$), a second phase of convergence manifests.
We now deal with the analysis of Algorithm~\ref{alg:ttsem} when variance reduction is applied \ie\ $\rho <1$.
Let $K$ be an independent discrete r.v.~drawn from $\{1,\dots,{ K}_{ f }\}$ with distribution  $\{ \gamma_{k+1}/ { P}_{ m}\}_{k=0}^{{ K}_{ f } - 1 }$, then, for any ${ K}_{ f } >0 $, the convergence criterion used in our study reads
\beq\notag
\EE[ \| \grd V( \hs{K} ) \|^2 ]  = \frac{1}{{ P}_{ m}} \sum_{k=0}^{{ K}_{ f }-1} \gamma_{k+1} \EE[ \| \grd V( \hs{k} ) \|^2 ] \eqs,
\eeq
where ${ P}_{ m} \eqdef \sum_{\ell=0}^{{ K}_{ f }-1} \gamma_\ell$ and the expectation is taken over the total randomness of the algorithm.
Denote $\Delta V \eqdef V( \hs{0} ) - V( \hs{{ K}_{ f }})$ and $ \|\Delta S\|^2 \eqdef \| \hs{k} - \stt^{(k)}\|^2$.
The \SAEMVR\ method satisfies:

\begin{theoremcoloured}\label{thm:vrsaem}
Assume A\ref{ass:compact}-A\ref{ass:mcerror}.
Consider the \SAEMVR\ sequence $\{\hat{\bss}^{(k)}\}_{k>0} \in \mathcal{S}$ for any $k \leq { K}_{ f }$ where ${ K}_{ f }$ is a positive integer.
Let $\{\gamma_{k+1} = 1/(k^a \overline{L})\}_{k>0}$, where $a \in (0,1)$, be a sequence of stepsizes, $\overline{L} = \max \{\Lip{\bss}, \Lip{V} \}$, $\rho = \mu/( c_1 \overline{L}  n^{2/3})$, $m = n c_1^2/(2 \mu^2+\mu c_1^2)$ and a constant $\mu \in (0,1)$. Then:
\beq\notag
\begin{split}
 \EE[ \| \grd V( \hs{K} ) \|^2 ] \leq  \frac{2 n^{2/3} \overline{L}}{\mu { P}_{ m} \upsilon_{\min}^2\upsilon_{\max}^2} ( \EE[ \Delta V ]+  \sum_{k=0}^{{ K}_{ f }-1}  \tilde{\eta}^{(k+1)}\hspace{-0.1cm} + \chi^{(k+1)} \EE[ \|\Delta S\|^2]).
\end{split}
\eeq
\end{theoremcoloured}

Furthermore, the \FISAEM\ method displays the following convergence rate:

\begin{theoremcoloured}\label{thm:fisaem}
Assume A\ref{ass:compact}-A\ref{ass:mcerror}.
Consider the \FISAEM\ sequence $\{\hat{\bss}^{(k)}\}_{k>0} \in \mathcal{S}$ for any $k \leq { K}_{ f }$ where ${ K}_{ f }$ be a positive integer.
Let $\{\gamma_{k+1} = 1/(k^a \alpha c_1 \overline{L}) \}_{k>0}$, where $a \in (0,1)$, be a sequence of positive stepsizes, $\alpha =\max\{2, 1+2\upsilon_{\min}\}$, $\overline{L} = \max\{ \Lip{\bss} , \Lip{V} \}$, $\beta = 1/(\alpha n)$, $\rho = 1/(\alpha c_1 \overline{L}n^{2/3})$ and $c_1(k\alpha-1) \geq c_1(\alpha-1) \geq 2$. Then:
\beq\notag
\begin{split}
  \EE[ \| \grd V( \hs{K} ) \|^2 ] \leq \frac{4\alpha  \overline{L} n^{2/3}}{{ P}_{ m}\upsilon_{\min}^2\upsilon_{\max}^2} ( \EE [ \Delta V ]   + \sum_{k=0}^{{ K}_{ f }-1}  \Xi^{(k+1)}  +\Gamma^{(k+1)} \EE [ \|\Delta S\|^2 ]) \eqs.
\end{split}
\eeq
\end{theoremcoloured}
Note that in those two bounds, $\tilde{\eta}^{(k+1)} $ and $ \Xi^{(k+1)} $ depend only on the Monte Carlo noises $\EE [\| \eta_{i_k}^{(k)}\|^2 ]$, $\EE[\| \EE[\eta_{i}^{(r)}|{\cal F}_r]\|^2]$, bounded under assumption A\ref{ass:mcerror}, and some constants.

\vspace{0.2in}

\noindent {Remarks:} Theorem~\ref{thm:vrsaem} and Theorem~\ref{thm:fisaem} exhibit in their convergence bounds {two different phases}.
The upper bounds display a {bias term} due to the initial conditions, \ie the term $ \Delta V$, and a {double dynamic} burden exemplified by the term $\EE [ \|\Delta S\|^2 $.
Indeed, we remark the following: (i) This term is the price we pay for the two-timescale dynamic and corresponds to the gap between the two {asynchronous} updates (one on  $\hs{k}$ and the other on $ \tilde{S}^{(k)}$).
(ii) It is readily understood that if $\rho = 1$, \ie\ there is no variance reduction, then for any $k >0$,
\beq\notag
\EE [ \|\Delta S\|^2] = \EE [\| \StocEstep^{(k+1)} - \stt^{(k+1)} \|^2]= 0  \eqsp,
\eeq
with $\hs{0} = \tilde{S}^{(0)}=0$, which strengthens the fact that this quantity characterizes the impact of the variance reduction technique introduced in our scheme.
The following Lemma describes this gap:

\begin{lemmacoloured} \label{lem:gap_dynamics}
Considering a decreasing stepsize $\gamma_k \in (0,1)$ and a constant $\rho \in (0,1)$, we have
\beq\notag
\begin{split}
\EE [ \|\Delta S\|^2]  \leq \frac{\rho}{1-\rho}\sum_{\ell = 0}^k (1-\gamma_{\ell} )^2 (   \StocEstep^{(\ell)} - \stt^{(\ell)})\eqs,
\end{split}
\eeq
where $\StocEstep^{(\ell)}  $ is defined by Line~2 (\SAEMVR ) or Line~3 (\FISAEM ).
\end{lemmacoloured}

\section{Proof Sketches}\label{sec:proofs}
We provide in the sequel sketches of the proofs of our main Theorems along important auxiliary Lemmas used throughout the proofs.

\subsection{Proof of Theorem~\ref{thm:isaem}}

The main convergence result for the \ISAEM\ algorithm, \ie Theorem~\ref{thm:isaem}, is derived under the control of the Monte Carlo fluctuations as described by assumption A\ref{ass:mcerror} and is built upon the following intermediary Lemma, characterizing the quantity of interest $ \stt^{(k+1)} - \hat{\bss}^{(k)} $ at each iteration index $k > 0$:

\begin{lemmacoloured}\label{lem:meanfield_isaem}
 Assume A\ref{ass:compact}. The \ISAEM\ update, Line~1 of Table~\ref{alg:prox}, is equivalent to the following update on the statistics
 $$\hat{\bss}^{(k+1)} =  \hat{\bss}^{(k)}  + \gamma_{k+1} (  \sum_{i=1}^n \tilde{S}_i^{(\tau_i^k)} - \hat{\bss}^{(k)} ) \eqsp .$$
Also:
\beq\notag
\begin{split}
\EE[\stt^{(k+1)} - \hat{\bss}^{(k)}]= \EE[\overline{\bss}^{(k)} - \hat{\bss}^{(k)}] + (1 - 1/n ) \EE[\frac{1}{n} \sum_{i=1}^n \tilde{S}_i^{(\tau_i^k)}- \overline{\bss}^{(k)}]+\frac{1}{n}\EE[\eta_{i_k}^{(k+1)}]\eqsp,
\end{split}
\eeq
where $\overline{\bss}^{(k)}$ is defined by~\eqref{eq:definition-overline-bss} and $\tau_i^k = \max \{ k' : i_{k'} = i,~k' < k \}$.
\end{lemmacoloured}

\begin{proof}
From update~\eqref{eq:isaem}, we have:
\beq\notag
\begin{split}
\stt^{(k+1)} - \hat{\bss}^{(k)} = \overline{\bss}^{(k)} - \hat{\bss}^{(k)} + \stt^{(k)}- \overline{\bss}^{(k)}  - \frac{1}{n}( \tilde{S}_{i_k}^{(\tau_i^k)} - \tilde{S}_{i_k}^{(k+1)}   ) \eqsp .
\end{split}
\eeq
Since $\tilde{S}_{i_k}^{(k+1)} = \overline{\bss}_{i_k}(\param^{(k+1)}) + \eta_{i_k}^{(k+1)}$ we have
\beq\notag
\begin{split}
\stt^{(k+1)} - \hat{\bss}^{(k)} = \overline{\bss}^{(k)} - \hat{\bss}^{(k)} + \stt^{(k)}- \overline{\bss}^{(k)}  - \frac{1}{n}( \tilde{S}_{i_k}^{(\tau_i^k)} -  \overline{\bss}_{i_k}(\param^{(k+1)})   ) + \frac{1}{n}\eta_{i_k}^{(k+1)}\eqsp .
\end{split}
\eeq
Taking the full expectation of both side of the equation leads to:
\beq\notag
\begin{split}
\EE[\stt^{(k+1)} - \hat{\bss}^{(k)}] = \EE[\overline{\bss}^{(k)} - \hat{\bss}^{(k)}]  + \EE[\frac{1}{n} \sum_{i=1}^n \tilde{S}_i^{(\tau_i^k)}-  \overline{\bss}^{(k)}] -\frac{1}{n} \EE[\EE[ \tilde{S}_i^{(\tau_i^k)}-  \overline{\bss}_{i_k}(\param^{(k+1)})  | \mathcal{F}_{k} ]] + \frac{1}{n} \EE[\eta_{i_k}^{(k+1)}] \eqsp.
\end{split}
\eeq
Since we have $\EE[ \tilde{S}_i^{(\tau_i^k)} | \mathcal{F}_{k} ] =\frac{1}{n} \sum_{i=1}^n \tilde{S}_i^{(\tau_i^k)}$ and $\EE[  \overline{\bss}_{i_k}(\param^{(k)})  | \mathcal{F}_{k} ]= \overline{\bss}^{(k)}$, we conclude the proof.
\end{proof}
We derive the following Lemma which establishes an upper bound of the quantity $\EE [ \|  \stt^{(k+1)} - \hs{k}   \|^2 ]$, another important quantity in order to characterize the convergence of our incremental scheme.
\begin{lemmacoloured}\label{lem:aux2}
For any $k \geq 0$ and consider the \ISAEM\ update in~\eqref{eq:isaem}, it holds that
{\small\beq\notag
\begin{split}
\EE [ \|  \stt^{(k+1)} - \hs{k}   \|^2 ] \leq & 4 \EE[ \|  \os^{(k)} - \hs{k} \|^2 ]
+ \frac{2\Lip{\bss}^2}{n^3} \sum_{i=1}^n \EE[ \| \hs{k} - \hs{t_i^k} \|^2 ]+ 2\frac{c_{\eta}}{M_k} \\
& + 4 \EE[\| \frac{1}{n} \sum_{i=1}^n \tilde{S}_i^{(\tau_i^k)}-  \overline{\bss}^{(k)}\|^2]  \eqsp.
\end{split}
\eeq}
\end{lemmacoloured}
\begin{proof}
Applying the \ISAEM\ update yields:
\beq\notag
\begin{split}
 \EE[ \|  \stt^{(k+1)} - \hs{k} \|^2 ]  =&  \EE[ \| \stt^{(k)} - \hs{k}  -\frac{1}{n}(\tilde{S}^{(\tau_i^k)}_{i_k} - \tilde{S}^{(k)}_{i_k}  )  \|^2 ]\\
 \leq  & 4 \EE[\|\frac{1}{n} \sum_{i=1}^n \tilde{S}_i^{(\tau_i^k)}-  \overline{\bss}^{(k)}\|^2 + 4 \|   \overline{\bss}^{(k)} - \hs{k} \|^2
 +  \frac{2}{n^2}  \| \os_{i_k}^{(k)} - \os_{i_k}^{(t_{i_k}^k)} \|^2] + 2\frac{c_{\eta}}{M_k} \eqsp.
\end{split}
\eeq
The last expectation can be further bounded by
\beq\notag
\begin{split}
&
\frac{2}{n^2}\EE[ \| \os_{i_k}^{(k)} - \os_{i_k}^{(t_{i_k}^k)} \|^2 ] = \frac{2}{n^3} \sum_{i=1}^n \EE[ \| \os_i^{(k)} - \os_i^{(t_i^k)} \|^2 ] \leq \frac{2\Lip{\bss}^2}{n^3}
\sum_{i=1}^n \EE[ \| \hs{k} - \hs{t_i^k} \|^2 ]\eqsp,
\end{split}
\eeq
where the inequality is due to Lemma~\ref{lem:smooth} and which concludes the proof of the Lemma.
\end{proof}

\textbf{Proof of Theorem~\ref{thm:isaem}:} Having established those two auxiliary results, we now give a proof sketch for Theorem~\ref{thm:isaem}.
We consider the \ISAEM\ sequence $\{\hat{\bss}^{(k)}\}_{k>0} \in \mathcal{S}$ obtained with $\rho_{k+1}=1$ via Algorithm~\ref{alg:ttsem} and Line~1 of Table~\ref{alg:prox}.
Under the classical smoothness assumption of the Lyapunov function $V$ (cf. Lemma~\ref{lem:smooth}), Lemma~\ref{lem:meanfield_isaem} yields:
\beq\notag
\begin{split}
 \EE [\pscal{  \stt^{(k+1)}  - \hs{k}}{ \grd V( \hs{k} ) } ] \leq & (\upsilon^2_{\max}\frac{\beta(n-1) + 1}{2n}-\upsilon_{\min}) \EE [\|  \overline{\bss}^{(k)}  - \hs{k}\|^2  ] +  \frac{1}{2 n} \EE [\| \eta_{i_k}^{(k)}\|^2 ]\\
 & + (1 - \frac{1}{n})/(2\beta)\EE[\| \frac{1}{n} \sum_{i=1}^n \tilde{S}_i^{(\tau_i^k)}-  \overline{\bss}^{(k)}\|^2] \eqsp,
\end{split}
\eeq
where the inequality is due to the growth condition~\eqref{lem:growth} and Young's inequality (with $\beta \to 1$).
Besides,
\beq\notag
\frac{1}{n} \sum_{i=1}^n \EE[ \| \hs{k+1} - \hs{t_i^{k+1}} \|^2 ] = \frac{1}{n} \sum_{i=1}^n
( \frac{1}{n} \EE[ \| \hs{k+1} - \hs{k} \|^2 ] + \frac{n-1}{n} \EE[ \| \hs{k+1} - \hs{\tau_i^k} \|^2 ]  )\eqsp,
\eeq
where the equality holds as $i_k$ and $j_k$ are drawn independently. For any $\beta > 0$, it holds
\beq\notag
\begin{split}
 \EE[ \| \hs{k+1} - \hs{t_i^k} \|^2 ] \leq  &\EE [ \| \hs{k+1} - \hs{k} \|^2 + \| \hs{k} - \hs{\tau_i^k} \|^2 +  \frac{\gamma_{k+1}}{\beta} \| \hs{k} - \stt^{(k+1)}\|^2 \\
 &+ \gamma_{k+1} \beta \| \hs{k}- \hs{\tau_i^k} \|^2 ]\eqsp,
\end{split}
\eeq
where the last inequality is due to Young's inequality. Subsequently, we have
\beq\notag
\begin{split}
\frac{1}{n} \sum_{i=1}^n \EE[ \| \hs{k+1} - \hs{\tau_i^{k+1}} \|^2 ]
 \leq & \EE[  \| \hs{k+1} - \hs{k} \|^2 ] + \frac{\gamma_{k+1}}{\beta} \|  \hs{k} - \stt^{(k+1)} \|^2 ]\\
 &+ \frac{n-1}{n^2} \sum_{i=1}^n \EE [ (1+\gamma_{k+1} \beta) \|  \hs{k} - \hs{\tau_i^k} \|^2 \eqsp.
\end{split}
\eeq
Applying Lemma~\ref{lem:aux2} gives
\beq\notag
\begin{split}
 \frac{1}{n} \sum_{i=1}^n \EE[ \| \hs{k+1} - \hs{\tau_i^{k+1}} \|^2 ]  \leq & 4(\gamma_{k+1}^2 +\frac{\gamma_{k+1}}{\beta}  )\EE [  \|   \os^{(k)} - \hs{k}  \|^2  ] + 2(\gamma_{k+1}^2 +\frac{\gamma_{k+1}}{\beta}  )\EE [\| \eta_{i_k}^{(k)}\|^2 ]\\
+&  4 (\gamma_{k+1}^2 +\frac{\gamma_{k+1}}{\beta}  )\EE[\|\frac{1}{n} \sum_{i=1}^n \tilde{S}_i^{(\tau_i^k)}-  \overline{\bss}^{(k)}\|^2] \\
+&  \sum_{i=1}^n \EE [ \frac{1 - \frac{1}{n} + \gamma_{k+1} \beta + \frac{2\gamma_{k+1} \Lip{\bss}^2}{n^2}(\gamma_{k+1} +\frac{1}{\beta})}{n} \|  \hs{k} - \hs{t_i^k} \|^2  ]  \eqsp,
\end{split}
\eeq
and define the following quantity
\beq\notag
\Delta^{(k)} \eqdef \frac{1}{n} \sum_{i=1}^n \EE[ \| \hs{k} - \hs{\tau_i^{k}} \|^2 ]\eqsp.
\eeq
Setting $c_1 = \upsilon_{\min}^{-1}$, $\alpha =\max\{8, 1+6\upsilon_{\min}\}$, $\overline{L} = \max\{ \Lip{\bss} , \Lip{V} \}$, $\gamma_{k+1} = \frac{1}{k \alpha c_1 \overline{L}}$, $\beta = \frac{c_1 \overline{L}}{n}$, we have that $c_1(k\alpha-1) \geq c_1(\alpha-1) \geq 6$ and observe that
\beq\notag
1 - \frac{1}{n} + \gamma_{k+1} \beta + \frac{2\gamma_{k+1} \Lip{\bss}^2}{n^2}(\gamma_{k+1} +\frac{1}{\beta})
 \leq 1 - \frac{c_1(k\alpha  - 1) - 4}{k\alpha n c_1 } \leq 1 - \frac{2}{k\alpha n c_1 }\eqsp,
\eeq
which shows that $1 - \frac{1}{n} + \gamma_{k+1} \beta + \frac{2\gamma_{k+1} \Lip{\bss}^2}{n^2}(\gamma_{k+1} +\frac{1}{\beta})  \in (0,1)$ for any $k >0$.
Denote $ \Lambda_{(k+1)} =\frac{1}{n} - \gamma_{k+1} \beta - \frac{2\gamma_{k+1} \Lip{\bss}^2}{n^2}(\gamma_{k+1} +\frac{1}{\beta}) $ and note that $\Delta^{(0)} = 0$, thus the telescoping sum yields:
\beq\notag
\begin{split}
\Delta^{(k+1)}  \leq & 4 \sum_{ \ell = 0 }^k \prod_{j = \ell +1}^k ( 1 -  \Lambda_{(j)} ) (\gamma_{\ell+1}^2 +\frac{\gamma_{\ell+1}}{\beta}  )  \EE[  \|  \os^{(\ell)} - \hs{\ell}  \|^2 ] \\
  &+ 2\sum_{ \ell = 0 }^k \prod_{j = \ell +1}^k ( 1 -  \Lambda_{(j)} ) (\gamma_{\ell+1}^2  +\frac{\gamma_{\ell+1}}{\beta}  ) \EE [\| \eta_{i_\ell}^{(\ell)}\|^2 ]\\
& +  4 \sum_{ \ell = 0 }^k   \prod_{j = \ell +1}^k ( 1 -  \Lambda_{(j)} )  (\gamma_{\ell+1}^2  +\frac{\gamma_{\ell+1}}{\beta}  )  \EE[\| \frac{1}{n} \sum_{i=1}^n \tilde{S}_i^{(\tau_i^\ell)}-  \overline{\bss}^{(\ell)}\|^2]\eqsp.
\end{split}
\eeq
Note $\omega_{k,\ell} = \prod_{j = \ell +1}^k ( 1 -  \Lambda_{(j)} )$
Summing on both sides over $k=0$ to $k = { K}_{ m }-1$ and upper bounding the quantity $\sum_{k=0}^{{ K}_{ m }-1} \Delta^{(k+1)}$ leads to the combination of the above equations and yields:
{\small \beq\notag
\begin{split}
\sum_{k=0}^{{ K}_{ m }-1}  \tilde{\alpha}_k \EE [\|  \overline{\bss}^{(k)}  - \hs{k}\|^2  ] + \sum_{k=0}^{{ K}_{ m }-1}  \tilde{\beta}_k \EE[\|\frac{1}{n} \sum_{i=1}^n \tilde{S}_i^{(\tau_i^k)}-  \overline{\bss}^{(k)}\|^2]
\leq   \EE [ V( \hs{0} ) - V( \hs{K} ) ]+ \sum_{k=0}^{{ K}_{ m }-1} \tilde{\Gamma}_k         \EE [\| \eta_{i_k}^{(k)}\|^2 ] \eqsp,
\end{split}
\eeq}
where the various quantities are provided in the Appendix for the sake of clarity.
For any $k >0$, $\tilde{\alpha}_k \geq 0$, we have by Lemma~\ref{lem:growth} that:
\beq\notag
\sum_{k=0}^{{ K}_{ m }} \tilde{\alpha}_k \EE [\| \grd V( \hs{k} )\|^2 ] \leq \upsilon_{\max}^2\sum_{k=0}^{{ K}_{ m }} \tilde{\alpha}_k \EE [\|  \overline{\bss}^{(k)}  - \hs{k}\|^2  ]  \eqsp,
\eeq
which yields an upper bound of the gradient of the Lyapunov function $V$ and concludes the proof.

\subsection{Proof of Theorem~\ref{thm:vrsaem}}
We first derive an identity for the drift term of the \SAEMVR\ :

\begin{lemmacoloured}\label{lem:auxvrsaem}
Consider the \SAEMVR\ update~\eqref{eq:vrsaem} with $\rho_k = \rho$, it holds for all $k>0$
\beq\notag
\begin{split}
  \EE [\| \hs{k} - \stt^{(k+1)}\|^2 ] \leq & 2\rho^2 \EE[ \| \hs{k} - \os^{(k)} \|^2] +  2\rho^2\Lip{\bss}^2 \EE[ \| \hs{k} - \hs{\ell(k)} \|^2 ]\\
  & +2(1-\rho)^2 \EE[ \| \hs{(k)} - \stt^{(k)} \|^2 ]+ 2\rho^2\EE[\|\eta_{i_k}^{(k+1)} \|^2]\eqs,
\end{split}
\eeq
where we recall that $\ell(k)$ is the first iteration number in the epoch that iteration $k$ is in.
\end{lemmacoloured}

\begin{proof}
Beforehand, we provide a rewriting of the quantity $ \hs{k+1} - \hs{k} $ that will be useful throughout this proof:
\beq\label{eq:vrsaem_drift_main}
\begin{split}
\hs{k+1} - \hs{k} & =-\gamma_{k+1}  ( \hs{k} - (1-\rho)\stt^{(k)} - \rho\StocEstep^{(k+1)})\\
& = -\gamma_{k+1} ((1-\rho)[\hs{k} - \stt^{(k)} ] +\rho[\hs{k} - \StocEstep^{(k+1)}] ) \eqsp.
\end{split}
\eeq
We observe, using the identity~\eqref{eq:vrsaem_drift_main}, that
\beq \label{eq:auxlemvrsaem_main}
\begin{split}
\EE[ \| \hs{k} -\stt^{(k+1)} \|^2 ] \leq 2\rho^2 \EE[ \| \hs{k} - \os^{(k)} \|^2] + 2\rho^2 \EE[ \| \os^{(k)} - \StocEstep^{(k+1)} \|^2 ] + 2(1-\rho)^2 \EE[ \| \hs{(k)} - \stt^{(k)} \|^2 ].
\end{split}
\eeq
For the latter term, we obtain its upper bound as 
\beq\notag
\begin{split}
& \EE[ \| \os^{(k)} - \StocEstep^{(k+1)} \|^2 ]
= \EE[ \| \frac{1}{n} \sum_{i=1}^n ( \os_i^{(k)} - \tilde{S}_i^{\ell(k)} ) - ( \os_{i_k}^{(k)} - \tilde{S}_{i_k}^{(\ell(k))} ) \|^2 ] \\
 \overset{(a)}{\leq} & \EE[ \| \os_{i_k}^{(k)} - \os_{i_k}^{(\ell(k))} \|^2 ] + \EE[\|\eta_{i_k}^{(k+1)} \|^2] \overset{(b)}{\leq}  \Lip{\bss}^2 \EE[ \| \hs{k} - \hs{\ell(k)} \|^2 ]+ \EE[\|\eta_{i_k}^{(k+1)} \|^2]\eqsp,
\end{split}
\eeq
where $(a)$ uses the variance inequality and $(b)$ uses Lemma~\ref{lem:smooth}.
Substituting into~\eqref{eq:auxlemvrsaem_main} proves the lemma.
\end{proof}

\textbf{Proof of Theorem~\ref{thm:vrsaem}:}
Similar arguments using the smoothness of the Lyapunov function as above are used at the beginning of the following proof.
The main different argument when dealing with two-timescale methods, rather than incremental ones, is in the construction of the following sequence:
\beq \label{eq:seq}
R_k \eqdef \EE[ V( \hs{k} ) + b_{{k}} \| \hs{k} - \hs{\ell(k)} \|^2 ]\eqsp,
\eeq
where for $k >0$, $b_k \eqdef \overline{b}_{k~{\rm mod}~m}$ is a periodic sequence where:
\beq\notag
\overline{b}_i = \overline{b}_{i+1} (1 + \gamma_{k+1} \beta + 2 \gamma_{k+1}^2\rho^2 \Lip{\bss}^2 ) + \gamma_{k+1}^2\rho^2 \Lip{V} \Lip{\bss}^2,~~i=0,1,\dots,m-1~~\text{with}~~\overline{b}_m = 0\eqsp.
\eeq
Note that $\overline{b}_i$ is decreasing with $i$ and this implies
\beq\notag
\overline{b}_i \leq \overline{b}_0 = \gamma_{k+1}^2\rho^2 \Lip{V} \Lip{\bss}^2 \frac{ (1 + \gamma_{k+1} \beta + 2 \gamma_{k+1}^2 \rho^2\Lip{\bss}^2 )^m - 1 }{ \gamma_{k+1} \beta + 2 \gamma_{k+1}^2 \rho^2\Lip{\bss}^2 },~i=1,2,\dots,m \eqsp.
\eeq
For $k+1 \leq \ell(k) + m$, we have the following inequality
\beq\notag
\begin{split}
R_{k+1 }  \leq  &
\EE [ V( \hs{k} ) ] - \gamma_{k+1}(  \rho \upsilon_{\min} +   \upsilon_{\max}^2  - \gamma_{k+1}\rho^2 \Lip{V} - b_{k+1}(\frac{\rho}{\beta}+ 2\gamma_{k+1} \rho^2) ) \EE[ \|  \hmean_{k} \|^2 ] \\
& + (  \underbrace{b_{k+1} (1 + \gamma \beta + 2 \gamma^2\rho^2 \Lip{\bss}^2 ) + \gamma^2\rho^2 \Lip{V} \Lip{\bss}^2}_{=b_k~~\text{since $k+1 \leq \ell(k)+m$}} ) \EE[  \| \hs{k} - \hs{\ell(k)} \|^2 ]+ \tilde{\eta}^{(k+1)} + \tilde{\chi}^{(k+1)}\eqsp,
\end{split}
\eeq
where we have used Lemma~\ref{lem:auxvrsaem}.
Then, using Lemma~\ref{lem:growth}, that for any $\gamma_{k+1}$, $\rho$ and $\beta$ such that $  \rho \upsilon_{\min} +   \upsilon_{\max}^2  - \gamma_{k+1}\rho^2 \Lip{V} - b_{k+1}(\frac{\rho}{\beta}+ 2\gamma_{k+1} \rho^2)  >0$,
\beq\notag
\begin{split}
\upsilon_{\max}^2 \EE[ \| \grd V( \hs{k} ) \|^2 ]  \leq \EE[ \| \hs{k} - \os^{(k)} \|^2 ]
\leq & \frac{  R_{k} - R_{k+1} }{ \gamma_{k+1}(  \rho \upsilon_{\min} +   \upsilon_{\max}^2  - \gamma_{k+1}\rho^2 \Lip{V} - b_{k+1}(\frac{\rho}{\beta}+ 2\gamma_{k+1} \rho^2) )}\\
& +\frac{ \tilde{\eta}^{(k+1)} + \tilde{\chi}^{(k+1)} }{ \gamma_{k+1}(  \rho \upsilon_{\min} +   \upsilon_{\max}^2  - \gamma_{k+1}\rho^2 \Lip{V} - b_{k+1}(\frac{\rho}{\beta}+ 2\gamma_{k+1} \rho^2) )} \eqsp.
\end{split}
\eeq
We first remark that
\beq\notag
\begin{split}
\gamma_{k+1}(  \rho \upsilon_{\min} +   \upsilon_{\max}^2  - \gamma_{k+1}\rho^2 \Lip{V} - b_{k+1}(\frac{\rho}{\beta}+ 2\gamma_{k+1} \rho^2) )
 \geq  \frac{\gamma_{k+1} \rho}{c_1}(1  - \gamma_{k+1}c_1\rho \Lip{V} - b_{k+1}(\frac{c_1}{\beta}+ 2\gamma_{k+1} \rho c_1) )\eqsp,
\end{split}
\eeq
where $c_1 = \upsilon_{\min}^{-1}$.
By setting $\overline{L} = \max \{\Lip{\bss}, \Lip{V} \}$, $\beta = \frac{c_1 \overline{L}}{n^{1/3}}$, $\rho = \frac{\mu}{ c_1 \overline{L}  n^{2/3}}$, $m = \frac{n c_1^2}{2 \mu^2+\mu c_1^2}$ and $\{ \gamma_{k+1}\}$ any sequence of decreasing stepsizes in $(0,1)$, it can be shown that there exists $\mu \in (0,1)$, such that the following lower bound holds
\beq\notag
\begin{split}
 1  - \gamma_{k+1}c_1\rho \Lip{V} - b_{k+1}(\frac{c_1}{\beta}+ 2\gamma_{k+1} \rho c_1)
 \overset{(a)}{\geq} 1 - \frac{\mu}{ n^{\frac{2}{3}}} - \frac{ \mu }{c_1^2 } (\rme-1) ( 1 + \frac{2 \mu}{n} )
 \geq 1 - \mu - \mu(1+2 \mu) \frac{\rme-1}{c_1^2} \overset{(b)}{ \geq} \frac{1}{2}\eqsp,
\end{split}
\eeq
where the simplification in (a) is due to
\beq\notag
\frac{\mu}{n} \leq \gamma \beta + 2 \gamma^2 \Lip{\bss}^2 \leq \frac{\mu}{n} + \frac{2 \mu^2}{c_1^2 n^{\frac{4}{3}}} \leq \frac{\mu c_1^2 + 2 \mu^2}{c_1^2} \frac{1}{n}~~\text{and}~~(1 + \gamma \beta + 2 \gamma^2 \Lip{\bss}^2 )^m \leq \rme-1 \eqs,
\eeq
where the required $\mu$ in (b) can be found by solving the quadratic equation.
Noting that $R_0 = \EE[ V( \hs{0} ) ]$ and if ${ K}_{ m }$ is a multiple of $m$, then $R_{ max} = \EE[ V( \hs{{ K}_{ m }}) ]$, hence concluding our proof.

\subsection{Proof of Theorem~\ref{thm:fisaem}}

We begin with the statement and proofs of two required Lemmas.
First, an equivalent update of Line~3 is given below for the purpose of the proof.

\begin{lemmacoloured} \label{lem:drift_fisaem}
 At iteration $k+1$, the drift term of update~\eqref{eq:fisaem}, with $\rho_{k+1} = \rho$, is equivalent to:
\beq\notag
\begin{split}
 \hs{k} -  \stt^{(k+1)}= \rho (\hs{k} - \overline{\bss}^{(k)})  + \rho \eta_{i_k}^{(k+1)}+ \rho [(\overline{\bss}_{i_k}^{(k)} - \tilde{S}_{i_k}^{(t_{i_k}^k)}) - \EE[\overline{\bss}_{i_k}^{(k)} - \tilde{S}_{i_k}^{(t_{i_k}^k)}] ] + (1-\rho)\left( \hs{k} - \tilde{S}^{(k)}\right),
\end{split}
\eeq
where we recall that $\eta_{i_k}^{(k+1)}$, defined in~\eqref{eq:mcerror}, which is the gap between the MC approximation and the expected statistics.
\end{lemmacoloured}
\begin{proof}
Using the \FISAEM\ update $ \stt^{(k+1)} = (1 - \rho)\stt^{(k)} + \rho \StocEstep^{(k+1)}$ where $\StocEstep^{(k+1)} = \overline{\StocEstep}^{(k)} + ( \tilde{S}_{i_k}^{(k)}  - \tilde{S}_{i_k}^{(t_{i_k}^k)} )$ leads to the following decomposition:
\beq\notag
\begin{split}
 \stt^{(k+1)} - \hs{k}  = \rho (\overline{\bss}^{(k)}-\hs{k}) + \rho \eta_{i_k}^{(k+1)} - \rho [(\overline{\bss}_{i_k}^{(k)} - \tilde{S}_{i_k}^{(t_{i_k}^k)}) - \EE[\overline{\bss}_{i_k}^{(k)} - \tilde{S}_{i_k}^{(t_{i_k}^k)}] ]  + (1-\rho)\left(\stt^{(k)} - \hs{k}\right) \eqsp,
\end{split}
\eeq
where we observe that $\EE[\overline{\bss}_{i_k}^{(k)} - \tilde{S}_{i_k}^{(t_{i_k}^k)}] =\overline{\bss}^{(k)} -   \overline{\StocEstep}^{(k)} $ and which concludes the proof.

\textit{Important Note:} Note that $\overline{\bss}_{i_k}^{(k)} - \tilde{S}_{i_k}^{(t_{i_k}^k)}$ is not equal to $\eta_{i_k}^{(k+1)}$, defined in~\eqref{eq:mcerror}, which is the gap between the MC approximation and the expected statistics, $\tilde{S}_{i_k}^{(t_{i_k}^k)}$ is not computed under the same model as $\overline{\bss}_{i_k}^{(k)}$.
\end{proof}
Then, we derive an identity for the quantity $\EE[ \| \hs{k} - \stt^{(k+1)}   \|^2 ]$:
\begin{lemmacoloured}\label{lem:aux1}
Consider the \FISAEM\ update~\eqref{eq:fisaem} with $\rho_k = \rho$. It holds for all $k>0$ that
\beq\notag
\begin{split}
  \EE [\| \hs{k} - \stt^{(k+1)}\|^2 ] \leq & 2\rho^2 \EE[ \| \hs{k} - \os^{(k)} \|^2] +  2\rho^2\frac{\Lip{\bss}^2}{n}
\sum_{i=1}^n \EE[ \| \hs{k} - \hs{t_i^k} \|^2 ]\\
& +2(1-\rho)^2 \EE[ \| \hs{(k)} - \stt^{(k)} \|^2 ]+ 2\rho^2\EE[\|\eta_{i_k}^{(k+1)} \|^2]\eqsp,
\end{split}
\eeq
where $\Lip{\bss}$ is the smoothness constant defined in Lemma~\ref{lem:smooth}.
\end{lemmacoloured}
\begin{proof}
Beforehand, we provide a rewriting of the quantity $ \hs{k+1} - \hs{k}$ as follows:
\beq\label{eq:fisaem_drift_main}
\begin{split}
\hs{k+1} - \hs{k}  =  -\gamma_{k+1} ((1-\rho)[\hs{k} - \stt^{(k)} ] +\rho[\hs{k} - \overline{\StocEstep}^{(k)} - ( \tilde{S}_{i_k}^{(k)}  -  \tilde{S}_{i_k}^{(t_{i_k}^k)}  )] ) \eqsp.
\end{split}
\eeq
We observe, using the identity~\eqref{eq:fisaem_drift_main}, that
\beq \label{eq:auxlemfisaem_main}
\begin{split}
\EE[ \| \hs{k} -\stt^{(k+1)} \|^2 ] \leq 2\rho^2 \EE[ \| \hs{k} - \os^{(k)} \|^2] + 2\rho^2 \EE[ \| \os^{(k)} - \StocEstep^{(k+1)} \|^2 ]+ 2(1-\rho)^2 \EE[ \| \hs{(k)} - \stt^{(k)} \|^2 ]\eqsp.
\end{split}
\eeq
For the latter term, we obtain its upper bound as 
\beq
\begin{split}
\EE[ \| \os^{(k)} - \StocEstep^{(k+1)} \|^2 ]  = \EE [ \| \frac{1}{n} \sum_{i=1}^n ( \os_i^{(k)} -\overline{\StocEstep}_i^{(k)} ) - ( \tilde{S}_{i_k}^{(k)} - \tilde{S}_{i_k}^{(t_{i_k}^k)} ) \|^2 ] \overset{(a)}{\leq} \EE[ \| \os_{i_k}^{(k)} - \os_{i_k}^{(\ell(k))} \|^2 ] + \EE[\|\eta_{i_k}^{(k+1)} \|^2] \eqsp,
\end{split}
\eeq
where $(a)$ uses the variance inequality.
We can further bound the last expectation using Lemma~\ref{lem:smooth}:
\beq\notag
\EE[ \| \os_{i_k}^{(k)} - \os_{i_k}^{(t_{i_k}^k)} \|^2 ] = \frac{1}{n} \sum_{i=1}^n \EE[ \| \os_i^{(k)} - \os_i^{(t_i^k)} \|^2 ] \overset{(a)}{\leq} \frac{\Lip{\bss}^2}{n}
\sum_{i=1}^n \EE[ \| \hs{k} - \hs{t_i^k} \|^2 ]\eqsp.
\eeq
Substituting into~\eqref{eq:auxlemfisaem_main} proves the lemma.
\end{proof}

\textbf{Proof of Theorem~\ref{thm:fisaem}:}
Using the smoothness of $V$ and update~\eqref{eq:fisaem}, we obtain:
\beq\label{eq:smoothfisaem_main}
\begin{split}
V( \hs{k+1} )  \leq V( \hs{k} ) - \gamma_{k+1} \pscal{  \hs{k} - \stt^{(k+1)} }{ \grd V( \hs{k} ) } + \frac{\gamma_{k+1}^2 \Lip{V}}{2} \| \hs{k}  -  \stt^{(k+1)}\|^2\eqsp.
\end{split}
\eeq
Denote $\Hdrift_{k+1} \eqdef   \hs{k} - \stt^{(k+1)} $ the drift term of the \FISAEM\ update in~\eqref{eq:rmstep} and  $\hmean_{k} = \hs{k} - \overline{\bss}^{(k)}$. Using Lemma~\ref{lem:drift_fisaem} and the additional following identity $\EE[(\overline{\bss}_{i_k}^{(k)} - \tilde{S}_{i_k}^{(t_{i_k}^k)}) - \EE[\overline{\bss}_{i_k}^{(k)} - \tilde{S}_{i_k}^{(t_{i_k}^k)}] ] = 0$,  we have
 \beq\notag
\begin{split}
 \EE[V( \hs{k+1} )]
 \leq &  -(\upsilon_{\min}\gamma_{k+1}\rho+\gamma_{k+1} \upsilon_{\max}^2) \EE[\norm{\hmean_{k}}^2 ] -\frac{\gamma_{k+1}\rho^2}{2} \xi^{(k+1)} \\
 &- \frac{\gamma_{k+1}(1-\rho)^2}{2} \EE[\| \hs{k} - \tilde{S}^{(k)}\|^2]
 + \frac{\gamma_{k+1}^2 \Lip{V}}{2} \| \Hdrift_{k+1}\|^2\eqsp,
\end{split}
\eeq
where $\xi^{(k+1)}  =\EE[\|\EE[\eta_{i_k}^{(k+1)}|{\cal F}_k]  \|^2 ]$.
The remaining of the proof is similar to the one for the \SAEMVR\ algorithm.
It consists of bounding the terms (i) $\EE[\|  \Hdrift_{k+1}  \|^2]$ and (ii) $\EE[ \| \hs{k+1} - \hs{t_i^k} \|^2 ]$ using respectively Lemma~\ref{lem:aux1} and noting that $\hs{k+1} - \hs{k} = -\gamma_{k+1} ( \hs{k} - \stt^{(k+1)}) = -\gamma_{k+1} \Hdrift_{k+1}$.
We also recall that in expectation $\EE[\Hdrift_{k+1}|{\cal F}_k] =  \rho \hmean_{k} + \rho\EE[\eta_{i_k}^{(k+1)}|{\cal F}_k] + (1-\rho) \EE[\stt^{(k)} - \hs{k}]$ where $\hmean_{k} = \hs{k} - \overline{\bss}^{(k)}$.
As for the \ISAEM\ method, an important step of our proof is to define the following quantity
\beq\notag
\Delta^{(k)} \eqdef \frac{1}{n} \sum_{i=1}^n \EE[ \| \hs{k} - \hs{t_i^{k}} \|^2 ]\eqsp,
\eeq
where we recall that $t_j^k = \{ k' : j_{k'} = j , k' < k \}$ is the iteration index where the sample $j \in \inter$ is last drawn as $j_k$ prior to iteration $k$ in addition to $\tau_i^k$ which was defined w.r.t. $i_k$, since \FISAEM\ update in Line~3 requires two independently drawn indices.
Then, from the bounds on (i) and (ii), we obtain
\beq\notag
\begin{split}
 \Delta^{(k+1)} \leq & \left( 1 - \frac{1}{n} +\gamma_{k+1}\beta+\gamma_{k+1}^2\rho^2 \Lip{\bss}^2\right) \Delta^{(k)} + \gamma_{k+1} (1-\rho)^2 \left( 2\gamma_{k+1} + \frac{1}{\beta} \right)\EE[ \|\hs{k} - \tilde{S}^{(k)}\|^2]  \\
 & + \left(2 \gamma_{k+1}^2 \rho^2 + \frac{\gamma_{k+1} \rho^2}{\beta}\right) \EE[\| \overline{\bss}^{(k)}-\hs{k}\|^2 ]+ \gamma_{k+1}\left(2 \gamma_{k+1} + \frac{ \rho^2}{\beta} \right)\EE[\norm{\eta_{i_k}^{(k+1)}}^2 ]\eqsp.
 \end{split}
\eeq
Setting $c_1 = \upsilon_{\min}^{-1}$, $\alpha =\max\{2, 1+2\upsilon_{\min}\}$, $\overline{L} = \max\{ \Lip{\bss} , \Lip{V} \}$, $\gamma_{k+1} = \frac{1}{k }$, $\beta = \frac{1}{\alpha n}$, $\rho = \frac{1}{\alpha c_1 \overline{L}n^{2/3}}$, then we have that $c_1(k\alpha-1) \geq c_1(\alpha-1) \geq 2$.
Hence, we observe
{\small
\beq\notag
1 - \frac{1}{n} +\gamma_{k+1}\beta+\gamma_{k+1}^2\rho^2 \Lip{\bss}^2
 \leq 1 - \frac{1}{n} + \frac{1}{\alpha kn} + \frac{ 1 }{ \alpha^2 c_1^2 k^2 n^{\frac{4}{3}} } \leq 1 - \frac{c_1(k\alpha  - 1) - 1}{k\alpha n c_1 } \leq 1 - \frac{1}{k\alpha n c_1 }
\eeq
}
showing that $1 - \frac{1}{n} +\gamma_{k+1}\beta+\gamma_{k+1}^2\rho^2 \Lip{\bss}^2  \in (0,1)$ for any $k >0$.
Denote $ \Lambda_{(k+1)} =\frac{1}{n} -\gamma_{k+1}\beta-\gamma_{k+1}^2\rho^2 \Lip{\bss}^2 $ and note that $\Delta^{(0)} = 0$, thus the telescoping sum yields:
{\small
\beq\notag
\begin{split}
\Delta^{(k+1)} \leq & \sum_{ \ell = 0 }^k \omega_{k, \ell} \left(2 \gamma_{\ell+1}^2 \rho^2 + \frac{\gamma_{\ell+1}^2 \rho^2}{\beta}\right)  \EE[\norm{\overline{\bss}^{(\ell)}-\hs{\ell}}^2 ]\\
& +\sum_{ \ell = 0 }^k \omega_{k, \ell} \gamma_{\ell+1} (1-\rho)^2 \left( 2\gamma_{\ell+1} +\frac{1}{\beta} \right)\EE[ \norm{\tilde{S}^{(\ell)} - \hs{\ell}}^2] + \sum_{ \ell = 0 }^k \omega_{k, \ell}\gamma_{\ell+1} \tilde{\epsilon}^{(\ell+1)}  \eqsp,
\end{split}
\eeq
}
where $ \omega_{k, \ell} =  \prod_{j = \ell +1}^k ( 1 -  \Lambda_{(j)} )$ and $\tilde{\epsilon}^{(\ell+1)}   = \left(2 \gamma_{k+1} + \frac{ \rho^2}{\beta} \right)\EE[\norm{\eta_{i_k}^{(k+1)}}^2 ]$.
Summing over the total number of iterations, making assumptions on the different hyperparameters of the algorithm and injecting in the smoothness inequality~\eqref{eq:smoothfisaem_main} leads to similar final steps of the proofs as for the two other methods detailed above.
For completeness, we refer readers to Appendix  where our proofs are explained in greater detail.

\section{Numerical Applications}\label{sec:numerical}
This section presents several numerical applications for our proposed class of Algorithms~\ref{alg:ttsem}.
The broad range of potential applications for our scheme include Gaussian Mixture Modeling, deformable template image analysis and nonlinear mixed-effects modeling.
For each example, we provide the formulation of the model, explicit the updates for the various training methods, including the baselines, and run numerical experiments along with visual plots showing the benefits of our proposed methods.

\subsection{Gaussian Mixture Models}
We begin by a simple and illustrative example.
The authors acknowledge that the following model can be trained using deterministic EM-type of algorithms but propose to apply stochastic methods, including theirs, in order to compare their performances.
Given $n$ observations $\{y_i\}_{i=1}^n$, the goal here is to fit a Gaussian Mixture Model (GMM)~\citep{xuan2001algorithms} whose distribution is modeled as a mixture of $M$ Gaussian components, each with a unit variance.
Let $z_i \in \inter[M]$ be the latent labels of each component, the complete log-likelihood is defined as follows:
\beq \notag \textstyle
 \log f( z_i, y_i; \param) =
\sum_{m=1}^{M} \indiacc{m}(z_i) [ \log(\omega_m) - \mu_m^2/2 ] + \sum_{m=1}^M \indiacc{m}(z_i) \mu_m y_i + {\rm cst.} \eqsp,
\eeq
where $\param \eqdef (\bomega, \bmu)$ with $\bomega= \{\omega_{m}\}_{m=1}^{M-1}$ are the mixing weights with the convention $\omega_M= 1 - \sum_{m=1}^{M-1} \omega_m$  and $\bmu= \{\mu_m \}_{m =1}^M$ are the means.
We use the penalization $\Pen(\param)= \frac{\delta}{2}\sum_{m=1}^M \mu_m^2 - \log \Dir(\bomega; M, \epsilon)$ where $\delta > 0$ and $\Dir(\cdot; M,\epsilon)$ is the $M$ dimensional symmetric Dirichlet distribution with concentration parameter $\epsilon > 0$.
The constraint set is given by $\Param = \{ \omega_m,~m=1,...,M-1 : \omega_m \geq 0,~\sum_{m=1}^{M-1} \omega_m \leq 1\} \times \{ \mu_m \in \rset ,~m=1,...,M \}$.

\vspace{0.1in}
\noindent \textbf{EM updates:}
We first recognize that the constraint set for $\param$ is given by $\Param = \Delta^M \times \rset^M$.
Using the partition of the sufficient statistics as
$$S( y_i,z_i ) = ( S^{(1)}( y_i,z_i)^\top , S^{(2)}( y_i,z_i )^\top, S^{(3)}(y_i,z_i) )^\top  \in \rset^{M-1} \times \rset^{M-1} \times \rset$$ the partition $\phi( \param ) = ( \phi^{(1)}( \param )^\top ,\phi^{(2)}( \param )^\top,\phi^{(3)}( \param ) )^\top \in \rset^{M-1} \times \rset^{M-1} \times \rset$ and the fact that $\indiacc{M}(z_i)= 1 - \sum_{m=1}^{M-1} \indiacc{m}(z_i)$, the complete data log-likelihood can be expressed as in~\eqref{eq:exp} with
\beq \label{eq:gmm_exp}
\begin{split}
& s_{i,m}^{(1)} = \indiacc{m}(z_i), \quad \phi_m^{(1)}(\param) =   \left\{\log(\omega_m) -\frac{\mu_m^2}{2}\right\} - \left\{\log(1 - {\textstyle  \sum_{j=1}^{M-1}} \omega_j) - \frac{\mu_M^2}{2}\right\} \eqsp,\\
& s_{i,m}^{(2)} =   \indiacc{m}(z_i) y_i, \quad \phi^{(2)}_m(\param) =  {\mu_m} \eqsp, \quad s_i^{(3)} = y_i, \quad \phi^{(3)}(\param) = \mu_M \eqsp,
\end{split}
\eeq
and $\psi(\param) =   - \left\{\log(1 - \sum_{m=1}^{M-1} \omega_m) - \frac{\mu_M^2}{2 \sigma^2}\right\}$.
We also define for each $m \in [M]$,  $j \in \{1, 2, 3 \}$, $s_{m}^{(j)} = n^{-1}\sum_{i=1}^n s_{i,m}^{(j)}$.
Consider the following latent sample used to compute an approximation of the conditional expected value $\EE_{\param}[ 1_{\{z_i=m\}} | y= y_{i} ]$:
\beq \label{eq:cexp}
z_{i,m} \sim \prob \left(z_i = m |y_i; \param\right) \eqsp,
\eeq
where $m \in [M]$, $i \in \inter$ and $\param = ({\bm w}, {\bm{\mu}}) \in \Param$.
In particular, given iteration $k+1$, the computation of the approximated quantity $ \tilde{S}_{i_k}^{(k)}$ during the { Inc-step} updates, see~\eqref{eq:sestep}, can be written as
\beq\label{eq:stat_gmm}
 \tilde{S}_{i_k}^{(k)} = ( \underbrace{ \{ \indiacc{m}(z_{i_k,m})\}_{m \in [M-1]}}_{\eqdef \tilde{s}_{i_k}^{(1)}} , \underbrace{ \{\indiacc{m}(z_{i_k,m})y_{i_k}\}_{m \in [M-1]}}_{\eqdef \tilde{s}_{i_k}^{(2)}},   , \underbrace{y_{i_k}}_{\eqdef \overline{\bss}_{i_k}^{(3)}( \param^{(k)} )} )^\top.
\eeq
Recall the regularizer $ \frac{\delta}{2}\sum_{m=1}^M \mu_m^2 - \log \Dir(\bomega; M, \epsilon)$ we used, which also reads:
\beq \textstyle \label{eq:regu}
\Pen( \param ) = \frac{\delta}{2} \sum_{m=1}^M \mu_m^2 - \epsilon \sum_{m=1}^M  \log ( \omega_m )  - \epsilon \log ( 1 - \sum_{m=1}^{M-1} \omega_m ) \eqsp.
\eeq
It can be shown that the regularized { M-step} evaluates to
\beq \label{eq:mstep_gmm}
\overline{\param} ( {\bm s} )
= \left(
\begin{array}{c}
( 1+\epsilon M )^{-1} ( {s}_1^{(1)} + \epsilon, \dots,  {s}_{M-1}^{(1)} + \epsilon )^\top \vspace{.2cm}\\
 ( ({s}_1^{(1)} + \delta )^{-1} {s}_1^{(2)}  , \dots, ({s}_{M-1}^{(1)} + \delta )^{-1} {s}_{M-1}^{(2)}  )^\top \vspace{.2cm} \\
  (1 - \sum_{m=1}^{M-1}s_m^{(1)} +  \delta)^{-1} ( s^{(3)} - \sum_{m=1}^{M-1} s_m^{(2)} )
\end{array}
\right)
= \left(
\begin{array}{c}
\overline{\bm{\omega}} ( {\bm s}) \\
\overline{\bm{\mu}} ( {\bm s}) \\
\overline{\mu}_M ( {\bm s})
\end{array}
\right) \eqsp,
\eeq
where we have defined for all $m \in [M]$ and $j \in \{1, 2, 3\}$ , $ {s}_m^{(j)}  = n^{-1} \sum\nolimits_{i=1}^n s_{i,m}^{(j)}$.

\vspace{0.1in}
\noindent \textbf{Synthetic data experiment:}
In the following experiments on synthetic data, we generate $50$ synthetic datasets of size $n = 10^5$ from a GMM model with $M=2$ components of means $\mu_1 = - \mu_2 = 0.5$.
\begin{figure}[t]
\begin{center}
\includegraphics[width=4in]{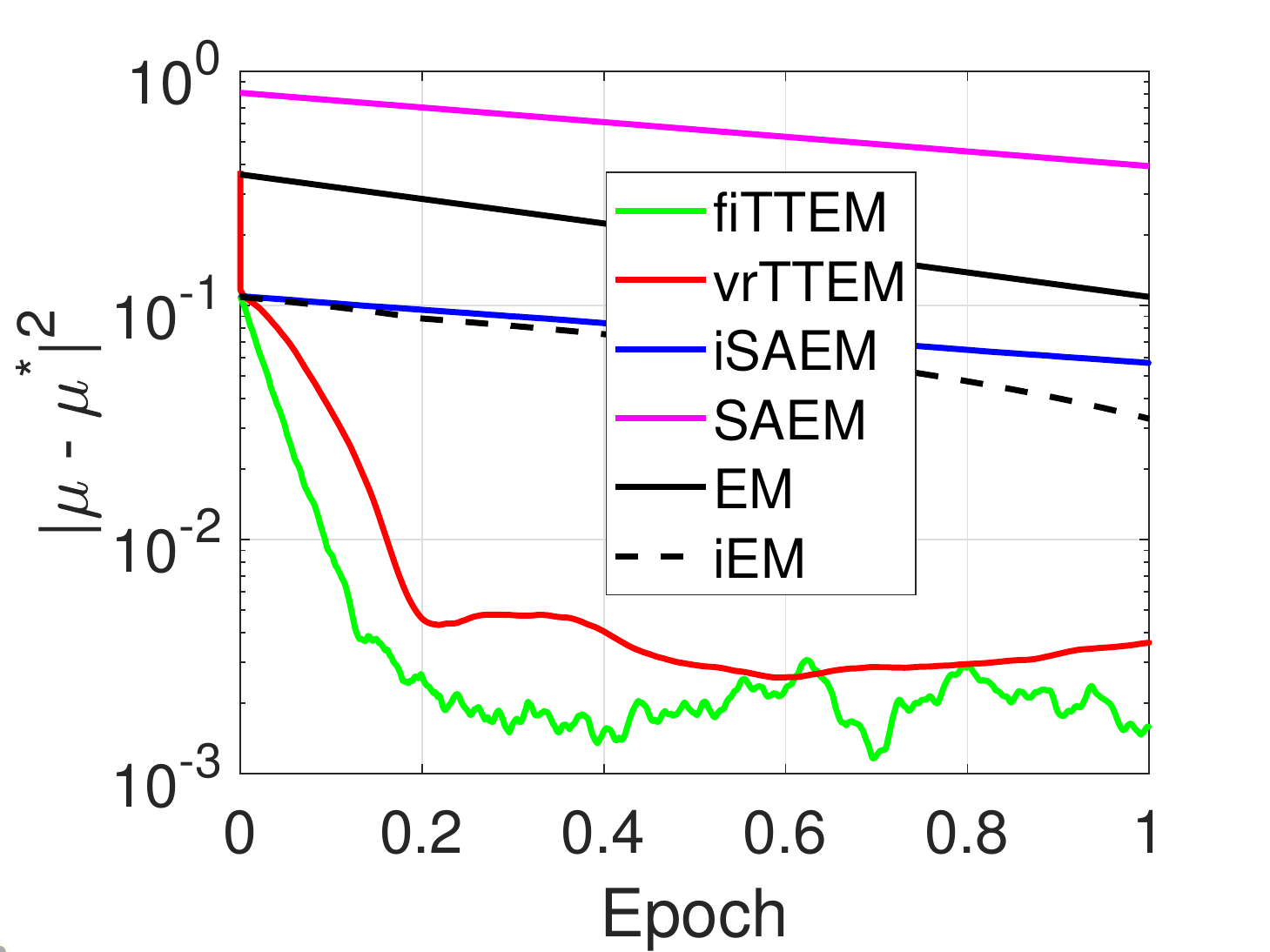}
\end{center}
\caption{Precision $|\mu^{(k)} - \mu^*|^2$ for our methods (\FISAEM\ in green,\SAEMVR\ in black and \ISAEM\ in red) versus deterministic baselines (EM in dashed blue line, iEM in solid blue line) or stochastic baseline (SAEM in solid red line) against epochs elapsed. Our two variance reduced methods, \ie \FISAEM\ and \SAEMVR\ are reaching the highest accuracy.}
\label{fig:gmm_tts}
\end{figure}
We run the EM method until convergence (to double precision) to obtain the ML estimate $\mu^\star$ averaged on $50$ datasets.
We compare the EM, iEM (incremental EM), SAEM, \ISAEM, \SAEMVR\ and \FISAEM\ methods in terms of their precision measured by $| \mu - \mu^\star |^2$.
We set the stepsize of the {SA-step} for all method as $\gamma_k = 1/k^{\alpha}$ with $\alpha = 0.5$, and the stepsize $\rho_k$ for the \SAEMVR\ and the \FISAEM\ to a constant stepsize equal to $1/n^{2/3}$.
The number of MC samples is fixed to $M=10$.
Figure~\ref{fig:gmm_tts} shows the precision $|\mu - \mu^*|^2$ for the different methods through the epoch(s) (one epoch equals $n$ iterations).
The \SAEMVR\ and \FISAEM\ methods outperform the other stochastic methods, supporting the benefits of our scheme.

\vspace{0.1in}
\noindent \textbf{Model Assumptions:}
We use the GMM example to illustrate the required assumptions.
Many practical models can satisfy the compactness of the sets as in assumption A\ref{ass:compact}.
For instance, the GMM example satisfies the conditions in~A\ref{ass:compact} as the sufficient statistics are composed of indicator functions and observations as defined in~\eqref{eq:gmm_exp}.
Assumptions A\ref{ass:expected} and A\ref{ass:reg} are standard for the curved exponential family models.
For GMM, the following (strongly convex) regularization $\Pen( \param )$ ensures A\ref{ass:reg}:
$$
\Pen( \param ) = \frac{\delta}{2} \sum_{m=1}^M \mu_m^2 - \epsilon \sum_{m=1}^M  \log ( \omega_m )  - \epsilon \log ( 1 - \sum_{m=1}^{M-1} \omega_m ) \eqsp,
$$
since it ensures $\param^{(k)}$ is unique and lies in ${\rm int}( \Delta^M ) \times \rset^M$.
We remark that for A\ref{ass:expected}, it is possible to define the Lipschitz constant $\Lip{p}$ independently for each data $y_i$ to yield a refined characterization.
Again, A\ref{ass:eigen} is satisfied by practical models. For GMM, it can be verified by deriving the closed form expression for $\operatorname{B}( \bss )$ and using A\ref{ass:compact}.
Under A\ref{ass:compact} and A\ref{ass:reg}, we have $\| \hat{\bm s}^{(k)} \| < \infty$ since $\Sset$ is compact and $\hat{\param}^{(k)} \in {\rm int}( \Param )$ for any $k \geq 0$ which thus ensure that the EM methods operate in a closed set throughout the optimization.

\newpage


\noindent \textbf{Algorithms updates:}
In the sequel, recall that, for all $i \in \inter[n]$ and iteration $k$, the computed statistic $ \tilde{S}_{i_k}^{(k)}$ is defined by~\eqref{eq:stat_gmm}.
At iteration $k$, the several E-steps defined by~\eqref{eq:isaem} or~\eqref{eq:vrsaem} and~\eqref{eq:fisaem} leads to the definition of the quantity $\hat{\bss}^{(k+1)} $.
Define the exact conditional expected value $\EE_{\param}[ 1_{\{z_i=m\}} | y= y_{i} ]$ as follows:
\beq \notag
\widetilde{\omega}_m ( y_{i} ; \param ) \eqdef \EE_{\param}[ 1_{\{z_i=m\}} | y= y_{i} ]
= \frac{ {\omega}_{m} \!~ {\rm exp}(-\frac{1}{2}( y_{i} - {\mu}_{i} )^2) }{  \sum_{j=1}^{M}{ {\omega}_{j} \!~ \exp(-\frac{1}{2}( y_{i} - {\mu}_{j} )^2)} } \eqsp.
\eeq
Then, for the GMM example, after the initialization of the quantity $\hat{\bss}^{(0)} = n^{-1} \sum\nolimits_{i=1}^n \overline{\bss}_i^{(0)} $, the E-step explicit updates are listed Table~\ref{alg:estepgmm}.

 \begin{protocol}[t]
  \floatname{algorithm}{Table}
\caption{Algorithms Updates for GMM}\label{alg:estepgmm}
  \begin{algorithmic}[1]
\STATE {Batch EM (EM)} \hspace{0.4cm} for all $i \in \inter[n]$, compute $\overline{\bss}_{i}^{(k)}$ and set $\hat{\bss}^{(k+1)} = n^{-1} \sum\nolimits_{i=1}^n \overline{\bss}_i^{(k)}$
\STATE {Incremental EM (iEM)} \hspace{0.4cm} draw  $i_k$ uniformly at random on $\inter[n]$, compute $\overline{\bss}_{i_k}^{(k)}$ and set $\hat{\bss}^{(k+1)} = n^{-1} \sum\nolimits_{i=1}^n \overline{\bss}_i^{(k)}$
\STATE {Batch SAEM (SAEM)} \hspace{0.4cm} for all $i \in \inter$ compute $ \tilde{S}_{i}^{(k)}$~\eqref{eq:stat_gmm} and set  $\hat{\bss}^{(k+1)} = \hat{\bss}^{(k)}(1 - \gamma_{k+1}) + \gamma_{k+1}\stt^{(k)}$
\STATE {Variance Reduced Two-Timescale EM (\SAEMVR)} \hspace{0.4cm} draw  $i_k$ uniformly at random on $\inter[n]$, compute $ \tilde{S}_{i_k}^{(k)}$ via~\eqref{eq:stat_gmm} and set $\hat{\bss}^{(k+1)} = \hat{\bss}^{(k)}(1 - \gamma_{k+1})+ \gamma_{k+1} (\stt^{(k)} (1 - \rho) + \rho (\tilde{S}^{(\ell(k))} +  ( \tilde{S}_{i_k}^{(k)}  -\tilde{S}_{i_k}^{(\ell(k))}   )) ) $
\STATE {Fast Incremental Two-Timescale EM (\FISAEM)} \hspace{0.4cm} draw  $i_k$ uniformly at random on $\inter[n]$, compute $ \tilde{S}_{i_k}^{(k)}$ via~\eqref{eq:stat_gmm} and set $ \hat{\bss}^{(k+1)} = \hat{\bss}^{(k)}(1 - \gamma_{k+1})+ \gamma_{k+1} (\stt^{(k)} (1 - \rho) + \rho (\overline{\StocEstep}^{(k)} + ( \tilde{S}_{i_k}^{(k)}  - \tilde{S}_{i_k}^{(t_{i_k}^k)}) ) \eqsp.$
  \end{algorithmic}
\end{protocol}

Finally, the $k$-th update reads $\hp{k+1} = \overline{\param} (\hat{\bss}^{(k+1)})$ where the function ${\bm s} \to \overline{\param}({\bm s})$ is defined by~\eqref{eq:mstep_gmm}.

\subsection{Deformable Template Model for Image Analysis}

\noindent \textbf{Model and EM Updates:} Let $(y_i, i \in \inter)$ be observed gray level images defined on a grid of pixels.
Let $u \in \mathcal{U} \subset \rset^2$ denote the pixel index on the image and $x_u \in \mathcal{D} \subset \rset^2$ its location.
The model used in this experiment suggests that each image $y_i$ is a deformation of a template, noted $I: \mathcal{D} \to \rset$, common to all images of the dataset:
\beq\label{eq:deformablemodel}
y_{i}(u)=I\left(x_{u}-\Phi_{i}\left(x_{u}, z_i\right)\right)+\varepsilon_{i}(u) \eqsp,
\eeq
where $\Phi_i: \rset^2 \to \rset^2$ is a deformation function, $z_i$ some latent variable parameterizing this deformation and $\varepsilon_{i} \sim \mathcal{N}(0,\sigma^2)$ is an observation error.
The template model, given $\{p_k\}_{k=1}^{k_p}$ landmarks on the template, a fixed known kernel $\mathbf{K}_{\mathbf{p}}$ and a vector of parameters $\beta \in \rset^{k_p}$ is defined as follows:
\beq\notag\label{eq:template}
I_{\xi}=\mathbf{K}_{\mathbf{p}} \beta, \quad \textrm{where} \quad \left(\mathbf{K}_{\mathbf{p}} \beta \right)(x)=\sum_{k=1}^{k_{p}} \mathbf{K}_{\mathbf{p}}\left(x, p_{k}\right) \beta_k\eqs.
\eeq
Given a set of landmarks $\{g_k\}_{k=1}^{k_g}$ and a fixed kernel $\mathbf{K}_{\mathbf{g}}$, we parameterize the deformation $\Phi_{i}$ as:
\beq\notag
\begin{split}
\Phi_{i}=\mathbf{K}_{\mathbf{g}} z_{i} \quad \textrm{where} \quad \left(\mathbf{K}_{\mathbf{g}} z_{i}\right)(x)=\sum_{k=1}^{k_{s}} \mathbf{K}_{\mathbf{g}}\left(x, g_{k}\right)\left(z_{i}^{(1)}(k), z_{i}^{(2)}(k)\right)\eqs,
\end{split}
\eeq
where we put a Gaussian prior on the latent variables, $z_i \sim \mathcal{N}(0,\Gamma)$ and $z_i \in \left( \rset^{k_g}\right)^2$.
Hence, the vector of parameters we want to estimate is $\param = ( \beta, \Gamma, \sigma  )$.

The complete model belongs to the curved exponential family, see~\citet{allassonniere2007towards}, and its vector of sufficient statistics, noted $S = (S_1(z),S_2(z),S_3(z) )$, reads:
\beq \label{eq:suffstat_deformable2}
 S_1(z) = \frac{1}{n} \sum_{i=1}^n \left(\mathbf{K}_{p}^{z_{i}}\right)^\top y_{i},\quad S_2(z) =\frac{1}{n} \sum_{i=1}^n \left(\mathbf{K}_{p}^{z_{i}}\right)^\top\left(\mathbf{K}_{p}^{z_{i}}\right),\quad S_3(z) = \frac{1}{n}  \sum_{i=1}^n  z_{i}^{t} z_{i} \eqsp,
\eeq
where for any pixel $u \in \rset^2$ and $j \in [k_g]$ we denote:
\beq\notag
\mathbf{K}_{p}^{z_{i}}(x_u,j) = \mathbf{K}_{p}^{z_{i}}(x_u - \phi_i(x_u,z_i), p_j)\eqsp.
\eeq
Finally, the Two-Timescale {M-step} yields the following parameter updates:
\beq\notag
\bar{\param}(\hat{s})
= \left(
\begin{array}{c}
\beta(\hat{s}) =   \hat{s}_2^{-1}(z) \hat{s}_1(z)    \\
\Gamma(\hat{s}) = \frac{1}{n} \hat{s}_3(z)   \\
 \sigma(\hat{s}) =\beta(\hat{s})^\top  \hat{s}_2(z) \beta(\hat{s}) - 2\beta(\hat{s}) \hat{s}_1(z)
\end{array}
\right)\eqsp,
\eeq
where $\hat{s} = (\hat{s}_1(z),\hat{s}_2(z),\hat{s}_3(z))$ is the vector of statistics obtained via the {SA-step}~\eqref{eq:rmstep} and using the MC approximation of the sufficient statistics $(S_1(z),S_2(z),S_3(z) )$ defined in~\eqref{eq:suffstat_deformable2}.

\vspace{0.2in}

\begin{figure}[h]
\includegraphics[width=\textwidth]{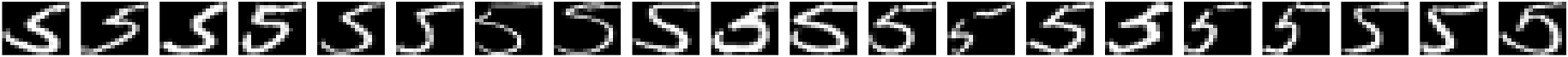}
\caption{Training set of the USPS database (20 images for digit $5$). The variability within a class in the dataset we consider for this experiment is exhibited here. We can observe that all the images present particular features both in terms of sharpness and shape.}
\label{fig:variancedigit}
\end{figure}

\noindent \textbf{Numerical Experiment on the U.S. Postal Service database:}
We apply model~\eqref{eq:deformablemodel} and Algorithm~\ref{alg:ttsem} to the US postal database~\citep{hull1994database}, a collection of handwritten digits featuring $n = 1\, 000$, $(16 \times 16)$-pixel images for each class of digits from $0$ to $9$.
The main challenge with this dataset stems from the geometric dispersion within each class of digit as shown Figure~\ref{fig:variancedigit} for digit $5$.
Hence, we ought to use our deformable template model~\eqref{eq:deformablemodel} in order to account for both sources of variability, \ie the intrinsic template of each class of digit and the small and local deformations in each observed image.

\begin{figure}[b!]
\includegraphics[width=\textwidth]{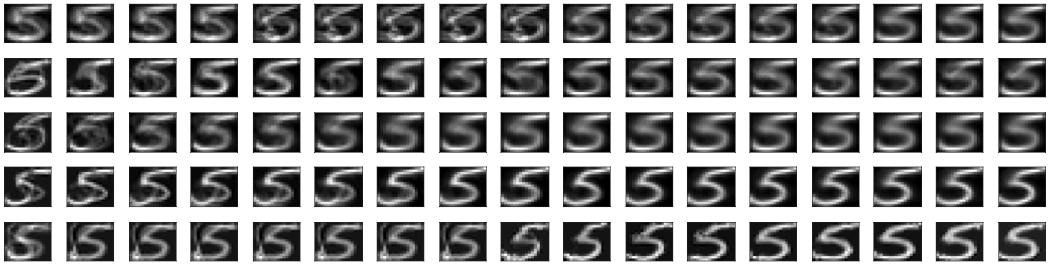}
\caption{(USPS Digits) Estimation of the template. From top to bottom: batch, online, \ISAEM,\ \SAEMVR\ and \FISAEM\ through 7 epochs. Batch method templates are replicated in-between epochs for a fair comparison with incremental variants. }
\label{fig:results}\vspace{-0.1in}
\end{figure}

Figure~\ref{fig:results} shows the resulting synthetic images for digit $5$ through several epochs, for the batch method, the online SAEM, the incremental SAEM and the various two-timescale methods.
For all methods, the initialization of the template~\eqref{eq:template} is the mean of the gray level images.
In our experiments, we have chosen Gaussian kernels for both, $\mathbf{K}_{\mathbf{p}}$ and $\mathbf{K}_{\mathbf{g}}$, defined on $\rset^2$ and centered on the landmark points$\{p_k\}_{k=1}^{k_p}$ and $\{g_k\}_{k=1}^{k_g}$ with standard respective standard deviations of $0.12$ and $0.3$.
We set $k_p = 15$  and  $k_g = 6$ equidistributed landmarks points on the grid for the training procedure.

The hyperparameters are kept the same and are set as $M = 400$, $ \gamma_k = 1/k^{0.6}$ and $ p = 16$.
The standard deviation of the measurement errors is set to $0.1$.
Those hyperparameters are inspired by relevant studies~\citep{allassonniere2010construction,allassonniere2013statistical}.
For the sampling phase of our methods, we use the Carlin and Chib MCMC procedure, see~\citet{carlin1995bayesian}, refer to~\citet{maire2016online} for more details.

In particular, the choice of the geometric covariance, indexed by $g$, in our study is critical since it has a direct impact on the {sharpness} of the templates.
As for the photometric hyperparameter, indexed by $p$, both the template and the geometry are impacted, in the sense that with a large photometric variance, the kernel centered on one landmark {spreads out} to many of its neighbors.

As the iterations proceed, the templates become progressively sharper.
Figure~\ref{fig:results} displays the virtue of the \SAEMVR\ and \FISAEM\ methods leading to a more {contrasted} and {accurate} template estimate.
The incremental and online versions are better in the very first epochs compared to the batch method, given the high computational cost of the latter.
After a few epochs, the batch SAEM estimates similar template as the incremental and online methods due to their high variance.
Our variance reduced and fast incremental variants are effective in the long run and sharpen the template estimates contrasting between the background and the regions of interest in the image.

\vspace{0.1in}
\subsection{Pharmacokinetics (PK) Model with Absorption Lag Time}
\vspace{0.1in}

The following numerical example deals characterizes the pharmacokinetics (PK) of orally administered drug to simulated patients, using a population approach, \ie the training set consists of numerous drug plasmatic concentration per patient of the cohort.
Specifically, $M = 50$ synthetic datasets were generated for $n = 5000$ patients with $10$ observations (concentration measures) per patient.
The goal is to model the evolution of the concentration of the absorbed drug using a {nonlinear} and {latent} variable model.
We consider a one-compartment PK model for oral administration with an absorption lag-time ($T^{\textrm{lag}}$), assuming first-order absorption and linear elimination processes.

\vspace{0.2in}
\noindent \textbf{Model and Explicit Updates:}
The final model includes the following variables: $ka$ the absorption rate constant, $V$ the volume of distribution, $k$ the elimination rate constant and $T^{\textrm{lag}}$ the absorption lag-time.
We also add several covariates to our model such as $D$ the dose of drug administered, $t$ the time at which measures are taken and the weight of the patient influencing the volume $V$. More precisely, the log-volume $\log(V)$ is a linear function of the log-weight $lw70= \log(wt/70)$.
Let $ z_i=(T_i^{\textrm{lag}}, ka_i, V_i, k_i)$ be the vector of individual PK parameters, different for each individual $i$.
The final model reads:
\begin{equation} \label{eq:pkmodel}
\begin{split}
 y_{ij} = f(t_{ij},z_i)+ \varepsilon_{ij} \quad  \textrm{where} \quad f(t_{ij},z_i) = \frac{D\,ka_i}{V(ka_i - k_i)}(\exponential^{-ka_i\,(t_{ij} - T_i^{\textrm{lag}})}-\exponential^{-k_i\,(t_{ij} - T_i^{\textrm{lag}})})\eqs,
\end{split}
\end{equation}
where $y_{ij}$ is the $j$-th concentration measurement of the drug of dosage $D$ injected at time $t_{ij}$ for patient $i$.
We assume in this example that the residual errors $\varepsilon_{ij}$ are independent and normally distributed with mean 0 and variance $\sigma^2$.
Lognormal distributions are used for the four PK parameters:
\begin{align}
& \log(T_i^{\textrm{lag}}) \sim \mathcal{N}(\log(T^{\textrm{lag}}_{\rm pop}), \omega^2_{T^{\textrm{lag}}} ), \quad \log(ka_i) \sim \mathcal{N}(\log(ka_{\rm pop}), \omega^2_{ka})\eqs,\notag\\
&\log(V_i) \sim \mathcal{N}(\log(V_{\rm pop}), \omega^2_{V}), \quad
 \log(k_i) \sim \mathcal{N}(\log(k_{\rm pop}), \omega^2_{k})\eqs.\notag
\end{align}

\newpage

We note that the complete model $p(y,z)$ defined by the structural model in~\eqref{eq:pkmodel} belongs to the curved exponential family, which vector of sufficient statistics $S = (S_1(z),S_2(z),S_3(z) )$ reads:
\beq \label{eq:suffstat_deformable3}
\begin{split}
S_1(z)  = \frac{1}{n} \sum_{i=1}^n z_i  , \quad S_2(z) =\frac{1}{n} \sum_{i=1}^n z_i^\top z_i , \quad S_3(z)  = \frac{1}{n}  \sum_{i=1}^n  \left(y_i - f(t_{i},z_i)\right)^2 \eqs ,
\end{split}
\eeq
where we have noted $y_i$ and $t_i$ the vector of observations and time for each patient $i \in \inter$.
At iteration $k$, and setting the number of MC samples to $1$ for the sake of clarity, the MC sampling $z_i^{(k)} \sim p(z_i |y_i; \param^{(k)})$ is performed using a Metropolis-Hastings procedure detailed in Algorithm~\ref{alg:mh}. The quantities $\stt^{(k+1)}$ and $\hat{\bss}^{(k+1)}$ are then updated according to the different methods introduced in our paper, see Table~\ref{alg:prox}.
Finally the maximization step yields:
\beq \label{eq:mstep_pk}
\overline{\param} ( {\bm s} )
= \left(
\begin{array}{c}
\hat{\bss}^{(k+1)}_1 \\
\hat{\bss}^{(k+1)}_2 - \hat{\bss}^{(k+1)}_1 \left(\hat{\bss}^{(k+1)}_1 \right)^\top \vspace{.2cm} \\
\hat{\bss}^{(k+1)}_3
\end{array}
\right)
= \left(
\begin{array}{c}
\overline{\bm{z_{\rm pop}}} ( \hat{\bss}^{(k+1)}) \\
\overline{\bm{\omega_{z}}} ( \hat{\bss}^{(k+1)}) \\
\overline{\bm{\sigma}} ( \hat{\bss}^{(k+1)})
\end{array}
\right) \eqsp,
\eeq
where $z_{\rm pop}$ denotes the vector of fixed effects $(T^{\textrm{lag}}_{\rm pop}, ka_{\rm pop}, V_{\rm pop}, k_{\rm pop})$.

\begin{algorithm}[t]
\begin{algorithmic}[1]
\STATE \textbf{Input:} initialization $z_{i,0} \sim q(z_{i}; {\bm \delta})$
\FOR {$m=1, \cdots ,M$}
\STATE Sample $z_{i,m} \sim q(z_{i}; {\bm \delta})$.
\STATE Sample $u \sim \mathcal{U}([0, 1])$.
\STATE Calculate the ratio $r = \frac{\pi(z_{i,m}; \param)/q(z_{i,m}); {\bm \delta})}{\pi(z_{i,m-1}; \param)/q(z_{i,m-1}); {\bm \delta})}$.
\STATE \textbf{if} $u < r$ \textbf{then} accept $z_{i,m}$ \textbf{else} $z_{i,m} \leftarrow z_{i,m-1}$
\ENDFOR
\STATE \textbf{Output:} $z_{i,M}$
\end{algorithmic}
\caption{Metropolis-Hastings algorithm}
\label{alg:mh}
        \end{algorithm}

\begin{figure}[t]
\centering
\includegraphics[width=4in]{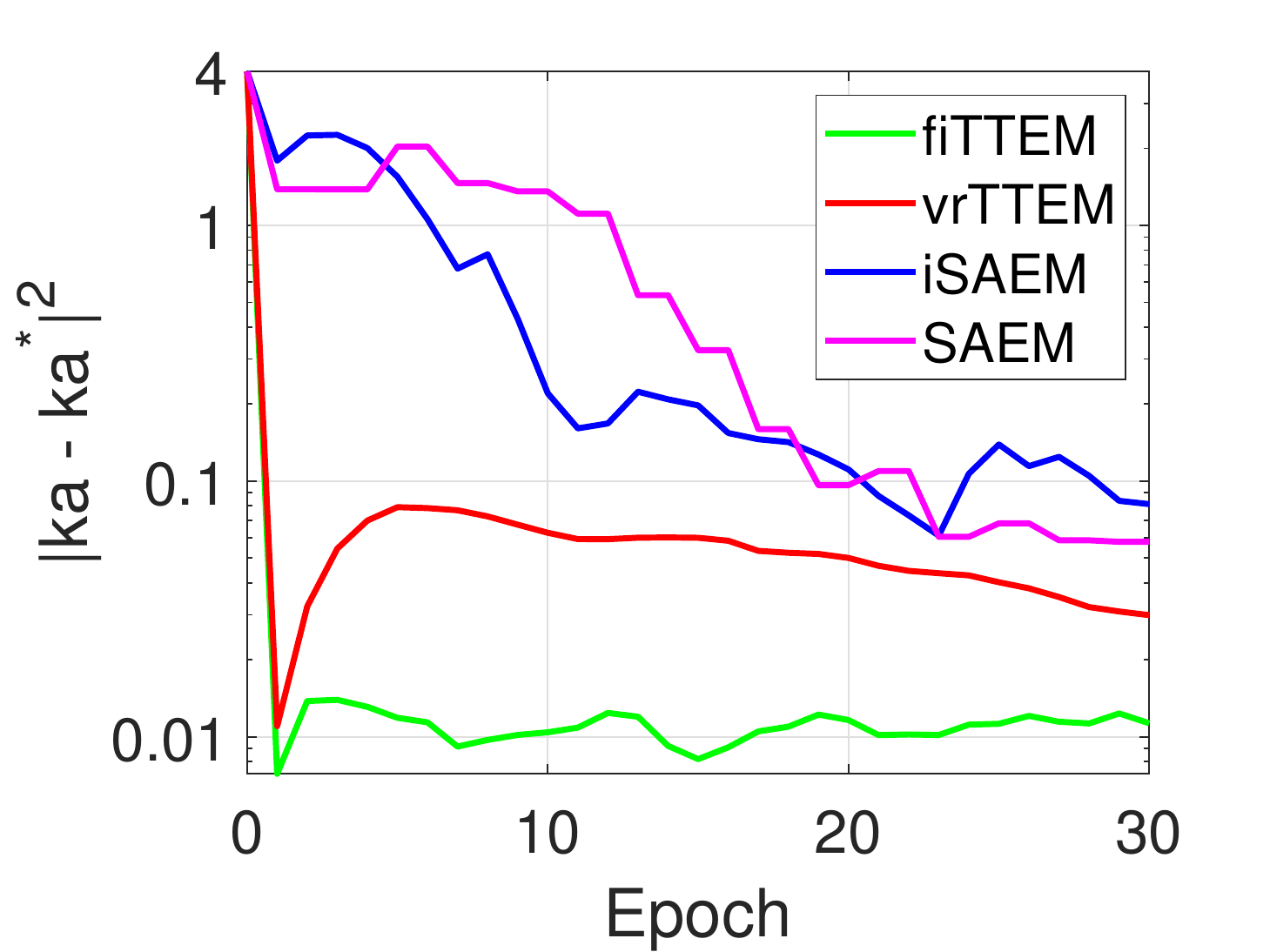}
\caption{Mean square errors $|ka^{(k)} - ka^*|^2$ for our methods (\FISAEM\ in green,\SAEMVR\ in black and \ISAEM\ in blue) versus the SAEM baseline in red, against epochs elapsed. The errors have been averaged over $M=50$ synthetic datasets for robustness. The \FISAEM\ appears to be the best method among the four. The other variance reduced TTSEM method, namely \SAEMVR\, is quickly reaching a similar accuracy but exhibits overfitting rather quickly. The two other plain SAEM methods are the slowest.}
\label{fig:pk_tts}
\end{figure}

\vspace{0.15in}
\noindent \textbf{Monte Carlo study:}
We conduct a Monte Carlo study to showcase the benefits of our scheme.
$M=50$ datasets have been simulated using the following PK parameters values:
$T^{\textrm{lag}}_{\rm pop} =1$, $ka_{\rm pop} =1$, $V_{\rm pop}= 8$, $k_{\rm pop}=0.1$, $ \omega_{T^{\textrm{lag}}}=0.4$, $\omega_{ka}=0.5$, $\omega_{V}=0.2$, $\omega_{k}=0.3$ and $\sigma^2=0.5$.
We define the mean square distance over the $M$ replicates as $E_k(\ell) = \frac{1}{M}\sum_{m=1}^{M}{\left(\param_k^{(m)}(\ell) - \param^* \right)^2} $, and plot it against the epochs (passes over the data) in Figure~\ref{fig:pk_tts}.	
Note that the { MC-step}~\eqref{eq:mcstep} is performed using a Metropolis-Hastings procedure since the posterior distribution under the model $\param$ noted $p(z_i | y_i; \param)$ is intractable, mainly due to the nonlinearity of the model~\eqref{eq:pkmodel}.
The Metropolis-Hastings (MH) algorithm~\citep{meyn2012markov} leverages a proposal distribution $q(z_{i}, \delta)$ where $\param = (z_{\rm pop}, \omega_{z})$ and $ \delta$ is the vector of parameters of the proposal distribution. Generally, and for simplicity, a Gaussian proposal is used.
The MH algorithm employed to sample from each individual posterior distribution $\left(p(z_i | y_i; \param), i\in \inter \right)$ is summarized in Algorithm~\ref{alg:mh}.

Figure~\ref{fig:pk_tts} shows clear advantage of variance reduced methods (\SAEMVR\ and \FISAEM ) avoiding the twists and turns displayed by the incremental and the batch methods (iSAEM and SAEM).
Both our newly proposed EM methods quickly reaches a neighborhood of the solution while baselines slowly converge to it empirically stressing on the benefits of our two-timescale methods that not only temper the noise of the incremental update but also reduce the MC noise stemming from a required approximation of the expectations.

\newpage

\section{Conclusion}

In this paper we have introduced a new class of two-timescale EM methods for learning latent variable models.
In particular, the models dealt with in this paper belong to the curved exponential family and are possibly nonconvex.
The nonconvexity of the problem is tackled using a Robbins-Monro type of update, which represents the {first level} of our class of methods.
The scalability with the number of samples is performed through a variance reduced and incremental update, the {second} and last level of the scheme we introduce in this paper.
The various algorithms are interpreted as scaled gradient methods, in the space of the sufficient statistics, and our convergence results are {global}, in the sense of independence of the initial values, and {non-asymptotic}, \ie true for any termination iteration index.
We singularly deal with the Monte Carlo noise introduced by the stochastic approximation in order to derive those convergence bounds.
We empirically and theoretically show that variance reduction techniques applied to Stochastic EM type of algorithms lead to a faster convergence of the optimization phase.
A panoply of numerical examples, carried out in various latent variable models, illustrate the benefits of our scheme on synthetic and real datasets.
In particular, our numerical runs validate the benefits of using variance reduce variants of the SAEM over standard incremental baselines.

\newpage
\bibliographystyle{plainnat}
\bibliography{references}
\newpage

\appendix

\section{Proofs for the \ISAEM\ Algorithm}
\subsection{Proof of Lemma~\ref{lem:growth}}\label{app:growth}
\begin{Lemma*}
Assume A\ref{ass:reg}, A\ref{ass:eigen}. For all $\bss \in \Sset$,
\beq \label{eq:semigrad2}
\upsilon_{\min}^{-1} \pscal{\grd V ( {\bss} ) }{ {\bss} - \os( \op ({\bss})) }
\geq \| {\bss} - \os( \op ({\bss})) \|^2 \geq \upsilon_{\max}^{-2} \| \grd V ( {\bss} ) \|^2.
\eeq
\end{Lemma*}
\begin{proof}
Using A\ref{ass:reg} and the fact that we can exchange integration with differentiation and the Fisher's identity,   we obtain
\beq \label{eq:grd_v}
\begin{split}
\grd_{ \bss} V( {\bss} ) & = \jacob{ \overline{\param} }{ \bss }{\bss}^\top
( \grd_\param \Pen( \mstep{\bss} )  + \grd_\param \calL( \overline\param( {\bss} ) )  ) \\
& =  \jacob{ \overline{\param} }{ \bss }{\bss}^\top ( \grd_\param \psi( \mstep{\bss}) + \grd_\param \Pen( \mstep{\bss} ) - \jacob{\phi}{\param}{\mstep{\bss} }^\top  \os( \op ({\bss})) )\\
& =   \jacob{ \overline{\param} }{ \bss }{\bss}^\top \jacob{\phi}{\param}{ \mstep{\bss} }^\top \!~ ({\bss} - \os( \op ({\bss})) ) \eqsp.
\end{split}
\eeq
Consider the following vector map:
\beq\notag
{\bss} \to \grd_{\param} L(\bss; \param) \vert_{\param= \mstep{\bss}}= \grd_\param \psi ( \mstep{\bss} ) + \grd_{ \param} \Pen(\mstep{\bss}  ) - \jacob{ \phi }{ \param }{\mstep{\bss}  }^\top \!~{\bss} \eqsp.
\eeq
Taking the gradient of the above map w.r.t. ${\bss}$ and using assumption A\ref{ass:reg}, we show that:
\beq\notag
{\bm 0} = - \jacob{\phi}{\param}{\mstep{\bss} } + ( \underbrace{ \grd_{\param}^2 ( \psi( \param ) + \Pen( \param ) - \pscal{ \phi( \param ) }{ {\bss} } )}_{= \hess{{L}}{\param} ( {\bss}; \param )} \big|_{\param = \mstep{\bss}  } ) \jacob{ \overline{\param} }{\bss}{\bss} \eqsp.
\eeq
The above yields
\beq\notag
\grd_{ \bss} V( {\bss} )  = \operatorname{B}(\bss) ({\bss} - \os( \op ({\bss})) ) \eqsp,
\eeq
where we recall $\operatorname{B}(\bss) = \jacob{ \phi }{ \param }{ \mstep{\bss} } ( \hess{{L}}{\param}( {\bss}; \mstep{\bss} )  )^{-1} \jacob{ \phi }{ \param }{\mstep{\bss} }^\top$. The proof of~\eqref{eq:semigrad2} follows directly from the assumption~A\ref{ass:eigen}.
\end{proof}

\vspace{0.2in}

\subsection{Proof of Theorem~\ref{thm:isaem}}\label{app:theoremisaem}
Beforehand, We present two intermediary Lemmas important for the analysis of the incremental update of the iSAEM algorithm.
The first one gives a characterization of the quantity $\EE[\stt^{(k+1)} - \hat{\bss}^{(k)}]$:
\begin{Lemma*}
 Assume A\ref{ass:compact}. The update~\eqref{eq:isaem} is equivalent to the following update on the resulting statistics
\beq\notag
\hat{\bss}^{(k+1)} =  \hat{\bss}^{(k)}  + \gamma_{k+1} ( \stt^{(k+1)} - \hat{\bss}^{(k)} ) \eqsp.
\eeq
Also:
\beq\notag
\EE[\stt^{(k+1)} - \hat{\bss}^{(k)}] = \EE[\overline{\bss}^{(k)} - \hat{\bss}^{(k)}] + (1 - 1/n ) \EE[\frac{1}{n} \sum_{i=1}^n \tilde{S}_i^{(\tau_i^k)}- \overline{\bss}^{(k)}]  +\frac{1}{n}\EE[\eta_{i_k}^{(k+1)}]\eqsp ,
\eeq
where $\overline{\bss}^{(k)}$ is defined by~\eqref{eq:definition-overline-bss} and $\tau_i^k = \max \{ k' : i_{k'} = i,~k' < k \}$.
\end{Lemma*}
\begin{proof}
From update~\eqref{eq:isaem}, we have:
\beq\notag
\begin{split}
\stt^{(k+1)} - \hat{\bss}^{(k)} & = \stt^{(k)} - \hat{\bss}^{(k)} +\frac{1}{n}( \tilde{S}_{i_k}^{(k+1)} - \tilde{S}_{i_k}^{(\tau_i^k)}  )\\
& = \overline{\bss}^{(k)} - \hat{\bss}^{(k)} + \stt^{(k)}- \overline{\bss}^{(k)}  - \frac{1}{n}( \tilde{S}_{i_k}^{(\tau_i^k)} - \tilde{S}_{i_k}^{(k+1)}   ) \eqsp .
\end{split}
\eeq
Since $\tilde{S}_{i_k}^{(k+1)} = \overline{\bss}_{i_k}(\param^{(k)}) + \eta_{i_k}^{(k+1)}$ we have
\beq\notag
\begin{split}
\stt^{(k+1)} - \hat{\bss}^{(k)} = \overline{\bss}^{(k)} - \hat{\bss}^{(k)} + \stt^{(k)}- \overline{\bss}^{(k)}  - \frac{1}{n}( \tilde{S}_{i_k}^{(\tau_i^k)} -  \overline{\bss}_{i_k}(\param^{(k)})   ) + \frac{1}{n}\eta_{i_k}^{(k+1)}\eqsp .
\end{split}
\eeq
Taking the full expectation of both side of the equation leads to:
\beq\notag
\begin{split}
\EE[\stt^{(k+1)} - \hat{\bss}^{(k)}] = \EE[\overline{\bss}^{(k)} - \hat{\bss}^{(k)}] & + \EE[\frac{1}{n} \sum_{i=1}^n \tilde{S}_i^{(\tau_i^k)}-  \overline{\bss}^{(k)}] \\
& -\frac{1}{n} \EE[\EE[ \tilde{S}_i^{(\tau_i^k)}-  \overline{\bss}_{i_k}(\param^{(k)})  | \mathcal{F}_{k} ]] + \frac{1}{n} \EE[\eta_{i_k}^{(k+1)}] \eqsp.
\end{split}
\eeq
Since we have $\EE[ \tilde{S}_i^{(\tau_i^k)} | \mathcal{F}_{k} ] =\frac{1}{n} \sum_{i=1}^n \tilde{S}_i^{(\tau_i^k)}$ and $\EE[  \overline{\bss}_{i_k}(\param^{(k)})  | \mathcal{F}_{k} ]= \overline{\bss}^{(k)}$, we conclude the proof of the Lemma.
\end{proof}

We also derive the following auxiliary Lemma which sets an upper bound for the quantity $\EE [ \|  \stt^{(k+1)} - \hs{k}   \|^2 ]$:
\begin{Lemma*}
For any $k \geq 0$ and consider the \ISAEM\ update in~\eqref{eq:isaem}, it holds that
\beq\notag
\begin{split}
\EE [ \|  \stt^{(k+1)} - \hs{k}   \|^2 ] \leq 4 \EE[ \|  \os^{(k)} - \hs{k} \|^2 ]
+ \frac{2\Lip{\bss}^2}{n^3} \sum_{i=1}^n \EE[ \| \hs{k} - \hs{t_i^k} \|^2 ]+ 2\frac{c_{\eta}}{M_k} + 4 \EE[\|\frac{1}{n} \sum_{i=1}^n \tilde{S}_i^{(\tau_i^k)}-  \overline{\bss}^{(k)}\|^2]  \eqsp.
\end{split}
\eeq
\end{Lemma*}

\begin{proof}
Applying the \ISAEM\ update yields the following inequality:
\beq\notag
\begin{split}
   \EE[ \|  \stt^{(k+1)} - \hs{k} \|^2 ]
 =&  \EE[ \| \stt^{(k)} - \hs{k}  -\frac{1}{n}(\tilde{S}^{(\tau_i^k)}_{i_k} - \tilde{S}^{(k)}_{i_k}  )  \|^2 ]\\
 \leq  & 4 \EE[\|\frac{1}{n} \sum_{i=1}^n \tilde{S}_i^{(\tau_i^k)}-  \overline{\bss}^{(k)}\|^2] + 4 \EE[\|   \overline{\bss}^{(k)} - \hs{k} \|^2] +  \frac{2}{n^2} \EE[ \| \os_{i_k}^{(k)} - \os_{i_k}^{(t_{i_k}^k)} \|^2] + 2\frac{c_{\eta}}{M_k} \eqsp.
\end{split}
\eeq
The last expectation can be further bounded by
\beq\notag
\begin{split}
&
\frac{2}{n^2}\EE[ \| \os_{i_k}^{(k)} - \os_{i_k}^{(t_{i_k}^k)} \|^2 ] = \frac{2}{n^3} \sum_{i=1}^n \EE[ \| \os_i^{(k)} - \os_i^{(t_i^k)} \|^2 ] \overset{(a)}{\leq} \frac{2\Lip{\bss}^2}{n^3}
\sum_{i=1}^n \EE[ \| \hs{k} - \hs{t_i^k} \|^2 ]\eqsp,
\end{split}
\eeq
where (a) is due to Lemma~\ref{lem:smooth} and which concludes the proof of the Lemma.

\end{proof}

\begin{Theorem*}
Assume A\ref{ass:compact}-A\ref{ass:mcerror}.
Consider the \ISAEM\ sequence $\{\hat{\bss}^{(k)}\}_{k>0} \in \mathcal{S}$ obtained with $\rho_{k+1}=1$ for any $k \leq { K}_{ m }$ where ${ K}_{ m }$ is a positive integer.
Let $\{\gamma_{k} = 1/(k^a \alpha c_1 \overline{L})\}_{k>0}$, where $a \in (0,1)$, be a sequence of stepsizes, $c_1 = \upsilon_{\min}^{-1}$, $\alpha = \max\{8, 1+6\upsilon_{\min}\}$, $\overline{L} = \max\{ \Lip{\bss} , \Lip{V} \}$, $\beta = c_1 \overline{L}/n$. Then:
\beq\notag
\upsilon_{\max}^{-2}\sum_{k=0}^{{ K}_{ m }} \tilde{\alpha}_k \EE [\|\grd V( \hs{k} )\|^2]  \leq   \EE  [V( \hs{0} ) - V( \hs{{ K}_{ m }} ) ] + \sum_{k=0}^{{ K}_{ m }-1} \tilde{\Gamma}_k         \EE [\| \eta_{i_k}^{(k)}\|^2] \eqs.
\eeq
\end{Theorem*}
\begin{proof}
Under the smoothness of the Lyapunov function $V$ (cf. Lemma~\ref{lem:smooth}), we can write:
\beq\notag
\begin{split}
V( \hs{k+1} ) & \leq V( \hs{k} ) + \gamma_{k+1} \pscal{  \stt^{(k+1)}  - \hs{k}}{ \grd V( \hs{k} ) } + \frac{\gamma_{k+1}^2 \Lip{V}}{2} \|\stt^{(k+1)} -  \hs{k}  \|^2 \eqsp.
\end{split}
\eeq
Taking the expectation on both sides yields:
\beq\notag
\begin{split}
\EE [V( \hs{k+1} ) ]  \leq \EE [ V( \hs{k} ) ]  + \gamma_{k+1} \EE [\pscal{  \stt^{(k+1)}  - \hs{k}}{ \grd V( \hs{k} ) }  ]+ \frac{\gamma_{k+1}^2 \Lip{V}}{2} \EE [\|\stt^{(k+1)} -  \hs{k}  \|^2  ]\eqsp.
\end{split}
\eeq
Using Lemma~\ref{lem:meanfield_isaem}, we obtain:
\beq\notag
\begin{split}
& \EE [\pscal{  \stt^{(k+1)}  - \hs{k}}{ \grd V( \hs{k} ) }  ] \\
= &  \EE [\pscal{  \overline{\bss}^{(k)}  - \hs{k}}{ \grd V( \hs{k} ) }  ]  + \left(1 - \frac{1}{n}\right)\EE[\pscal{ \frac{1}{n} \sum_{i=1}^n \tilde{S}_i^{(\tau_i^k)}-  \overline{\bss}^{(k)}}{ \grd V( \hs{k} ) }] \\
& +  \frac{1}{n} \EE [\pscal{ \eta_{i_k}^{(k)}}{ \grd V( \hs{k} ) }  ]\\
 \overset{(a)}{\leq} & -\upsilon_{\min}\EE [\|  \overline{\bss}^{(k)}  - \hs{k}\|^2  ] + \left(1 - \frac{1}{n}\right)\EE[\pscal{ \frac{1}{n} \sum_{i=1}^n \tilde{S}_i^{(\tau_i^k)}-  \overline{\bss}^{(k)}}{ \grd V( \hs{k} ) }] \\
 & +  \frac{1}{n} \EE [\pscal{ \eta_{i_k}^{(k)}}{ \grd V( \hs{k} ) }  ]\\
 \overset{(b)}{\leq} & -\upsilon_{\min}\EE [\|  \overline{\bss}^{(k)}  - \hs{k}\|^2  ] + \frac{1 - \frac{1}{n}}{2\beta}\EE[\|\frac{1}{n} \sum_{i=1}^n \tilde{S}_i^{(\tau_i^k)}-  \overline{\bss}^{(k)}\|^2]\\
& +  \frac{\beta(n-1) + 1}{2n}\EE[ \norm{\grd V( \hs{k} )}^2]  +  \frac{1}{2 n} \EE [\| \eta_{i_k}^{(k)}\|^2 ] \\
 \overset{(a)}{\leq} & \left(\upsilon^2_{\max}\frac{\beta(n-1) + 1}{2n}-\upsilon_{\min}\right) \EE [\|  \overline{\bss}^{(k)}  - \hs{k}\|^2  ] + \frac{1 - \frac{1}{n}}{2\beta}\EE[\|\frac{1}{n} \sum_{i=1}^n \tilde{S}_i^{(\tau_i^k)}-  \overline{\bss}^{(k)}\|^2]\\
 & +  \frac{1}{2 n} \EE [\| \eta_{i_k}^{(k)}\|^2 ] \eqsp,
\end{split}
\eeq
where (a) is due to the growth condition~\eqref{lem:growth} and (b) is due to Young's inequality (with $\beta \to 1$).
Note $a_k = \gamma_{k+1}\left(\upsilon_{\min} - \upsilon^2_{\max}\frac{\beta(n-1) + 1}{2n}\right) $ and
\beq\label{eq:final1}
\begin{split}
a_k \EE [\|  \overline{\bss}^{(k)}  - \hs{k}\|^2  ]  \leq & \EE [ V( \hs{k} ) - V( \hs{k+1} ) ] + \frac{\gamma_{k+1}^2 \Lip{V}}{2} \EE [\|\stt^{(k+1)} -  \hs{k}  \|^2  ]\\
&+ \frac{\gamma_{k+1}(1 - \frac{1}{n})}{2\beta}\EE[\|\frac{1}{n} \sum_{i=1}^n \tilde{S}_i^{(\tau_i^k)}-  \overline{\bss}^{(k)}\|^2]+  \frac{\gamma_{k+1}}{2 n} \EE [\| \eta_{i_k}^{(k)}\|^2 ] \eqsp.
\end{split}
\eeq
We now give an upper bound of $\EE [\|\stt^{(k+1)} -  \hs{k}  \|^2  ]$ using Lemma~\ref{lem:aux2} and plug it into~\eqref{eq:final1}:
\beq\label{eq:final2}
\begin{split}
 ( a_k - 2\gamma_{k+1}^2 \Lip{V} ) \EE [\|  \overline{\bss}^{(k)}  - \hs{k}\|^2 ]
\leq  &  \EE [ V( \hs{k} ) - V( \hs{k+1} ) ] \\
&  +   \gamma_{k+1} \left(\frac{1}{2 \beta}(1 - 1/n ) + 2 \gamma_{k+1}\Lip{V} \right)            \EE[\|\frac{1}{n} \sum_{i=1}^n \tilde{S}_i^{(\tau_i^k)}-  \overline{\bss}^{(k)}\|^2]\\
& + \gamma_{k+1} \left(\gamma_{k+1} \Lip{V} +    \frac{1}{2 n}\right)           \EE [\| \eta_{i_k}^{(k)}\|^2 ] \\
& + \frac{\gamma_{k+1}^2 \Lip{V}\Lip{\bss}^2}{n^3} \sum_{i=1}^n \EE[ \| \hs{k} - \hs{\tau_i^k} \|^2 ] \eqsp.
\end{split}
\eeq
Next, we observe that
\beq\notag
\frac{1}{n} \sum_{i=1}^n \EE[ \| \hs{k+1} - \hs{t_i^{k+1}} \|^2 ] = \frac{1}{n} \sum_{i=1}^n
( \frac{1}{n} \EE[ \| \hs{k+1} - \hs{k} \|^2 ] + \frac{n-1}{n} \EE[ \| \hs{k+1} - \hs{\tau_i^k} \|^2 ]  )\eqsp,
\eeq
where the equality holds as $i_k$ and $j_k$ are drawn independently. For any $\beta > 0$, it holds
\beq\notag
\begin{split}
& \EE[ \| \hs{k+1} - \hs{t_i^k} \|^2 ] \\
 =& \EE [ \| \hs{k+1} - \hs{k} \|^2 + \| \hs{k} - \hs{\tau_i^k} \|^2 + 2 \pscal{\hs{k+1} - \hs{k}}{\hs{k}- \hs{\tau_i^k}} ] \\
=& \EE [ \| \hs{k+1} - \hs{k} \|^2 + \| \hs{k} - \hs{\tau_i^k} \|^2 - 2 \gamma_{k+1} \pscal{ \hs{k} - \stt^{(k+1)} }{\hs{k}- \hs{\tau_i^k}} ] \\
\leq&  \EE [ \| \hs{k+1} - \hs{k} \|^2 + \| \hs{k} - \hs{\tau_i^k} \|^2 +  \frac{\gamma_{k+1}}{\beta} \| \hs{k} - \stt^{(k+1)}\|^2 \\
 &+ \gamma_{k+1} \beta \| \hs{k}- \hs{\tau_i^k} \|^2 ]\eqsp,
\end{split}
\eeq
where the last inequality is due to Young's inequality. Subsequently, we have
\beq\notag
\begin{split}
  \frac{1}{n} \sum_{i=1}^n \EE[ \| \hs{k+1} - \hs{\tau_i^{k+1}} \|^2 ]
 \leq & \EE[  \| \hs{k+1} - \hs{k} \|^2 ] + \frac{n-1}{n^2} \sum_{i=1}^n \EE [ (1+\gamma_{k+1} \beta) \|  \hs{k} - \hs{\tau_i^k} \|^2
\\
 & + \frac{\gamma_{k+1}}{\beta} \|  \hs{k} - \stt^{(k+1)} \|^2 ]\eqsp.
\end{split}
\eeq
Observe that $\hs{k+1} - \hs{k} = - \gamma_{k+1} ( \hs{k} - \stt^{(k+1)} )$. Applying Lemma~\ref{lem:aux2} yields
\beq\notag
\begin{split}
& \frac{1}{n} \sum_{i=1}^n \EE[ \| \hs{k+1} - \hs{\tau_i^{k+1}} \|^2 ] \\
 \leq &(\gamma_{k+1}^2 +\frac{n-1}{n}\frac{\gamma_{k+1}}{\beta}  )\EE [  \|   \stt^{(k+1)} -  \hs{k} \|^2  ] + \sum_{i=1}^n \EE [  \frac{1 - \frac{1}{n} + \gamma_{k+1} \beta}{n} \|  \hs{k} - \hs{\tau_i^k} \|^2  ] \\
 \leq & 4(\gamma_{k+1}^2 +\frac{\gamma_{k+1}}{\beta}  )\EE [  \|   \os^{(k)} - \hs{k}  \|^2  ] + 2(\gamma_{k+1}^2 +\frac{\gamma_{k+1}}{\beta}  )\EE [\| \eta_{i_k}^{(k)}\|^2 ]\\
+&  4 (\gamma_{k+1}^2 +\frac{\gamma_{k+1}}{\beta}  )\EE[\|\frac{1}{n} \sum_{i=1}^n \tilde{S}_i^{(\tau_i^k)}-  \overline{\bss}^{(k)}\|^2] \\
+&  \sum_{i=1}^n \EE [ \frac{1 - \frac{1}{n} + \gamma_{k+1} \beta + \frac{2\gamma_{k+1} \Lip{\bss}^2}{n^2}(\gamma_{k+1} +\frac{1}{\beta})}{n} \|  \hs{k} - \hs{t_i^k} \|^2  ]  \eqsp.
\end{split}
\eeq
Let us define
\beq\notag
\Delta^{(k)} \eqdef \frac{1}{n} \sum_{i=1}^n \EE[ \| \hs{k} - \hs{\tau_i^{k}} \|^2 ]\eqsp.
\eeq
From the above, we obtain
\beq\notag
\begin{split}
 \Delta^{(k+1)} & \leq  (1 - \frac{1}{n} + \gamma_{k+1} \beta + \frac{2\gamma_{k+1} \Lip{\bss}^2}{n^2}(\gamma_{k+1} +\frac{1}{\beta})  ) \Delta^{(k)} \\
& +4 (\gamma_{k+1}^2 +\frac{\gamma_{k+1}}{\beta}  ) \EE [  \|   \os^{(k)} - \hs{k}  \|^2  ]  + 2(\gamma_{k+1}^2  +\frac{\gamma_{k+1}}{\beta}  )\EE [\| \eta_{i_k}^{(k)}\|^2 ]\\
&+  4 (\gamma_{k+1}^2 +\frac{\gamma_{k+1}}{\beta}  ) \EE[\|\frac{1}{n} \sum_{i=1}^n \tilde{S}_i^{(\tau_i^k)}-  \overline{\bss}^{(k)}\|^2]\eqsp.
\end{split}
\eeq
Setting $c_1 = \upsilon_{\min}^{-1}$, $\alpha =\max\{8, 1+6\upsilon_{\min}\}$, $\overline{L} = \max\{ \Lip{\bss} , \Lip{V} \}$, $\gamma_{k+1} = \frac{1}{k \alpha c_1 \overline{L}}$, $\beta = \frac{c_1 \overline{L}}{n}$, we remark $c_1(k\alpha-1) \geq c_1(\alpha-1) \geq 6$ and we observe that
\beq\notag
1 - \frac{1}{n} + \gamma_{k+1} \beta + \frac{2\gamma_{k+1} \Lip{\bss}^2}{n^2}(\gamma_{k+1} +\frac{1}{\beta})
 \leq 1 - \frac{c_1(k\alpha  - 1) - 4}{k\alpha n c_1 } \leq 1 - \frac{2}{k\alpha n c_1 }\eqsp,
\eeq
which shows that $1 - \frac{1}{n} + \gamma_{k+1} \beta + \frac{2\gamma_{k+1} \Lip{\bss}^2}{n^2}(\gamma_{k+1} +\frac{1}{\beta})  \in (0,1)$ for any $k >0$.
Denote $ \Lambda_{(k+1)} =\frac{1}{n} - \gamma_{k+1} \beta - \frac{2\gamma_{k+1} \Lip{\bss}^2}{n^2}(\gamma_{k+1} +\frac{1}{\beta}) $ and note that $\Delta^{(0)} = 0$, thus the telescoping sum yields:
\beq\notag
\begin{split}
\Delta^{(k+1)} & \leq  4 \sum_{ \ell = 0 }^k \prod_{j = \ell +1}^k ( 1 -  \Lambda_{(j)} ) (\gamma_{\ell+1}^2 +\frac{\gamma_{\ell+1}}{\beta}  )  \EE[  \|  \os^{(\ell)} - \hs{\ell}  \|^2 ] \\
&+ 2\sum_{ \ell = 0 }^k \prod_{j = \ell +1}^k ( 1 -  \Lambda_{(j)} ) (\gamma_{\ell+1}^2  +\frac{\gamma_{\ell+1}}{\beta}  ) \EE [\| \eta_{i_\ell}^{(\ell)}\|^2 ]\\
& +  4 \sum_{ \ell = 0 }^k   \prod_{j = \ell +1}^k ( 1 -  \Lambda_{(j)} )  (\gamma_{\ell+1}^2+\frac{\gamma_{\ell+1}}{\beta}  )  \EE[\| \frac{1}{n} \sum_{i=1}^n \tilde{S}_i^{(\tau_i^\ell)}-  \overline{\bss}^{(\ell)}\|^2]\eqsp.
\end{split}
\eeq
Note $\omega_{k,\ell} = \prod_{j = \ell +1}^k ( 1 -  \Lambda_{(j)} )$
Summing on both sides over $k=0$ to $k = { K}_{ m }-1$ yields:

\beq\label{eq:Delta}
\begin{split}
& \sum_{k=0}^{{ K}_{ m }-1} \Delta^{(k+1)}\\
=&  4 \sum_{k=0}^{{ K}_{ m }-1} (\gamma_{k+1}^2 +\frac{\gamma_{k+1}}{\beta}  ) \omega_{k,1} \EE[  \|  \os^{(k)} - \hs{k}  \|^2 ] + 2 \sum_{k=0}^{{ K}_{ m }-1} (\gamma_{k+1}^2  +\frac{\gamma_{k+1}}{\beta}  )\omega_{k,1}\EE [\norm{ \eta_{i_\ell}^{(k)}}^2 ]\\
+ &  \sum_{k=0}^{{ K}_{ m }-1} 4 (\gamma_{k+1}^2 +\frac{\gamma_{k+1}}{\beta}  ) \omega_{k,1}  \EE[\|\frac{1}{n} \sum_{i=1}^n \tilde{S}_i^{(\tau_i^k)}-  \overline{\bss}^{(k)}\|^2]\\
\leq &   \sum_{k=0}^{{ K}_{ m }-1}\frac{4(\gamma_{k+1}^2 +\frac{\gamma_{k+1}}{\beta}  )}{ \Lambda_{(k+1)}}   \EE[  \|  \os^{(k)} - \hs{k}  \|^2 ] + \sum_{k=0}^{{ K}_{ m }-1}\frac{2(\gamma_{k+1}^2 +\frac{\gamma_{k+1}}{\beta}  )}{ \Lambda_{(k+1)}}  \EE [\norm{ \eta_{i_\ell}^{(k)}}^2 ]\\
 +&  \sum_{k=0}^{{ K}_{ m }-1}\frac{4(\gamma_{k+1}^2 +\frac{\gamma_{k+1}}{\beta}  )}{ \Lambda_{(k+1)}}  \EE[\|\frac{1}{n} \sum_{i=1}^n \tilde{S}_i^{(\tau_i^k)}-  \overline{\bss}^{(k)}\|^2]\eqsp.
\end{split}
\eeq
We recall~\eqref{eq:final2} where we have summed on both sides from $k=0$ to $k = { K}_{ m }-1$:
\beq\label{eq:final3}
\begin{split}
&\sum_{k=0}^{{ K}_{ m }-1}  \left( a_k - 2\gamma_{k+1}^2 \Lip{V} \right) \EE [\|  \overline{\bss}^{(k)}  - \hs{k}\|^2  ] \\
 \leq &  \EE [ V( \hs{0} ) - V( \hs{K} ) ] + \sum_{k=0}^{{ K}_{ m }-1} \gamma_{k+1} \left(\frac{1}{2 \beta}(1 - 1/n ) + 2 \gamma_{k+1}\Lip{V} \right)            \EE[\|\frac{1}{n} \sum_{i=1}^n \tilde{S}_i^{(\tau_i^k)}-  \overline{\bss}^{(k)}\|^2]\\
+& \sum_{k=0}^{{ K}_{ m }-1} \gamma_{k+1} \left(\gamma_{k+1} \Lip{V} +    \frac{1}{2 n}\right)           \EE [\| \eta_{i_k}^{(k)}\|^2 ] + \sum_{k=0}^{{ K}_{ m }-1} \frac{\gamma_{k+1}^2 \Lip{V}\Lip{\bss}^2}{n^2} \Delta^{(k)}\eqsp.
\end{split}
\eeq
Plugging~\eqref{eq:Delta} into~\eqref{eq:final3} results in:
\beq\notag
\begin{split}
\sum_{k=0}^{{ K}_{ m }-1}  \tilde{\alpha}_k \EE [\|  \overline{\bss}^{(k)}  - \hs{k}\|^2  ] + \sum_{k=0}^{{ K}_{ m }-1}  \tilde{\beta}_k \EE[\|\frac{1}{n} \sum_{i=1}^n \tilde{S}_i^{(\tau_i^k)}-  \overline{\bss}^{(k)}\|^2]
\leq   \EE [ V( \hs{0} ) - V( \hs{K} ) ]
+ \sum_{k=0}^{{ K}_{ m }-1} \tilde{\Gamma}_k         \EE [\| \eta_{i_k}^{(k)}\|^2 ] \eqsp,
\end{split}
\eeq
where
\begin{align*}
&  \tilde{\alpha}_k = a_k - 2\gamma_{k+1}^2 \Lip{V} -  \frac{\gamma_{k+1}^2 \Lip{V}\Lip{\bss}^2}{n^2}\frac{4(\gamma_{k+1}^2 +\frac{\gamma_{k+1}}{\beta}  )}{ \Lambda_{(k+1)}} \eqsp,  \\
&  \tilde{\beta}_k =  \gamma_{k+1} \left(\frac{1}{2 \beta}(1 - 1/n ) + 2 \gamma_{k+1}\Lip{V} \right) -  \frac{\gamma_{k+1}^2 \Lip{V}\Lip{\bss}^2}{n^2}\frac{4(\gamma_{k+1}^2 +\frac{\gamma_{k+1}}{\beta}  )}{ \Lambda_{(k+1)}}\eqsp, \\
&  \tilde{\Gamma}_k = \gamma_{k+1} \left(\gamma_{k+1} \Lip{V} +    \frac{1}{2 n}\right)  +  \frac{\gamma_{k+1}^2 \Lip{V}\Lip{\bss}^2}{n^2} \frac{2(\gamma_{k+1}^2 +\frac{\gamma_{k+1}}{\beta}  )}{ \Lambda_{(k+1)}}\eqsp,
\end{align*}
and
\begin{align*}
&  a_k  = \gamma_{k+1}\left(\upsilon_{\min} - \upsilon^2_{\max}\frac{\beta(n-1) + 1}{2n}\right) \eqsp, \\
& \Lambda_{(k+1)} =\frac{1}{n} - \gamma_{k+1} \beta - \frac{2\gamma_{k+1} \Lip{\bss}^2}{n^2}(\gamma_{k+1} +\frac{1}{\beta})\eqsp, \\
& c_1 = \upsilon_{\min}^{-1}, \alpha = \max\{8, 1+6\upsilon_{\min}\}, \overline{L} = \max\{ \Lip{\bss} , \Lip{V} \}, \gamma_{k+1} = \frac{1}{k \alpha c_1 \overline{L}}, \beta = \frac{c_1 \overline{L}}{n}\eqsp.
\end{align*}
When, for any $k >0$, $\tilde{\alpha}_k \geq 0$, we have by Lemma~\ref{lem:growth} that:
\beq\notag
\sum_{k=0}^{{ K}_{ m }} \tilde{\alpha}_k \EE [\| \grd V( \hs{k} )\|^2 ] \leq \upsilon_{\max}^2\sum_{k=0}^{{ K}_{ m }} \tilde{\alpha}_k \EE [\|  \overline{\bss}^{(k)}  - \hs{k}\|^2  ]  \eqsp,
\eeq
concluding the proof of the Theorem.
\end{proof}

\section{Proofs for the \SAEMVR\ and the \FISAEM\ Algorithms}
\subsection{Additional Intermediary Results}

We introduce additional Lemmas below before getting into the proofs of the desired results.
\begin{lemmacoloured}
Consider the \SAEMVR\ update~\eqref{eq:vrsaem} with $\rho_k = \rho$, it holds for all $k>0$
\beq\notag
\begin{split}
  \EE [\| \hs{k} - \stt^{(k+1)}\|^2 ] \leq& 2\rho^2 \EE[ \| \hs{k} - \os^{(k)} \|^2] +  2\rho^2\Lip{\bss}^2 \EE[ \| \hs{k} - \hs{\ell(k)} \|^2 ]\\
  &+2(1-\rho)^2 \EE[ \| \hs{(k)} - \stt^{(k)} \|^2 ]+ 2\rho^2\EE[\|\eta_{i_k}^{(k+1)} \|^2]\eqs,
\end{split}
\eeq
where we recall that $\ell(k)$ is the first iteration number in the epoch that iteration $k$ is in.
\end{lemmacoloured}
\begin{proof}
Beforehand, we provide an alternate expression of the quantity $ \hs{k+1} - \hs{k} $ that will be useful throughout this proof:
\beq\label{eq:vrsaem_drift}
\begin{split}
\hs{k+1} - \hs{k}  & = -\gamma_{k+1}  ( \hs{k} - \stt^{(k+1)}) \\
&=-\gamma_{k+1}  ( \hs{k} - (1-\rho)\stt^{(k)} - \rho\StocEstep^{(k+1)})\\
& = -\gamma_{k+1} \left((1-\rho)[\hs{k} - \stt^{(k)} ] +\rho[\hs{k} - \StocEstep^{(k+1)}] \right) \eqsp.
\end{split}
\eeq
We observe, using the identity~\eqref{eq:vrsaem_drift}, that
\beq \label{eq:auxlemvrsaem}
\begin{split}
\EE[ \| \hs{k} -\stt^{(k+1)} \|^2 ] \leq & 2\rho^2 \EE[ \| \hs{k} - \os^{(k)} \|^2] + 2\rho^2 \EE[ \| \os^{(k)} - \StocEstep^{(k+1)} \|^2 ] \\
&+ 2(1-\rho)^2 \EE[ \| \hs{(k)} - \stt^{(k)} \|^2 ].
\end{split}
\eeq
For the latter term, we obtain its upper bound as 
\beq\notag
\begin{split}
\EE[ \| \os^{(k)} - \StocEstep^{(k+1)} \|^2 ] = &\EE[ \| \frac{1}{n} \sum_{i=1}^n ( \os_i^{(k)} - \tilde{S}_i^{\ell(k)} ) - ( \os_{i_k}^{(k)} - \tilde{S}_{i_k}^{(\ell(k))} ) \|^2 ] \\
 \overset{(a)}{\leq} & \EE[ \| \os_{i_k}^{(k)} - \os_{i_k}^{(\ell(k))} \|^2 ] + \EE[\|\eta_{i_k}^{(k+1)} \|^2] \\
 & \overset{(b)}{\leq}  \Lip{\bss}^2 \EE[ \| \hs{k} - \hs{\ell(k)} \|^2 ]+ \EE[\|\eta_{i_k}^{(k+1)} \|^2]\eqsp,
\end{split}
\eeq
where $(a)$ uses the variance inequality and $(b)$ uses Lemma~\ref{lem:smooth}.
Substituting into~\eqref{eq:auxlemvrsaem} proves the lemma.
\end{proof}
\begin{lemmacoloured}
Consider the \FISAEM\ update~\eqref{eq:fisaem} with $\rho_k = \rho$. It holds for all $k>0$ that
\beq\notag
\begin{split}
  \EE [\| \hs{k} - \stt^{(k+1)}\|^2 ] \leq& 2\rho^2 \EE[ \| \hs{k} - \os^{(k)} \|^2] +  2\rho^2\frac{\Lip{\bss}^2}{n}
\sum_{i=1}^n \EE[ \| \hs{k} - \hs{t_i^k} \|^2 ]\\
  &+2(1-\rho)^2 \EE[ \| \hs{(k)} - \stt^{(k)} \|^2 ]+ 2\rho^2\EE[\|\eta_{i_k}^{(k+1)} \|^2]\eqsp,
\end{split}
\eeq
where $\Lip{\bss}$ is the smoothness constant defined in Lemma~\ref{lem:smooth}.
\end{lemmacoloured}

\begin{proof}
Beforehand, we provide a rewriting of the quantity $ \hs{k+1} - \hs{k} $ that will be useful throughout this proof:
\beq\label{eq:fisaem_drift}
\begin{split}
\hs{k+1} - \hs{k}  & = -\gamma_{k+1}  ( \hs{k} - \stt^{(k+1)}) \\
& =-\gamma_{k+1}  ( \hs{k} - (1-\rho)\stt^{(k)} - \rho\StocEstep^{(k+1)})\\
& = -\gamma_{k+1} \left((1-\rho)[\hs{k} - \stt^{(k)} ] +\rho[\hs{k} - \StocEstep^{(k+1)}] \right)\\
& =  -\gamma_{k+1} \left((1-\rho)[\hs{k} - \stt^{(k)} ] +\rho[\hs{k} - \overline{\StocEstep}^{(k)} - ( \tilde{S}_{i_k}^{(k)}  -  \tilde{S}_{i_k}^{(t_{i_k}^k)}  )] \right) \eqsp.
\end{split}
\eeq
We observe, using the identity~\eqref{eq:fisaem_drift}, that
{\small
\beq \label{eq:auxlemfisaem}
\EE[ \| \hs{k} -\stt^{(k+1)} \|^2 ] \leq 2\rho^2 \EE[ \| \hs{k} - \os^{(k)} \|^2] + 2\rho^2 \EE[ \| \os^{(k)} - \StocEstep^{(k+1)} \|^2 ]+ 2(1-\rho)^2 \EE[ \| \hs{(k)} - \stt^{(k)} \|^2 ]\eqsp.
\eeq}
For the latter term, we obtain its upper bound as 
\beq\notag
\begin{split}
\EE[ \| \os^{(k)} - \StocEstep^{(k+1)} \|^2 ] & = \EE[ \| \frac{1}{n} \sum_{i=1}^n ( \os_i^{(k)} -\overline{\StocEstep}_i^{(k)} ) - ( \tilde{S}_{i_k}^{(k)} - \tilde{S}_{i_k}^{(t_{i_k}^k)} ) \|^2 ] \\
& \overset{(a)}{\leq} \EE[ \| \os_{i_k}^{(k)} - \os_{i_k}^{(\ell(k))} \|^2 ] + \EE[\|\eta_{i_k}^{(k+1)} \|^2] \eqsp,
\end{split}
\eeq
where $(a)$ uses the variance inequality.
We can further bound the last expectation using Lemma~\ref{lem:smooth}:
\beq\notag
\EE[ \| \os_{i_k}^{(k)} - \os_{i_k}^{(t_{i_k}^k)} \|^2 ] = \frac{1}{n} \sum_{i=1}^n \EE[ \| \os_i^{(k)} - \os_i^{(t_i^k)} \|^2 ] \overset{(a)}{\leq} \frac{\Lip{\bss}^2}{n}
\sum_{i=1}^n \EE[ \| \hs{k} - \hs{t_i^k} \|^2 ]\eqsp.
\eeq
Substituting the above into~\eqref{eq:auxlemfisaem} proves the lemma.
\end{proof}

\begin{lemmacoloured}
Considering a decreasing stepsize $\gamma_k \in (0,1)$ and a constant $\rho \in (0,1)$, we have
\beq\notag
\begin{split}
\EE [\| \hs{k} - \stt^{(k)}   \|^2]  \leq \frac{\rho}{1-\rho}\sum_{\ell = 0}^k (1-\gamma_{\ell} )^2 (   \StocEstep^{(\ell)} - \tilde{S}^{(\ell)})\eqs,
\end{split}
\eeq
where $\StocEstep^{(k)}  $ is defined either by Line~2 (\SAEMVR ) or Line~3 (\FISAEM ).
\end{lemmacoloured}
\begin{proof}
We begin by writing the two-timescale update:
\beq\label{eq:updatetwo}
\begin{split}
& \stt^{(k+1)} = \stt^{(k)} + \rho ( \StocEstep^{(k+1)}- \stt^{(k)}  )\eqsp,\\
&  \hat{\bss}^{(k+1)} =  \hat{\bss}^{(k)}  + \gamma_{k+1}(\stt^{(k+1)} - \hat{\bss}^{(k)} ) \eqsp,
\end{split}
\eeq
where $\StocEstep^{(k+1)} = \frac{1}{n}\sum_{i=1}^n \tilde{S}_i^{(t_i^k)} + ( \tilde{S}_{i_k}^{(k)}  - \tilde{S}_{i_k}^{(t_{i_k}^k)} ) $ according to~\eqref{eq:fisaem}.
Denote $\delta^{(k+1)} =  \hs{k+1} - \stt^{(k+1)} $.
Then from~\eqref{eq:updatetwo}, doing the subtraction of both equations yields:
\beq\notag
\delta^{(k+1)} = (1-\gamma_{k+1} ) \delta^{(k)} + \frac{\rho}{1-\rho}(1-\gamma_{k+1} )(  \StocEstep^{(k+1)} -  \stt^{(k+1)})\eqsp.
\eeq
Using the telescoping sum and noting that $\delta^{(0)} = 0$, we have
\beq\notag
\delta^{(k+1)} \leq \frac{\rho}{1-\rho}\sum_{\ell = 0}^k (1-\gamma_{\ell+1} )^2 (   \StocEstep^{(\ell+1)} - \tilde{S}^{(\ell+1)} )\eqsp.
\eeq
\end{proof}

\subsection{Proofs of Auxiliary Lemmas ( Lemma~\ref{lem:auxvrsaem}, Lemma~\ref{lem:aux1} and Lemma~\ref{lem:gap_dynamics})} \label{app:bothauxvrsaem}
\begin{Lemma*}
 At iteration $k+1$,the drift term of update~\eqref{eq:fisaem}, with $\rho_{k+1} = \rho$, is equivalent to :
\beq\notag
\begin{split}
 \hs{k} -  \stt^{(k+1)}= & \rho (\hs{k} - \overline{\bss}^{(k)})  + \rho \eta_{i_k}^{(k+1)}+ \rho [(\overline{\bss}_{i_k}^{(k)} - \tilde{S}_{i_k}^{(t_{i_k}^k)}) - \EE[\overline{\bss}_{i_k}^{(k)} - \tilde{S}_{i_k}^{(t_{i_k}^k)}] ] \\
 &+ (1-\rho)\left( \hs{k} - \tilde{S}^{(k)}\right)\eqsp,
\end{split}
\eeq
where we recall that $\eta_{i_k}^{(k+1)}$, defined in~\eqref{eq:mcerror}, which is the gap between the MC approximation and the expected statistics.
\end{Lemma*}
\begin{proof}
Using the \FISAEM\ update $ \stt^{(k+1)} = (1 - \rho)\stt^{(k)} + \rho \StocEstep^{(k+1)}$ where $\StocEstep^{(k+1)} = \overline{\StocEstep}^{(k)} + ( \tilde{S}_{i_k}^{(k)}  - \tilde{S}_{i_k}^{(t_{i_k}^k)} )$ leads to the following decomposition:
{\small
\beq\notag
\begin{split}
 & \stt^{(k+1)} - \hs{k} \\
 =& (1 - \rho)\stt^{(k)} + \rho \left( \overline{\StocEstep}^{(k)} + ( \tilde{S}_{i_k}^{(k)}  - \tilde{S}_{i_k}^{(t_{i_k}^k)} ) \right) - \hs{k}+\rho \overline{\bss}^{(k)} - \rho \overline{\bss}^{(k)} \\
 =& \rho (\overline{\bss}^{(k)} - \hs{k}) + \rho(\tilde{S}_{i_k}^{(k)} - \overline{\bss}^{(k)}_{i_k}) + (1-\rho)\left(\stt^{(k)} - \hs{k}\right) + \rho \left( \overline{\StocEstep}^{(k)} - \overline{\bss}^{(k)}+ ( \overline{\bss}_{i_k}^{(k)}   - \tilde{S}_{i_k}^{(t_{i_k}^k)} ) \right)\\
 =& \rho (\overline{\bss}^{(k)}-\hs{k}) + \rho \eta_{i_k}^{(k+1)} - \rho [(\overline{\bss}_{i_k}^{(k)} - \tilde{S}_{i_k}^{(t_{i_k}^k)}) - \EE[\overline{\bss}_{i_k}^{(k)} - \tilde{S}_{i_k}^{(t_{i_k}^k)}] ] \\
 +& (1-\rho)\left(\stt^{(k)} - \hs{k}\right) \eqsp,
\end{split}
\eeq
}
where we observe that $\EE[\overline{\bss}_{i_k}^{(k)} - \tilde{S}_{i_k}^{(t_{i_k}^k)}] =\overline{\bss}^{(k)} -   \overline{\StocEstep}^{(k)} $ and which concludes the proof.

{Important Note:} Note that $\overline{\bss}_{i_k}^{(k)} - \tilde{S}_{i_k}^{(t_{i_k}^k)}$ is not equal to $\eta_{i_k}^{(k+1)}$, defined in~\eqref{eq:mcerror}, which is the gap between the MC approximation and the expected statistics. Indeed $\tilde{S}_{i_k}^{(t_{i_k}^k)}$ is not computed under the same model as $\overline{\bss}_{i_k}^{(k)}$.
\end{proof}

\subsection{Proof of Theorem~\ref{thm:vrsaem}}\label{app:theoremvrsaem}
\begin{Theorem*}
Assume A\ref{ass:compact}-A\ref{ass:mcerror}.
Consider the \SAEMVR\ sequence $\{\hat{\bss}^{(k)}\}_{k>0} \in \mathcal{S}$ for any $k \leq { K}_{ m }$ where ${ K}_{ m }$ is a positive integer.
Let $\{\gamma_{k+1} = 1/(k^a \overline{L})\}_{k>0}$, where $a \in (0,1)$, be a sequence of stepsizes, $\overline{L} = \max \{\Lip{\bss}, \Lip{V} \}$, $\rho = \mu/( c_1 \overline{L}  n^{2/3})$, $m = n c_1^2/(2 \mu^2+\mu c_1^2)$ and a constant $\mu \in (0,1)$. Then:
\beq\notag
\EE[ \| \grd V( \hs{K} ) \|^2 ] \leq  \frac{2 n^{2/3} \overline{L}}{\mu { P}_{ m} \upsilon_{\min}^2\upsilon_{\max}^2}\left( \EE[ \Delta V ]+  \sum_{k=0}^{{ K}_{ m }-1}  \tilde{\eta}^{(k+1)}\hspace{-0.1cm} + \chi^{(k+1)} \EE[\| \hs{k} - \tilde{S}^{(k)}\|^2]\right)  \eqsp.
\eeq
\end{Theorem*}

\begin{proof}

Using the smoothness of $V$ and update~\eqref{eq:vrsaem}, we obtain:
\beq\label{eq:smoothvrsaem}
\begin{split}
V( \hs{k+1} ) & \leq V( \hs{k} ) + \pscal{  \hs{k+1} - \hs{k}  }{ \grd V( \hs{k} ) } + \frac{ \Lip{V}}{2} \| \hs{k+1} - \hs{k} \|^2\\
& \leq V( \hs{k} ) - \gamma_{k+1} \pscal{  \hs{k}-  \stt^{(k+1)} }{ \grd V( \hs{k} ) } + \frac{\gamma_{k+1}^2 \Lip{V}}{2} \|  \hs{k}  - \stt^{(k+1)} \|^2\eqsp.
\end{split}
\eeq
Denote $\Hdrift_{k+1} \eqdef  \hs{k} -  \stt^{(k+1)} $ the drift term of the \FISAEM\ update in~\eqref{eq:rmstep} and  $\hmean_{k} =\hs{k} - \overline{\bss}^{(k)}$. Taking expectations on both sides show that
\beq \label{eq:lips_con}
\begin{split}
& \EE[ V( \hs{k+1} ) ] \\
 \overset{(a)}{\leq} &\EE[ V( \hs{k} ) ] - \gamma_{k+1}(1-\rho) \EE [ \pscal{ \hs{k} - \stt^{(k)} }{\grd V( \hs{k} ) } ]\\
 &- \gamma_{k+1} \rho \EE [ \pscal{ \hs{k} - \StocEstep^{(k+1)}  }{\grd V( \hs{k} ) } ]  +  \frac{\gamma_{k+1}^2 \Lip{V}}{2} \EE[ \| \Hdrift_{k+1} \|^2 ] \\
 \overset{(b)}{\leq}&  \EE[ V( \hs{k} ) ] - \gamma_{k+1} \rho \EE [ \pscal{ \hmean_{k}  }{\grd V( \hs{k} ) } ]- \gamma_{k+1}(1-\rho) \EE [ \pscal{ \hs{k} - \stt^{(k)} }{\grd V( \hs{k} ) } ] \\
  -&  \gamma_{k+1}\rho \EE [ \pscal{ \eta_{i_k}^{(k+1)} }{\grd V( \hs{k} ) } ] + \frac{\gamma_{k+1}^2 \Lip{V}}{2} \EE[ \| \Hdrift_{k+1} \|^2 ] \\
 \overset{(c)}{\leq}&  \EE[ V( \hs{k} ) ] - \left(\gamma_{k+1} \rho \upsilon_{\min} + \gamma_{k+1}  \upsilon_{\max}^2 \right)  \EE [ \norm{\hmean_{k}}^2 ]+ \frac{\gamma_{k+1}^2 \Lip{V}}{2} \EE[ \| \Hdrift_{k+1} \|^2 ]\\
 - &  \gamma_{k+1} \rho \EE[\norm{\eta_{i_k}^{(k+1)}}^2 ] - \gamma_{k+1}(1-\rho) \EE [ \| \hs{k} - \tilde{S}^{(k)}\|^2 ]  \eqsp,
\end{split}
\eeq
where we have used~\eqref{eq:vrsaem_drift} in $(a)$ and $\EE [ \StocEstep^{(k+1)} ] = \overline{\bss}^{(k)} + \EE[\eta_{i_k}^{(k+1)}]$ in $(b)$, the growth condition in Lemma~\ref{lem:growth} and Young's inequality with the constant equal to $1$ in $(c)$.
Furthermore, for $k+1 \leq \ell(k) + m$ (\ie $k+1$ is in the same epoch as $k$), we have
\beq\notag
\begin{split}
& \EE[ \| \hs{k+1} -  \hs{\ell(k)} \|^2 ] = \EE[ \| \hs{k+1} - \hs{k} + \hs{k} - \hs{\ell(k)} \|^2 ] \\
= & \EE [  \| \hs{k} -  \hs{\ell(k)} \|^2 + \| \hs{k+1} - \hs{k}  \|^2 + 2 \pscal{\hs{k} -  \hs{\ell(k)} }{\hs{k+1} - \hs{k} } ] \\
= &  \EE [ \| \hs{k} -  \hs{\ell(k)} \|^2 + \gamma_{k+1}^2 \| \Hdrift_{k+1} \|^2 \\
-&2\gamma_{k+1} \pscal{\hs{k} -  \hs{\ell(k)} }{ \rho(\hmean_{k} - \eta_{i_k}^{(k+1)}) + (1-\rho)( \hs{k} - \stt^{(k)} )  } ] \\
 \leq &\EE [ (1 + \gamma_{k+1} \beta) \| \hs{k} -  \hs{\ell(k)} \|^2 + \gamma_{k+1}^2 \| \Hdrift_{k+1} \|^2 + \frac{\gamma_{k+1}\rho}{\beta} \| \hmean_{k} \|^2\\
 +  & \frac{\gamma_{k+1}\rho}{\beta} \|\eta_{i_k}^{(k+1)} \|^2 + \frac{\gamma_{k+1}(1-\rho)}{\beta} \| \hs{k} - \stt^{(k)} \|^2 ]\eqsp,
\end{split}
\eeq
where we first used~\eqref{eq:vrsaem_drift} and the last inequality is due to Young's inequality.
Consider the following sequence:
\beq\notag
R_k \eqdef \EE[ V( \hs{k} ) + b_{{k}} \| \hs{k} - \hs{\ell(k)} \|^2 ]\eqsp,
\eeq
where $b_k \eqdef \overline{b}_{k~{\rm mod}~m}$ is a periodic sequence where:
\beq\notag
\overline{b}_i = \overline{b}_{i+1} (1 + \gamma_{k+1} \beta + 2 \gamma_{k+1}^2\rho^2 \Lip{\bss}^2 ) + \gamma_{k+1}^2\rho^2 \Lip{V} \Lip{\bss}^2,~~i=0,1,\dots,m-1~~\text{with}~~\overline{b}_m = 0\eqsp.
\eeq
Note that $\overline{b}_i$ is decreasing with $i$ and this implies
\beq\notag
\overline{b}_i \leq \overline{b}_0 = \gamma_{k+1}^2\rho^2 \Lip{V} \Lip{\bss}^2 \frac{ (1 + \gamma_{k+1} \beta + 2 \gamma_{k+1}^2 \rho^2\Lip{\bss}^2 )^m - 1 }{ \gamma_{k+1} \beta + 2 \gamma_{k+1}^2 \rho^2\Lip{\bss}^2 },~i=1,2,\dots,m \eqsp.
\eeq
For $k+1 \leq \ell(k) + m$, we have the following inequality
\beq\notag
\begin{split}
R_{k+1 } & \leq  \EE [ V( \hs{k} )  - \left(\gamma_{k+1} \rho \upsilon_{\min} + \gamma_{k+1}  \upsilon_{\max}^2 \right)  \| \hmean_{k} \|^2 + \frac{\gamma_{k+1}^2 \Lip{V}}{2} \| \Hdrift_{k+1}  \|^2 ] \\
& + \gamma_{k+1} \EE[\rho \norm{\eta_{i_k}^{(k+1)}}^2 -(1-\rho)\| \hs{k} - \tilde{S}^{(k)}\|^2 ]\\
& + b_{k+1} \EE [ (1 + \gamma_{k+1} \beta) \| \hs{k} -  \hs{\ell(k)} \|^2 + \gamma_{k+1}^2 \| \Hdrift_{k+1} \|^2 + \frac{\gamma_{k+1}\rho}{\beta} \| \hmean_{k} \|^2 ]\\
& + b_{k+1} \EE [ \frac{\gamma_{k+1}\rho}{\beta} \|\eta_{i_k}^{(k+1)} \|^2 + \frac{\gamma_{k+1}(1-\rho)}{\beta} \| \hs{k} - \stt^{(k)} \|^2 ]\eqsp.
\end{split}
\eeq
And using Lemma~\ref{lem:auxvrsaem} we obtain:
\beq\notag
\begin{split}
& R_{k+1 }  \\
\leq & \EE [ V( \hs{k} )  - \left(\gamma_{k+1} \rho \upsilon_{\min} + \gamma_{k+1}  \upsilon_{\max}^2  - \gamma_{k+1}^2\rho^2 \Lip{V}\right)  \| \hmean_{k} \|^2  \\
&+ \gamma_{k+1}^2\rho^2 \Lip{V} \Lip{\bss}^2 \| \hs{k} - \hs{\ell(k)} \|^2 ] \\
& + b_{k+1} \EE [ (1 + \gamma_{k+1} \beta + 2\gamma_{k+1}^2 \rho^2 \Lip{\bss}^2) \| \hs{k} -  \hs{\ell(k)} \|^2  + (\frac{\gamma_{k+1}\rho}{\beta}+ 2\gamma_{k+1}^2 \rho^2) \| \hmean_{k} \|^2 ]\\
& + \gamma_{k+1} \EE[(\rho+\rho^2\gamma_{k+1}\Lip{V}) \norm{\eta_{i_k}^{(k+1)}}^2 -(1-\rho - (1-\rho)^2\gamma_{k+1}\Lip{V})\| \hs{k} - \tilde{S}^{(k)}\|^2 ]\\
& + b_{k+1} \EE [ (\frac{\gamma_{k+1}\rho}{\beta}+ 2\gamma_{k+1}^2 \rho^2) \|\eta_{i_k}^{(k+1)} \|^2 \\
&+ (\frac{\gamma_{k+1}(1-\rho)}{\beta}+ 2\gamma_{k+1}^2 (1-\rho)^2) \| \hs{k} - \stt^{(k)} \|^2 ]\eqsp.
\end{split}
\eeq
Rearranging the terms yields:
\beq\notag
\begin{split}
R_{k+1 } & \leq
\EE [ V( \hs{k} ) ] - \gamma_{k+1}(  \rho \upsilon_{\min} +   \upsilon_{\max}^2  - \gamma_{k+1}\rho^2 \Lip{V} - b_{k+1}(\frac{\rho}{\beta}+ 2\gamma_{k+1} \rho^2) ) \EE[ \|  \hmean_{k} \|^2 ] \\
& + (  \underbrace{b_{k+1} (1 + \gamma \beta + 2 \gamma^2\rho^2 \Lip{\bss}^2 ) + \gamma^2\rho^2 \Lip{V} \Lip{\bss}^2}_{=b_k~~\text{since $k+1 \leq \ell(k)+m$}} ) \EE[  \| \hs{k} - \hs{\ell(k)} \|^2 ]+ \tilde{\eta}^{(k+1)} + \tilde{\chi}^{(k+1)}\eqsp,
\end{split}
\eeq
where
\beq\notag
\begin{split}
&  \tilde{\eta}^{(k+1)}  = \left( \gamma_{k+1}(\rho+\rho^2\gamma_{k+1}\Lip{V}) + b_{k+1} (\frac{\gamma_{k+1}\rho}{\beta}+ 2\gamma_{k+1}^2 \rho^2) \right) \EE[ \norm{\eta_{i_k}^{(k+1)}}^2 ]\\
& \chi^{(k+1)} = \left( b_{k+1} (\frac{\gamma_{k+1}(1-\rho)}{\beta}+ 2\gamma_{k+1}^2 (1-\rho)^2) - \gamma_{k+1}(1-\rho - (1-\rho)^2\gamma_{k+1}\Lip{V}) \right) \\
& \tilde{\chi}^{(k+1)} = \chi^{(k+1)} \EE[\| \hs{k} - \stt^{(k)} \|^2 ]\eqsp.
\end{split}
\eeq
This leads, using Lemma~\ref{lem:growth}, that for any $\gamma_{k+1}$, $\rho$ and $\beta$ such that $  \rho \upsilon_{\min} +   \upsilon_{\max}^2  - \gamma_{k+1}\rho^2 \Lip{V} - b_{k+1}(\frac{\rho}{\beta}+ 2\gamma_{k+1} \rho^2)  >0$,
\beq\notag
\begin{split}
& \upsilon_{\max}^2 \EE[ \| \grd V( \hs{k} ) \|^2 ]  \leq \EE[ \| \hs{k} - \os^{(k)} \|^2 ] \\
\leq & \frac{  R_{k} - R_{k+1} }{ \gamma_{k+1}(  \rho \upsilon_{\min} +   \upsilon_{\max}^2  - \gamma_{k+1}\rho^2 \Lip{V} - b_{k+1}(\frac{\rho}{\beta}+ 2\gamma_{k+1} \rho^2) )}\\
& +\frac{ \tilde{\eta}^{(k+1)} + \tilde{\chi}^{(k+1)} }{ \gamma_{k+1}(  \rho \upsilon_{\min} +   \upsilon_{\max}^2  - \gamma_{k+1}\rho^2 \Lip{V} - b_{k+1}(\frac{\rho}{\beta}+ 2\gamma_{k+1} \rho^2) )} \eqsp.
\end{split}
\eeq
We first remark that
\beq\notag
\begin{split}
&\gamma_{k+1}(  \rho \upsilon_{\min} +   \upsilon_{\max}^2  - \gamma_{k+1}\rho^2 \Lip{V} - b_{k+1}(\frac{\rho}{\beta}+ 2\gamma_{k+1} \rho^2) ) \\
&\geq  \frac{\gamma_{k+1} \rho}{c_1}(1  - \gamma_{k+1}c_1\rho \Lip{V} - b_{k+1}(\frac{c_1}{\beta}+ 2\gamma_{k+1} \rho c_1) )\eqsp,
\end{split}
\eeq
where $c_1 = \upsilon_{\min}^{-1}$.
By setting $\overline{L} = \max \{\Lip{\bss}, \Lip{V} \}$, $\beta = \frac{c_1 \overline{L}}{n^{1/3}}$, $\rho = \frac{\mu}{ c_1 \overline{L}  n^{2/3}}$, $m = \frac{n c_1^2}{2 \mu^2+\mu c_1^2}$ and $\{ \gamma_{k+1}\}$ any sequence of decreasing stepsizes in $(0,1)$, it can be shown that there exists $\mu \in (0,1)$, such that the following lower bound holds
\beq\notag
\begin{split}
& 1  - \gamma_{k+1}c_1\rho \Lip{V} - b_{k+1}(\frac{c_1}{\beta}+ 2\gamma_{k+1} \rho c_1)
\\
 \geq & 1 - \frac{\mu}{n^{\frac{2}{3}}} - \overline{b}_0 ( \frac{n^{\frac{1}{3}}}{\overline{L}} + \frac{2 \mu}{\overline{L} n^{\frac{2}{3}}} ) \\
 \geq & 1 - \frac{\mu }{n^{\frac{2}{3}}} - \frac{ \Lip{V} \mu^2 }{c_1^2 n^{\frac{4}{3}}} \frac{ (1 + \gamma \beta + 2 \gamma^2 \Lip{\bss}^2 )^m - 1 }{ \gamma \beta + 2 \gamma^2 \Lip{\bss}^2 } ( \frac{n^{\frac{1}{3}}}{\overline{L}} + \frac{2 \mu}{\overline{L} n^{\frac{2}{3}}} ) \\
  \overset{(a)}{\geq} &1 - \frac{\mu}{ n^{\frac{2}{3}}} - \frac{ \mu }{c_1^2 } (\rme-1) ( 1 + \frac{2 \mu}{n} )
 \geq 1 - \mu - \mu(1+2 \mu) \frac{\rme-1}{c_1^2} \overset{(b)}{ \geq} \frac{1}{2}\eqsp,
 \end{split}
\eeq
where the simplification in (a) is due to
\beq\notag
\frac{\mu}{n} \leq \gamma \beta + 2 \gamma^2 \Lip{\bss}^2 \leq \frac{\mu}{n} + \frac{2 \mu^2}{c_1^2 n^{\frac{4}{3}}} \leq \frac{\mu c_1^2 + 2 \mu^2}{c_1^2} \frac{1}{n}~~\text{and}~~(1 + \gamma \beta + 2 \gamma^2 \Lip{\bss}^2 )^m \leq \rme-1 \eqsp,
\eeq
and the required $\mu$ in (b) can be found by solving the quadratic equation.
Finally, these results yield:
\beq\notag
\begin{split}
\upsilon_{\max}^2 \sum_{k=0}^{{ K}_{ m }-1}\gamma_{k+1} \EE[ \| \grd V( \hs{k} ) \|^2 ]  \leq  \frac{2(R_0 - R_{{ K}_{ m }})}{ \upsilon_{\min} \rho} + 2\sum_{k=0}^{{ K}_{ m }-1}  \frac{ \tilde{\eta}^{(k+1)} + \tilde{\chi}^{(k+1)}}{ \upsilon_{\min} \rho}\eqsp.
 \end{split}
\eeq
Note that $R_0 = \EE[ V( \hs{0} ) ]$ and if ${ K}_{ m }$ is a multiple of $m$, then $R_{ max} = \EE[ V( \hs{{ K}_{ m }}) ]$. Under the latter condition, we have
\beq\notag
\begin{split}
 \sum_{k=0}^{{ K}_{ m }-1}\gamma_{k+1} \EE[ \| \grd V( \hs{k} ) \|^2 ] \leq &  \frac{2 n^{2/3} \overline{L}}{\mu \upsilon_{\min}^2\upsilon_{\max}^2}\EE[ V( \hs{0} ) - V( \hs{{ K}_{ m }}) ]  \\
 &+ \frac{2 n^{2/3} \overline{L}}{\mu \upsilon_{\min}^2\upsilon_{\max}^2} \sum_{k=0}^{{ K}_{ m }-1} [  \tilde{\eta}^{(k+1)} + \tilde{\chi}^{(k+1)}]\eqsp.
\end{split}
\eeq
This concludes our proof.

\end{proof}

\subsection{Proof of Theorem~\ref{thm:fisaem}}\label{app:theoremfisaem}
\begin{Theorem*}
Assume A\ref{ass:compact}-A\ref{ass:mcerror}.
Consider the \FISAEM\ sequence $\{\hat{\bss}^{(k)}\}_{k>0} \in \mathcal{S}$ for any $k \leq { K}_{ m }$ where ${ K}_{ m }$ be a positive integer.
Let $\{\gamma_{k+1} = 1/(k^a \alpha c_1 \overline{L}) \}_{k>0}$, where $a \in (0,1)$, be a sequence of positive stepsizes, $\alpha =\max\{2, 1+2\upsilon_{\min}\}$, $\overline{L} = \max\{ \Lip{\bss} , \Lip{V} \}$, $\beta = 1/(\alpha n)$, $\rho = 1/(\alpha c_1 \overline{L}n^{2/3})$ and $c_1(k\alpha-1) \geq c_1(\alpha-1) \geq 2$. Then:
\beq\notag
 \EE[ \| \grd V( \hs{K} ) \|^2 ] \leq \frac{4\alpha  \overline{L} n^{2/3}}{{ P}_{ m}\upsilon_{\min}^2\upsilon_{\max}^2} \left( \EE [ \Delta V ]   + \sum_{k=0}^{{ K}_{ m }-1}  \Xi^{(k+1)}  +\Gamma^{(k+1)} \EE [\| \hs{k} - \tilde{S}^{(k)}\|^2 ]\right)\eqs.
\eeq
\end{Theorem*}

\begin{proof}
Using the smoothness of $V$ and update~\eqref{eq:fisaem}, we obtain:
\beq\label{eq:smoothfisaem}
\begin{split}
V( \hs{k+1} ) & \leq V( \hs{k} ) + \pscal{  \hs{k+1} - \hs{k}  }{ \grd V( \hs{k} ) } + \frac{ \Lip{V}}{2} \| \hs{k+1} - \hs{k} \|^2\\
& \leq V( \hs{k} ) - \gamma_{k+1} \pscal{  \hs{k} - \stt^{(k+1)} }{ \grd V( \hs{k} ) } + \frac{\gamma_{k+1}^2 \Lip{V}}{2} \| \hs{k}  -  \stt^{(k+1)}\|^2\eqsp.
\end{split}
\eeq
Denote $\Hdrift_{k+1} \eqdef   \hs{k} - \stt^{(k+1)} $ the drift term of the \FISAEM\ update in~\eqref{eq:rmstep} and  $\hmean_{k} = \hs{k} - \overline{\bss}^{(k)}$. Using Lemma~\ref{lem:drift_fisaem} and the additional following identity:
\beq
\EE[(\overline{\bss}_{i_k}^{(k)} - \tilde{S}_{i_k}^{(t_{i_k}^k)}) - \EE[\overline{\bss}_{i_k}^{(k)} - \tilde{S}_{i_k}^{(t_{i_k}^k)}] ] = 0\eqsp,
\eeq
 we have
 \beq\notag
\begin{split}
 \EE[V( \hs{k+1} )]   \leq & \EE[ V( \hs{k} )] - \gamma_{k+1}\rho \EE[\pscal{ \hmean_{k}  }{ \grd V( \hs{k} ) }] +\frac{\gamma_{k+1}^2 \Lip{V}}{2} \| \Hdrift_{k+1}\|^2\\
 & -\gamma_{k+1}  \EE[\pscal{ \rho \EE[\eta_{i_k}^{(k+1)} |{\cal F}_k] + (1-\rho) \EE[ \hs{k} - \tilde{S}^{(k)}]}{ \grd V( \hs{k} ) }]\\
 \overset{(a)}{\leq} & -\upsilon_{\min}\gamma_{k+1}\rho \EE[\norm{\hmean_{k}}^2 ]  -\gamma_{k+1}\EE[\norm{\grd V( \hs{k} ) }^2 ] \\
 &-\frac{\gamma_{k+1}\rho^2}{2} \xi^{(k+1)} - \frac{\gamma_{k+1}(1-\rho)^2}{2} \EE[\| \hs{k} - \tilde{S}^{(k)}\|^2]+ \frac{\gamma_{k+1}^2 \Lip{V}}{2} \|\Hdrift_{k+1}\|^2\\
 \overset{(b)}{\leq}&  -(\upsilon_{\min}\gamma_{k+1}\rho+\gamma_{k+1} \upsilon_{\max}^2) \EE[\norm{\hmean_{k}}^2 ] \\
 &-\frac{\gamma_{k+1}\rho^2}{2} \xi^{(k+1)} - \frac{\gamma_{k+1}(1-\rho)^2}{2} \EE[\| \hs{k} - \tilde{S}^{(k)}\|^2] + \frac{\gamma_{k+1}^2 \Lip{V}}{2} \| \Hdrift_{k+1}\|^2\eqsp,
\end{split}
\eeq
where $\xi^{(k+1)}  \eqdef \EE[\|\EE[\eta_{i_k}^{(k+1)}|{\cal F}_k]  \|^2 ]$.
Next, we bound the quantity $\EE[\|  \Hdrift_{k+1}  \|^2]$.
Using Lemma~\ref{lem:aux1}, we obtain
\beq\label{eq:finalfisaem}
\begin{split}
& \gamma_{k+1}(\upsilon_{\min}\rho+\upsilon_{\max}^2 - \gamma_{k+1}\rho^2 \Lip{V})  \EE[\norm{\hmean_{k}}^2 ]\\
\leq &  \EE[V( \hs{k} ) - V( \hs{k+1} ) ] +\tilde{\xi}^{(k+1)} + \frac{ \gamma_{k+1}^2\Lip{V}\rho^2\Lip{\bss}^2}{n} \sum_{i=1}^n \EE[ \| \hs{k} - \hs{t_i^k} \|^2 ]\\
&+ \left( (1-\rho)^2 \gamma_{k+1}^2 \Lip{V} - \frac{\gamma_{k+1}(1-\rho)^2}{2} \right)  \EE[\| \hs{k} - \tilde{S}^{(k)}\|^2]\eqsp,
\end{split}
\eeq
where $ \tilde{\xi}^{(k+1)} =  \gamma_{k+1}^2 \rho^2 \Lip{V}\EE[\|\eta_{i_k}^{(k+1)} \|^2] - \frac{\gamma_{k+1}\rho^2}{2} \xi^{(k+1)}$.
Next, we observe that
\beq\label{eq:auxdelta}
\frac{1}{n} \sum_{i=1}^n \EE[ \| \hs{k+1} - \hs{t_i^{k+1}} \|^2 ] = \frac{1}{n} \sum_{i=1}^n
( \frac{1}{n} \EE[ \| \hs{k+1} - \hs{k} \|^2 ] + \frac{n-1}{n} \EE[ \| \hs{k+1} - \hs{t_i^k} \|^2 ]  )\eqsp,
\eeq
where the equality holds as $i_k$ and $j_k$ are drawn independently.
Then,
\beq\notag
\begin{split}
 \EE[ \| \hs{k+1} - \hs{t_i^k} \|^2 ] = \EE [ \| \hs{k+1} - \hs{k} \|^2 + \| \hs{k} - \hs{t_i^k} \|^2 + 2 \pscal{\hs{k+1} - \hs{k}}{\hs{k}- \hs{t_i^k}} ]\eqsp.
\end{split}
\eeq
Note that $\hs{k+1} - \hs{k} = -\gamma_{k+1} ( \hs{k} - \stt^{(k+1)}) = -\gamma_{k+1} \Hdrift_{k+1}$ and that in expectation we recall that $\EE[\Hdrift_{k+1}|{\cal F}_k] =  \rho \hmean_{k} + \rho\EE[\eta_{i_k}^{(k+1)}|{\cal F}_k] + (1-\rho) \EE[\stt^{(k)} - \hs{k}]$ where $\hmean_{k} = \hs{k} - \overline{\bss}^{(k)}$.
Thus, for any $\beta > 0$, it holds
\beq\notag
\begin{split}
& \EE[ \| \hs{k+1} - \hs{t_i^k} \|^2 ] =   \EE [ \| \hs{k+1} - \hs{k} \|^2 + \| \hs{k} - \hs{t_i^k} \|^2 + 2 \pscal{\hs{k+1} - \hs{k}}{\hs{k}- \hs{t_i^k}} ]\\
 \leq  & \EE [ \| \hs{k+1} - \hs{k} \|^2 + (1+ \gamma_{k+1} \beta) \| \hs{k} - \hs{t_i^k} \|^2 +  \frac{\gamma_{k+1} \rho^2}{\beta} \| \hmean_{k} \|^2 +  \frac{\gamma_{k+1} \rho^2}{\beta} \EE[\norm{\eta_{i_k}^{(k+1)}}^2 ]\\
&+ \frac{\gamma_{k+1}(1- \rho)^2}{\beta}  \EE[\| \hs{k} - \tilde{S}^{(k)}\|^2 ]]\eqsp,
\end{split}
\eeq
where the last inequality is due to Young's inequality.
Plugging this into~\eqref{eq:auxdelta} yields:
\beq\notag
\begin{split}
& \EE[ \| \hs{k+1} - \hs{t_i^k} \|^2 ] \\
 = & \EE [ \| \hs{k+1} - \hs{k} \|^2 + \| \hs{k} - \hs{t_i^k} \|^2 + 2 \pscal{\hs{k+1} - \hs{k}}{\hs{k}- \hs{t_i^k}} ]\\
 \leq &  \EE [ \| \hs{k+1} - \hs{k} \|^2 + (1+ \gamma_{k+1} \beta) \| \hs{k} - \hs{t_i^k} \|^2 +  \frac{\gamma_{k+1} \rho^2}{\beta} \| \hmean_{k} \|^2 \\
&+  \frac{\gamma_{k+1} \rho^2}{\beta} \EE[\norm{\eta_{i_k}^{(k+1)}}^2 ]+  \frac{\gamma_{k+1}(1- \rho)^2}{\beta}  \EE[\norm{\hs{k} - \tilde{S}^{(k)}}^2 ]]\eqsp.
\end{split}
\eeq
Subsequently, we have
\beq\notag
\begin{split}
& \frac{1}{n} \sum_{i=1}^n \EE[ \| \hs{k+1} - \hs{t_i^{k+1}} \|^2 ] \\
\leq&  \EE[  \| \hs{k+1} - \hs{k} \|^2 ] + \frac{n-1}{n^2} \sum_{i=1}^n \EE [(1+ \gamma_{k+1} \beta) \| \hs{k} - \hs{t_i^k} \|^2 +  \frac{\gamma_{k+1} \rho^2}{\beta} \| \hmean_{k} \|^2 \\
&+  \frac{\gamma_{k+1} \rho^2}{\beta} \EE[\|\eta_{i_k}^{(k+1)}\|^2 ]
  + \frac{\gamma_{k+1}(1- \rho)^2}{\beta}  \EE[\|\hs{k} - \tilde{S}^{(k)}|^2 ]]\eqsp.
\end{split}
\eeq
We now use Lemma~\ref{lem:aux1} on $\| \hs{k+1} - \hs{k} \|^2 = \gamma_{k+1}^2\|  \hs{k} - \stt^{(k+1)} \|^2$ and obtain:
\beq\notag
\begin{split}
&  \frac{1}{n} \sum_{i=1}^n \EE[ \| \hs{k+1} - \hs{t_i^{k+1}} \|^2 ]\\
 \leq &  \left(2 \gamma_{k+1}^2 \rho^2 + \frac{\gamma_{k+1} \rho^2}{\beta}\right) \EE[\| \overline{\bss}^{(k)}-\hs{k}\|^2 ]  \\
 &+ \sum_{i=1}^n \left( \frac{\gamma_{k+1}^2\rho^2 \Lip{\bss}^2}{n} + \frac{(n-1) (1+ \gamma_{k+1} \beta)}{n^2}  \right) \EE [ \| \hs{k} - \hs{t_i^k} \|^2 ]\\
 &+  \gamma_{k+1} (1-\rho)^2 \left( 2\gamma_{k+1} + \frac{1}{\beta} \right)\EE[ \|\hs{k} - \tilde{S}^{(k)}\|^2] + \left(2 \gamma_{k+1}^2 + \frac{\gamma_{k+1} \rho^2}{\beta} \right)\EE[\norm{\eta_{i_k}^{(k+1)}}^2 ]\\
 \leq &  \left(2 \gamma_{k+1}^2 \rho^2 + \frac{\gamma_{k+1} \rho^2}{\beta}\right) \EE[\| \overline{\bss}^{(k)}-\hs{k}\|^2 ]  \\
 &+ \sum_{i=1}^n \left( \frac{ 1 - \frac{1}{n} +\gamma_{k+1}\beta+\gamma_{k+1}^2\rho^2 \Lip{\bss}^2 }{n}   \right) \EE [ \| \hs{k} - \hs{t_i^k} \|^2 ]\\
&+  \gamma_{k+1} (1-\rho)^2 \left( 2\gamma_{k+1} + \frac{1}{\beta} \right)\EE[ \|\hs{k} - \tilde{S}^{(k)}\|^2] + \left(2 \gamma_{k+1}^2 + \frac{\gamma_{k+1} \rho^2}{\beta} \right)\EE[\norm{\eta_{i_k}^{(k+1)}}^2 ]\eqsp.
\end{split}
\eeq
Let us define
\beq\notag
\Delta^{(k)} \eqdef \frac{1}{n} \sum_{i=1}^n \EE[ \| \hs{k} - \hs{t_i^{k}} \|^2 ]\eqsp.
\eeq
From the above, we obtain
\beq\notag
\begin{split}
& \Delta^{(k+1)} \\
\leq & \left( 1 - \frac{1}{n} +\gamma_{k+1}\beta+\gamma_{k+1}^2\rho^2 \Lip{\bss}^2\right) \Delta^{(k)} + \left(2 \gamma_{k+1}^2 \rho^2 + \frac{\gamma_{k+1} \rho^2}{\beta}\right) \EE[\| \overline{\bss}^{(k)}-\hs{k}\|^2 ]\\
& + \gamma_{k+1} (1-\rho)^2 \left( 2\gamma_{k+1} + \frac{1}{\beta} \right)\EE[ \|\hs{k} - \tilde{S}^{(k)}\|^2] + \gamma_{k+1}\left(2 \gamma_{k+1} + \frac{ \rho^2}{\beta} \right)\EE[\norm{\eta_{i_k}^{(k+1)}}^2 ]\eqsp.
 \end{split}
\eeq

Setting $c_1 = \upsilon_{\min}^{-1}$, $\alpha =\max\{2, 1+2\upsilon_{\min}\}$, $\overline{L} = \max\{ \Lip{\bss} , \Lip{V} \}$, $\gamma_{k+1} = \frac{1}{k }$, $\beta = \frac{1}{\alpha n}$, $\rho = \frac{1}{\alpha c_1 \overline{L}n^{2/3}}$, then we have that $c_1(k\alpha-1) \geq c_1(\alpha-1) =
\max\{\frac{1}{\upsilon_{\min}}, 2\}
\geq 2$.
Hence, we observe that
\beq\notag
1 - \frac{1}{n} +\gamma_{k+1}\beta+\gamma_{k+1}^2\rho^2 \Lip{\bss}^2
 \leq 1 - \frac{1}{n} + \frac{1}{\alpha kn} + \frac{ 1 }{ \alpha^2 c_1^2 k^2 n^{\frac{4}{3}} } \leq 1 - \frac{c_1(k\alpha  - 1) - 1}{k\alpha n c_1 } \leq 1 - \frac{1}{k\alpha n c_1 },
\eeq
which shows that $1 - \frac{1}{n} +\gamma_{k+1}\beta+\gamma_{k+1}^2\rho^2 \Lip{\bss}^2  \in (0,1)$ for any $k >0$.
Denote $ \Lambda_{(k+1)} =\frac{1}{n} -\gamma_{k+1}\beta-\gamma_{k+1}^2\rho^2 \Lip{\bss}^2 $ and note that $\Delta^{(0)} = 0$, thus the telescoping sum yields
\beq\notag
\begin{split}
\Delta^{(k+1)} \leq & \sum_{ \ell = 0 }^k \omega_{k, \ell} \left(2 \gamma_{\ell+1}^2 \rho^2 + \frac{\gamma_{\ell+1}^2 \rho^2}{\beta}\right)  \EE[\norm{\overline{\bss}^{(\ell)}-\hs{\ell}}^2 ]\\
& +\sum_{ \ell = 0 }^k \omega_{k, \ell} \gamma_{\ell+1} (1-\rho)^2 \left( 2\gamma_{\ell+1} +\frac{1}{\beta} \right)\EE[ \norm{\tilde{S}^{(\ell)} - \hs{\ell}}^2] + \sum_{ \ell = 0 }^k \omega_{k, \ell}\gamma_{\ell+1} \tilde{\epsilon}^{(\ell+1)}  \eqsp,
\end{split}
\eeq
where $ \omega_{k, \ell} =  \prod_{j = \ell +1}^k ( 1 -  \Lambda_{(j)} )$ and $\tilde{\epsilon}^{(\ell+1)}   = \left(2 \gamma_{k+1} + \frac{ \rho^2}{\beta} \right)\EE[\norm{\eta_{i_k}^{(k+1)}}^2 ]$.

Summing on both sides over $k=0$ to $k = { K}_{ m }-1$ yields:
\beq\notag
\begin{split}
\sum_{k=0}^{{ K}_{ m }-1} \Delta^{(k+1)} & \leq \sum_{k=0}^{{ K}_{ m }-1}  \frac{2 \gamma_{k+1}^2 \rho^2 + \frac{\gamma_{k+1} \rho^2}{\beta}}{\Lambda_{(k+1)}}  \EE[\| \overline{\bss}^{(k)}-\hs{k}\|^2 ]\\
&+\sum_{k=0}^{{ K}_{ m }-1} \frac{\gamma_{k+1} (1-\rho)^2 \left( 2\gamma_{k+1} +\frac{1}{\beta} \right)}{\Lambda_{(k+1)}}\EE[ \|\hs{k} - \tilde{S}^{(k)}\|^2] + \sum_{k=0}^{{ K}_{ m }-1} \frac{\gamma_{k+1}}{\Lambda_{(k+1)}} \tilde{\epsilon}^{(k+1)}  \eqsp.
\end{split}
\eeq
We recall~\eqref{eq:finalfisaem} where we have summed on both sides from $k=0$ to $k = { K}_{ m }-1$:
\beq\label{eq:finalboundfi}
\begin{split}
& \EE [ V(\hat{\bss}^{({ K}_{ m })}) - V(\hat{\bss}^{(0)} ) ] \\
 \leq &   \sum_{k=0}^{{ K}_{ m }-1} \Big\{ \gamma_{k+1}( -(\upsilon_{\min}\rho+\upsilon_{\max}^2) + \gamma_{k+1}\rho^2 \Lip{V})  \EE[\norm{\hmean_{k}}^2 ]   + \gamma^2 \Lip{V}\rho^2 \Lip{\bss}^2 \Delta^{(k)}\Big\}\\
& +   \sum_{k=0}^{{ K}_{ m }-1} \Big\{ \tilde{\xi}^{(k+1)} + \left( (1-\rho)^2 \gamma_{k+1}^2 \Lip{V} - \frac{\gamma_{k+1}(1-\rho)^2}{2} \right)  \EE[\| \hs{k} - \tilde{S}^{(k)}\|^2]\Big\}\\
 \leq &  \sum_{k=0}^{{ K}_{ m }-1} \Big\{ \gamma_{k+1}\left[ -(\upsilon_{\min}\rho+\upsilon_{\max}^2) + \gamma_{k+1}\rho^2 \Lip{V} + \frac{\rho^2\gamma_{k+1} \Lip{V}\Lip{\bss}^2\left(2 \gamma_{k+1}^2 \rho^2 + \frac{\gamma_{k+1} \rho^2}{\beta}\right)}{\Lambda_{(k+1)}} \right ] \EE[\norm{\hmean_{k}}^2 ]\Big\}\\
  &+   \sum_{k=0}^{{ K}_{ m }-1} \Xi^{(k+1)}  +  \sum_{k=0}^{{ K}_{ m }-1}\Gamma^{(k+1)} \EE[\| \hs{k} - \tilde{S}^{(k)}\|^2]\eqsp,
\end{split}
\eeq
where
$$
\Xi^{(k+1)} =\tilde{\xi}^{(k+1)} +\frac{\gamma_{k+1}^3\Lip{V}\rho^2\Lip{\bss}^2}{\Lambda_{(k+1)}} \tilde{\epsilon}^{(k+1)}
$$
and
$$
\Gamma^{(k+1)} =  \left( (1-\rho)^2 \gamma_{k+1}^2 \Lip{V} - \frac{\gamma_{k+1}(1-\rho)^2}{2} \right)  +\frac{\gamma_{k+1}^3\Lip{V}\rho^2\Lip{\bss}^2 (1-\rho)^2 \left( 2\gamma_{k+1} +\frac{1}{\beta} \right)}{\Lambda_{(k+1)}}  \eqsp.
$$
Furthermore, given the values set for $c_1$, $\alpha$, $\overline{L}$, $\gamma_{k+1}$, $\beta$ and $\rho$, then
\beq\label{eq:stepsizeineq}
\begin{split}
& \gamma_{k+1}\rho^2 \Lip{V} + \frac{\rho^2\gamma_{k+1} \Lip{V}\Lip{\bss}^2\left(2 \gamma_{k+1}^2 \rho^2 + \frac{\gamma_{k+1} \rho^2}{\beta}\right)}{\frac{1}{n} -\gamma_{k+1}\beta-\gamma_{k+1}^2\rho^2 \Lip{\bss}^2} \\
 \leq & \frac{1}{k \alpha^2 c_1^2 \overline{L} n^{4/3}} + \frac{\overline{L} (k\alpha^2 c_1^2  n^{4/3})^{-1} ( \frac{2}{k^2 \alpha^2 c_1^2 \overline{L}^2 n^{4/3}} + \frac{1}{k \alpha c_1^2 \overline{L}^2 n^{1/3}} ) }{\frac{1}{n} - \frac{1}{k \alpha n} - \frac{1}{k^2 \alpha^2 c_1^2 n^{4/3}} }\\
 = & \frac{1}{k \alpha^2 c_1^2 \overline{L} n^{4/3}} + \frac{  \overline{L}( \frac{2}{k^2 \alpha^2 c_1^2 \overline{L}^2 n^{4/3}} + \frac{1}{k \alpha c_1^2 \overline{L}^2 n^{1/3}} ) }{ (k\alpha c_1  n^{1/3}) (k\alpha-1) c_1 - 1 } \\
\overset{(a)}{\leq}&  \frac{1}{k\alpha^2 c_1^2 \overline{L} n^{4/3}} + \frac{ \frac{1}{k \alpha c_1^2 \overline{L} n^{1/3}} (\frac{2}{k\alpha n}  +1 ) }{ 2(\alpha c_1  n^{1/3}) - 1 } \\
\leq & \frac{1}{k^2\alpha c_1^2 \overline{L} n^{4/3} } + \frac{1}{4 k \alpha^2 c_1^3\overline{L} n^{2/3} } \leq  \frac{3/4}{\alpha c_1^2 \overline{L} n^{2/3} }\eqsp,
\end{split}
\eeq
where $(a)$ is due to $c_1(k\alpha-1) \geq c_1(\alpha-1) \geq 2$ and $k\alpha c_1 n^{1/3} \geq 1$.
Note also that
$$
 -(\upsilon_{\min}\rho+\upsilon_{\max}^2) \leq  -\rho \upsilon_{\min} = -\frac{1}{\alpha c_1^2 \overline{L}n^{2/3}} \eqsp,
 $$
which yields that
 $$
 \left[ -(\upsilon_{\min}\rho+\upsilon_{\max}^2) + \gamma_{k+1}\rho^2 \Lip{V} + \frac{\rho^2\gamma_{k+1} \Lip{V}\Lip{\bss}^2\left(2 \gamma_{k+1}^2 \rho^2 + \frac{\gamma_{k+1} \rho^2}{\beta}\right)}{\Lambda_{(k+1)}} \right] \leq -\frac{1/4}{\alpha c_1^2 \overline{L} n^{2/3} }\eqsp.
  $$
Using the Lemma~\ref{lem:growth}, we know that $\upsilon_{\max}^2 \| \grd V( \hs{k} ) \|^2 \leq \| \hs{k} - \os^{(k)} \|^2$ and using~\eqref{eq:stepsizeineq} on~\eqref{eq:finalboundfi} yields:

\beq\notag
\begin{split}
\upsilon_{\max}^2 \sum_{k=0}^{{ K}_{ m }-1}\gamma_{k+1} \EE[ \| \grd V( \hs{k} ) \|^2 ]
\leq &  \frac{4\alpha  \overline{L} n^{2/3}}{\upsilon_{\min}^2} [ V(\hat{\bss}^{(0)} )  - V(\hat{\bss}^{({ K}_{ m })}) ]\\
&   + \frac{4\alpha  \overline{L} n^{2/3}}{\upsilon_{\min}^2} \sum_{k=0}^{{ K}_{ m }-1} \Xi^{(k+1)}  +  \sum_{k=0}^{{ K}_{ m }-1}\Gamma^{(k+1)} \EE[\| \hs{k} - \tilde{S}^{(k)}\|^2] \eqsp,
\end{split}
\eeq
proving the bound on the second order moment of the gradient of the Lyapunov function:
\beq\notag
\begin{split}
\sum_{k=0}^{{ K}_{ m }-1}\gamma_{k+1} \EE[ \| \grd V( \hs{k} ) \|^2 ]  \leq& \frac{4\alpha  \overline{L} n^{2/3}}{\upsilon_{\min}^2\upsilon_{\max}^2}  [ V(\hat{\bss}^{(0)} )  - V(\hat{\bss}^{({ K}_{ m })}) ]\\
 &   + \frac{4\alpha  \overline{L} n^{2/3}}{\upsilon_{\min}^2\upsilon_{\max}^2} \sum_{k=0}^{{ K}_{ m }-1} \Xi^{(k+1)}  +  \sum_{k=0}^{{ K}_{ m }-1}\Gamma^{(k+1)} \EE[\| \hs{k} - \tilde{S}^{(k)}\|^2]\eqsp.
\end{split}
\eeq

\end{proof}

\end{document}